\documentclass[10pt,twocolumn,letterpaper]{article}

\usepackage{cvpr}              
\usepackage[dvipsnames]{xcolor}

\usepackage{url}
\usepackage{xspace}
\usepackage{subcaption}
\usepackage{graphicx}
\usepackage[autolanguage]{numprint}
\usepackage[utf8]{inputenc}
\usepackage{microtype}
\usepackage{soul}
\usepackage{float}
\usepackage{textcomp, gensymb}
\usepackage{amssymb}
\usepackage{booktabs}
\usepackage{array}
\usepackage{relsize}
\usepackage[export]{adjustbox}

\definecolor{cvprblue}{rgb}{0.21,0.49,0.74}
\usepackage[pagebackref,breaklinks,colorlinks,citecolor=cvprblue]{hyperref}

\def\paperID{7877} 
\def\confName{CVPR}
\def\confYear{2024}

\newcommand{\websiteURL}[0]{https://gsgen3d.github.io/}
\newcommand{\websiteURLText}[0]{gsgen3d.github.io}
\newcommand{\codeURL}[0]{https://github.com/gsgen3d/gsgen}

\newcommand{\websiteLink}[0]{{\color{blue}{\href{\websiteURL}{\texttt{\websiteURLText}}}}\xspace}
\newcommand{\codeLink}[0]{{\color{blue}{\href{\codeURL}{\texttt{\codeURL}}}}\xspace}
\newcommand{\approach}[0]{\textsc{Gsgen}\xspace}

\title{Text-to-3D using Gaussian Splatting}

\author{
Zilong Chen, \quad Feng Wang, \quad Yikai Wang, \quad Huaping Liu$^\dagger$\\
Beijing National Research Center for Information Science and Technology (BNRist),\\
Department of Computer Science and Technology, Tsinghua University\\
{\tt\small chenzl22@mails.tsinghua.edu.cn, hpliu@tsinghua.edu.cn}\\
Project Page: \websiteLink
}

\begin{document}

\twocolumn[{%
\maketitle
\begin{figure}[H]
    \hsize=\textwidth
    \centering
    \vspace{-10mm}
    \includegraphics[width=0.9\textwidth]{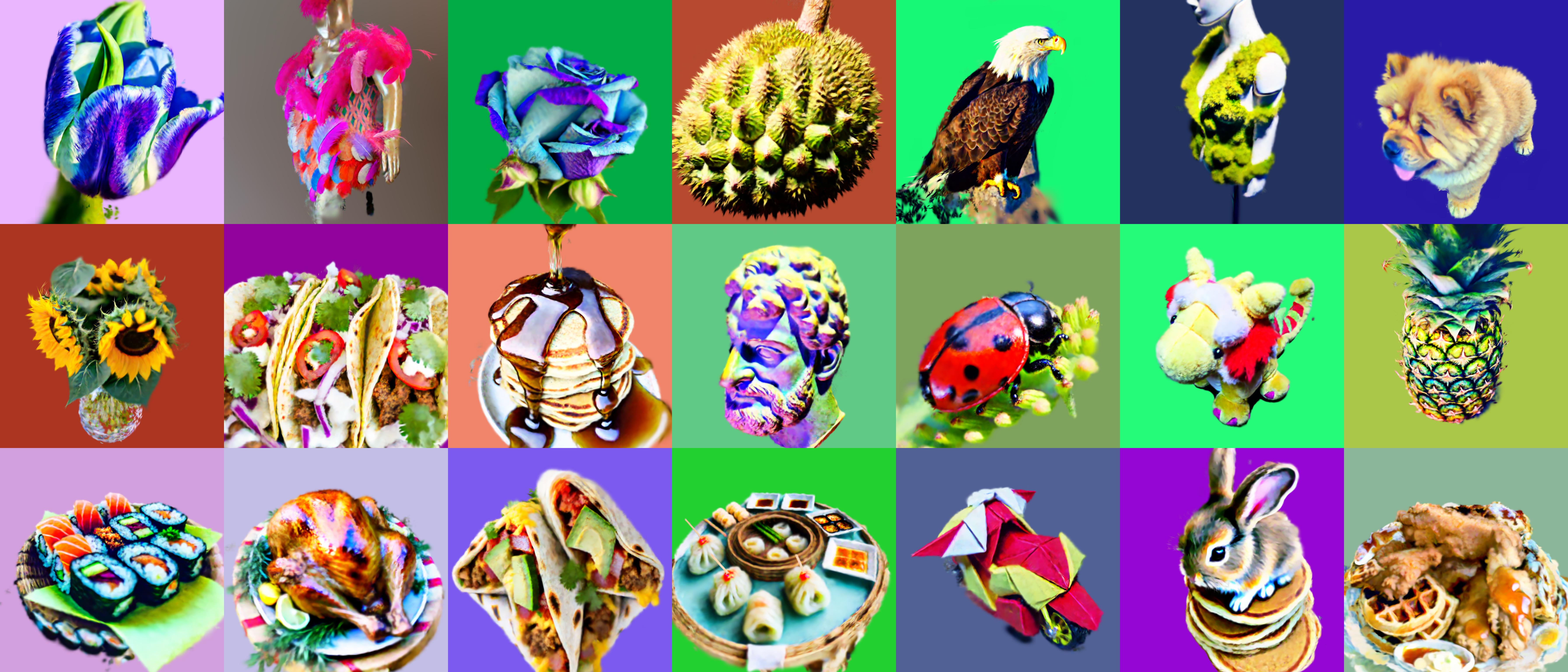}
    \begin{minipage}{0.9\textwidth}
        \vspace{3mm}
        \caption{Delicate 3D assets generated by the proposed \approach. See the project page for videos.}
    \end{minipage}
    \label{fig:gallery}
\end{figure}
}]


\let\thefootnote\relax\footnotetext{$^\dagger$Corresponding author}
\begin{abstract}
    Automatic text-to-3D generation that combines Score Distillation Sampling (SDS) with the optimization of volume rendering has achieved remarkable progress in synthesizing realistic 3D objects. Yet most existing text-to-3D methods by SDS and volume rendering suffer from inaccurate geometry, e.g., the Janus issue, since it is hard to explicitly integrate 3D priors into implicit 3D representations. Besides, it is usually time-consuming for them to generate elaborate 3D models with rich colors. In response, this paper proposes \approach, a novel method that adopts Gaussian Splatting, a recent state-of-the-art representation, to text-to-3D generation. \approach aims at generating high-quality 3D objects and addressing existing shortcomings by exploiting the explicit nature of Gaussian Splatting that enables the incorporation of 3D prior. Specifically, our method adopts a progressive optimization strategy, which includes a geometry optimization stage and an appearance refinement stage. In geometry optimization, a coarse representation is established under 3D point cloud diffusion prior along with the ordinary 2D SDS optimization, ensuring a sensible and 3D-consistent rough shape. Subsequently, the obtained Gaussians undergo an iterative appearance refinement to enrich texture details. In this stage, we increase the number of Gaussians by compactness-based densification to enhance continuity and improve fidelity. With these designs, our approach can generate 3D assets with delicate details and accurate geometry. Extensive evaluations demonstrate the effectiveness of our method, especially for capturing high-frequency components. Our code is available at \codeLink. 
    
\end{abstract}

\begin{figure*}[t]
    \vspace{-9mm}
    \centering
    \begin{minipage}{0.33\textwidth}
        \centering
        \begin{subfigure}{\linewidth}
            \centering
            \includegraphics[width=1.0\linewidth]{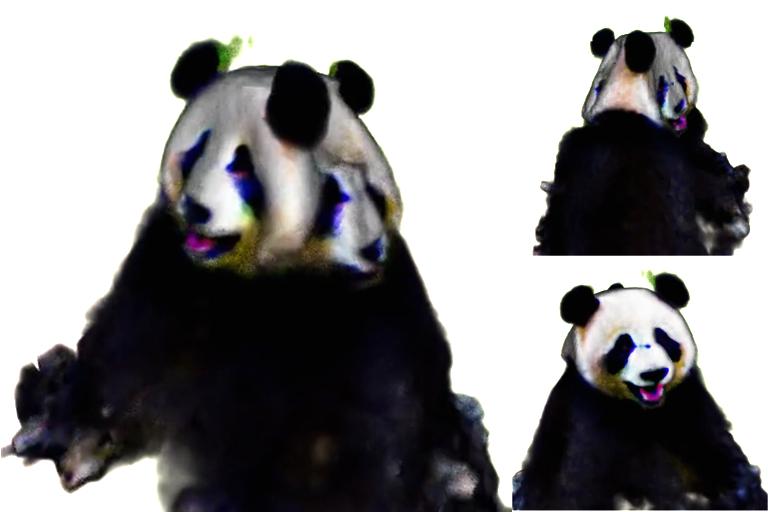} 
            \caption*{Stable DreamFusion~\citep{stable-dreamfusion, dreamfusion}}
        \end{subfigure}
    \end{minipage}
    \hfill
    \begin{minipage}{0.33\textwidth}
        \centering
        \begin{subfigure}{\linewidth}
            \centering
            \includegraphics[width=1.0\linewidth]{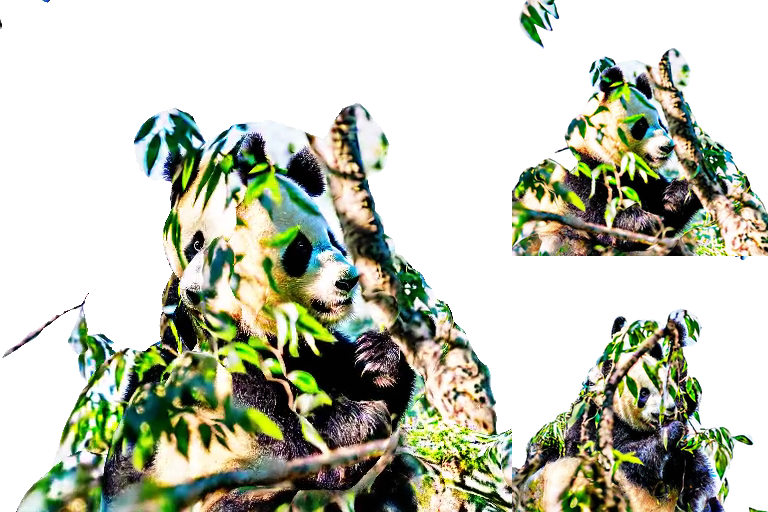} 
            \caption*{threestudio ProlificDreamer~\citep{prolificdreamer, threestudio2023}}
        \end{subfigure}
    \end{minipage}
    \hfill
    \begin{minipage}{0.33\textwidth}
        \centering
        \begin{subfigure}{\linewidth}
            \centering
            \includegraphics[width=1.0\linewidth]{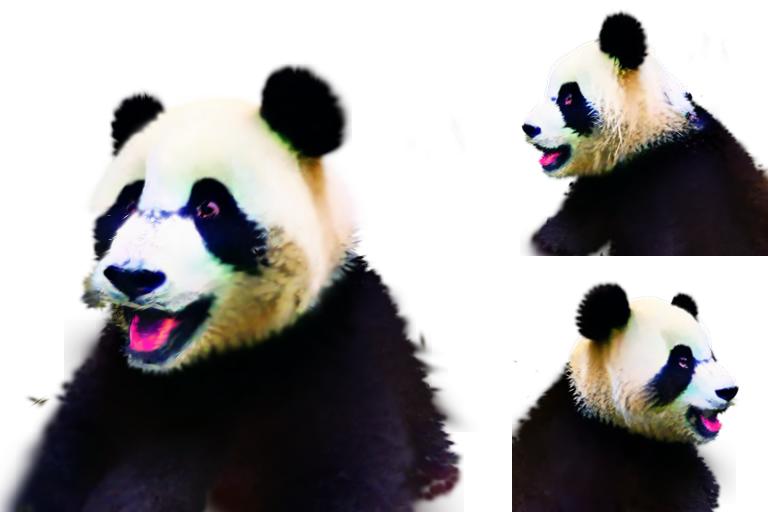} 
            \caption*{\approach (Ours)}
        \end{subfigure}
    \end{minipage}
    \\ \textit{\small A DSLR photo of a panda}
    
    \begin{minipage}{0.33\textwidth}
        \centering
        \begin{subfigure}{\linewidth}
            \centering
            \includegraphics[width=1.0\linewidth]{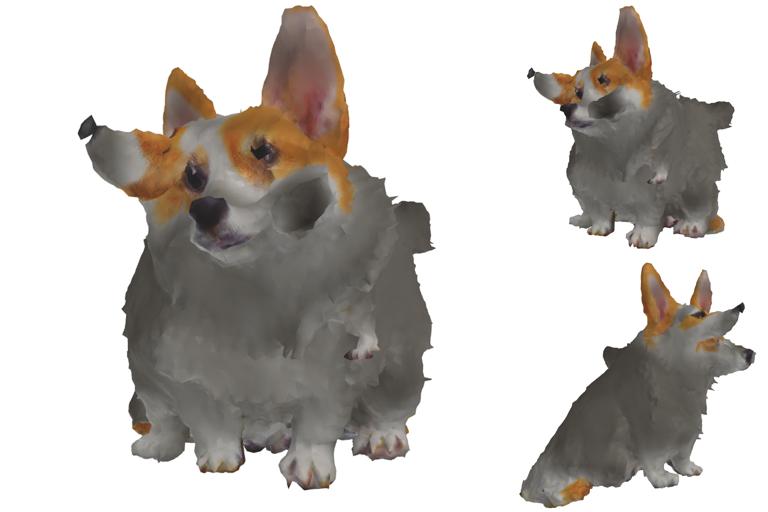} 
            \caption*{Fantasia3D~\citep{fantasia3d}}
        \end{subfigure}
    \end{minipage}
    \hfill
    \begin{minipage}{0.33\textwidth}
        \centering
        \begin{subfigure}{\linewidth}
            \centering
            \includegraphics[width=1.0\linewidth]{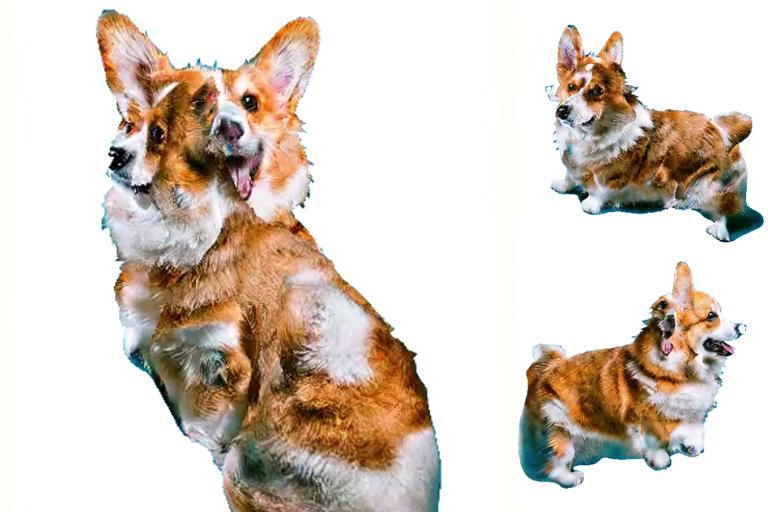} 
            \caption*{threestudio ProlificDreamer~\citep{prolificdreamer, threestudio2023}}
        \end{subfigure}
    \end{minipage}
    \hfill
    \begin{minipage}{0.33\textwidth}
        \centering
        \begin{subfigure}{\linewidth}
            \centering
            \includegraphics[width=1.0\linewidth]{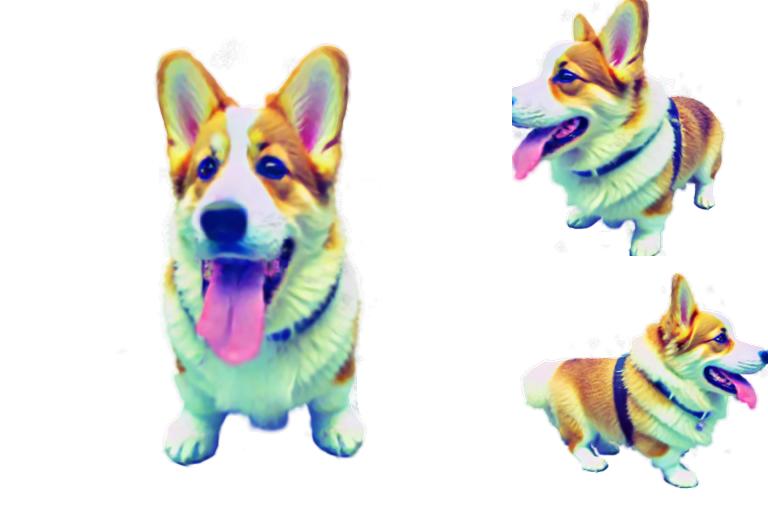} 
            \caption*{\approach (Ours)}
        \end{subfigure}
    \end{minipage}
    \\ \textit{\small A high quality photo of a furry corgi}

    \begin{minipage}{0.33\textwidth}
        \centering
        \begin{subfigure}{\linewidth}
            \centering
            \includegraphics[width=1.0\linewidth]{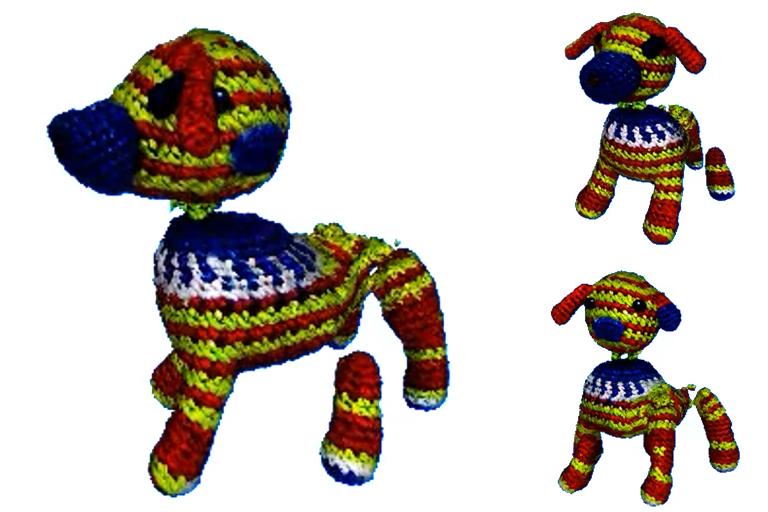} 
            \caption*{threestudio Magic3D \citep{threestudio2023, magic3d}}
        \end{subfigure}
    \end{minipage}
    \hfill
    \begin{minipage}{0.33\textwidth}
        \centering
        \begin{subfigure}{\linewidth}
            \centering
            \includegraphics[width=1.0\linewidth]{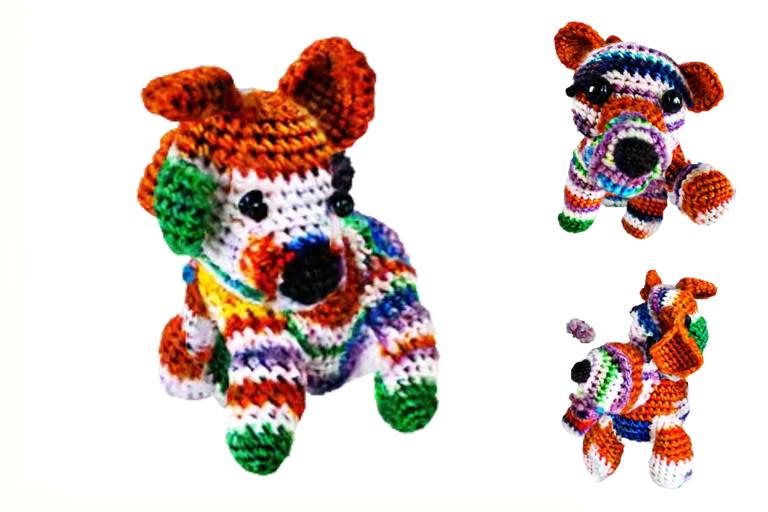} 
            \caption*{threestudio ProlificDreamer~\citep{threestudio2023, prolificdreamer}}
        \end{subfigure}
    \end{minipage}
    \hfill
    \begin{minipage}{0.33\textwidth}
        \centering
        \begin{subfigure}{\linewidth}
            \centering
            \includegraphics[width=1.0\linewidth]{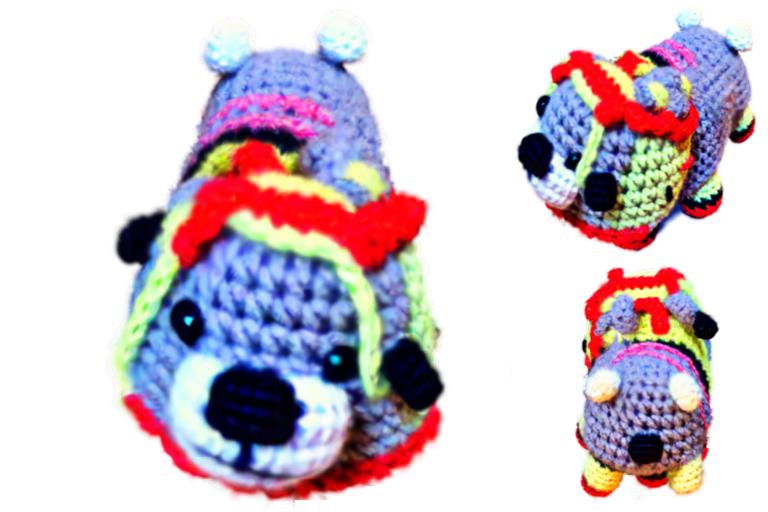} 
            \caption*{\approach (Ours)}
        \end{subfigure}
    \end{minipage}
    \\ \textit{\small A zoomed out DSLR photo of an amigurumi bulldozer}
    
    \caption{Compared to previous methods, \approach alleviates the Janus problem by representing the 3D scene using 3D Gaussian Splatting, which is capable of applying direct 3D geometry guidance and expressing content with delicate details. Note that the results of DreamFusion, Magic3D, and ProlificDreamer are obtained using Stable DreamFusion \citep{stable-dreamfusion} and threestudio \citep{threestudio2023} since the official implementations have not been publicly available till the date of this work.}
    \label{fig:janus}
    \vspace{-4mm}
\end{figure*}
\section{Introduction}
\label{sec:intro}
Diffusion model based text-to-image generation \citep{imagen, stablediffusion, dalle2, Alex2023deep} has achieved remarkable success in synthesizing photo-realistic images from textual prompts. Nevertheless, for high-quality text-to-3D content generation, the advancements lag behind that of image generation due to the inherent complexity of real-world 3D scenes.
Recently, DreamFusion \citep{dreamfusion} has made great progress in generating delicate assets by utilizing score distillation sampling with a pre-trained text-to-image diffusion prior. Its follow-up works further improve this paradigm in quality \citep{prolificdreamer, fantasia3d}, training speed \citep{magic3d, latentnerf}, and generating more reasonable geometry \citep{perpneg, hifa, seo2023let}. 
However, most existing text-to-3D methods still suffer greatly from collapsed geometry and limited fidelity, and are difficult to incorporate 3D priors due to the implicit nature of NeRF \citep{nerf} and \textsc{DMTet} \citep{dmtet}.

Recently, 3D Gaussian Splatting \citep{kerbl3Dgaussians} has garnered significant attention in the field of 3D reconstruction, primarily due to its remarkable ability to represent intricate scenes and capability of real-time rendering. By modeling a scene using a set of 3D Gaussians, \citet{kerbl3Dgaussians} adopt an explicit and object-centric approach that fundamentally diverges from implicit representations like NeRF and \textsc{DMTet}. This distinctive approach paves the way for the integration of explicit 3D priors into text-to-3D generation. Building upon this insight, instead of a straightforward replacement of NeRFs with Gaussians, we propose to guide the generation with an additional 3D point cloud diffusion prior to enhancing geometrical coherence. By adopting this strategy, we can better harness the inherent advantages of 3D Gaussians in the creation of complex and 3D-consistent assets. 

Specifically, we propose to represent the generated 3D content with a set of Gaussians and optimize them progressively in two stages, namely geometry optimization and appearance refinement. In the geometry optimization stage, we optimize the Gaussians under the guidance of a 3D point cloud diffusion prior along with the ordinary 2D image prior. The incorporation of this extra 3D SDS loss ensures a 3D-consistent rough geometry. In the subsequent refinement stage, the Gaussians undergo an iterative enhancement to enrich delicate details. Due to the sub-optimal performance of the original adaptive control under SDS loss, we introduce an additional compactness-based densification technique to enhance appearance and fidelity. Besides, to prevent potential degeneration and break the symmetry in the early stage, the Gaussians are initialized with a coarse point cloud generated by a text-to-point-cloud diffusion model. As a result of these techniques, our approach can generate 3D assets with consistent geometry and exceptional fidelity. Fig.~\ref{fig:janus} illustrates a comparison between \approach and previous state-of-the-art methods on generating assets with asymmetric geometry. In summary, our contributions are:
\begin{itemize}
    \item We propose \approach, a text-to-3D generation method using 3D Gaussians as representation. By incorporating direct geometric priors, we highlight the distinctive advantages of Gaussian Splatting in text-to-3D generation.
    \item We introduce a two-stage optimization strategy that first exploits joint guidance of 2D and 3D diffusion prior to shaping a coherent rough structure in geometry optimization; then enriches the details with compactness-based densification in appearance refinement.
    \item We validate \approach on various textual prompts. Experiments show that our method can generate 3D assets with accurate geometry and enhanced fidelity. Especially, \approach demonstrates superior performance in capturing \textit{high-frequency components}, such as feathers, surfaces with intricate textures, animal fur, etc.
\end{itemize}

\section{Related Work}
\subsection{3D Scene Representations}
Representing 3D scenes in a differentiable way has achieved remarkable success in recent years. NeRFs \citep{nerf} demonstrates outstanding performance in novel view synthesis by representing 3D scenes with a coordinate-based neural network. After works have emerged to improve NeRF in reconstruction quality \citep{barron2021mip, barron2023zipnerf, wang20224k}, handling large-scale \citep{tancik2022block, zhang2020nerf++, martinbrualla2020nerfw, chen2022hallucinated} and dynamic scenes \citep{park2021hypernerf, attal2023hyperreel, wang2022mixed, kplanes_2023, pumarola2021d}, improving training \citep{yu2021plenoxels, chen2022tensorf, sun2022direct, mueller2022instant}, rendering \citep{merf, hedman2021snerg, yu2021plenoctrees} speed and facilitating down-stream tasks~\citep{ye2023learning, GraspNeRF, zhou2023general, ozer2023low, zhong2023touching}. 
Although great progress has been made, NeRF-based methods still suffer from low rendering speed and high training-time memory usage due to their implicit nature. 
To tackle these challenges, \citet{kerbl3Dgaussians} propose to represent the 3D scene as a set of anisotropic Gaussians and render novel views using GPU-optimized tile-based rasterization. Gaussian Splatting could achieve better reconstruction results while being capable of real-time rendering. Our research highlights the distinctive advantages of Gaussian Splatting within text-to-3D by incorporating explicit 3D prior, generating 3D consistent and highly detailed assets.

\subsection{Diffusion Models}
Diffusion models have arisen as a promising paradigm for learning and sampling from a complex distribution. Inspired by the diffusion process in physics, these models involve a forward process to gradually add noise and an inverse process to denoise a noisy sample with a trained neural network. After DDPM \citep{ddpm, sde} highlights the effectiveness of diffusion models in capturing real-world image data, a plethora of research has emerged to improve the inherent challenges, including fast sampling \citep{lu2022dpm, DBLP:conf/iclr/BaoLZZ22, ddim} and architecture improvements \citep{bao2022all, SDXL, liu2023insta, guided_diffusion, simple_diffusion, Peebles2022DiT}. One of the most successful applications of diffusion models lies in text-to-image generation, where they have shown remarkable progress in generating realistic images from text prompts \citep{cfg, dalle2, Alex2023deep}. 
To produce high-resolution images, current methods utilize either a cascaded structure combining a low-resolution diffusion model with super-resolution models \citep{imagen, ediffi, Alex2023deep} or train a diffusion model in latent space using an auto-encoder \citep{stablediffusion, DBLP:conf/cvpr/GuCBWZCYG22}.
Our proposed \approach is built upon StableDiffusion \citep{stablediffusion}, an open-source latent diffusion model that provides fine-grained guidance for delicate 3D content generation.

\begin{figure*}[t]
    \vspace{-8mm}
    \centering
    \includegraphics[width=1.0\textwidth]{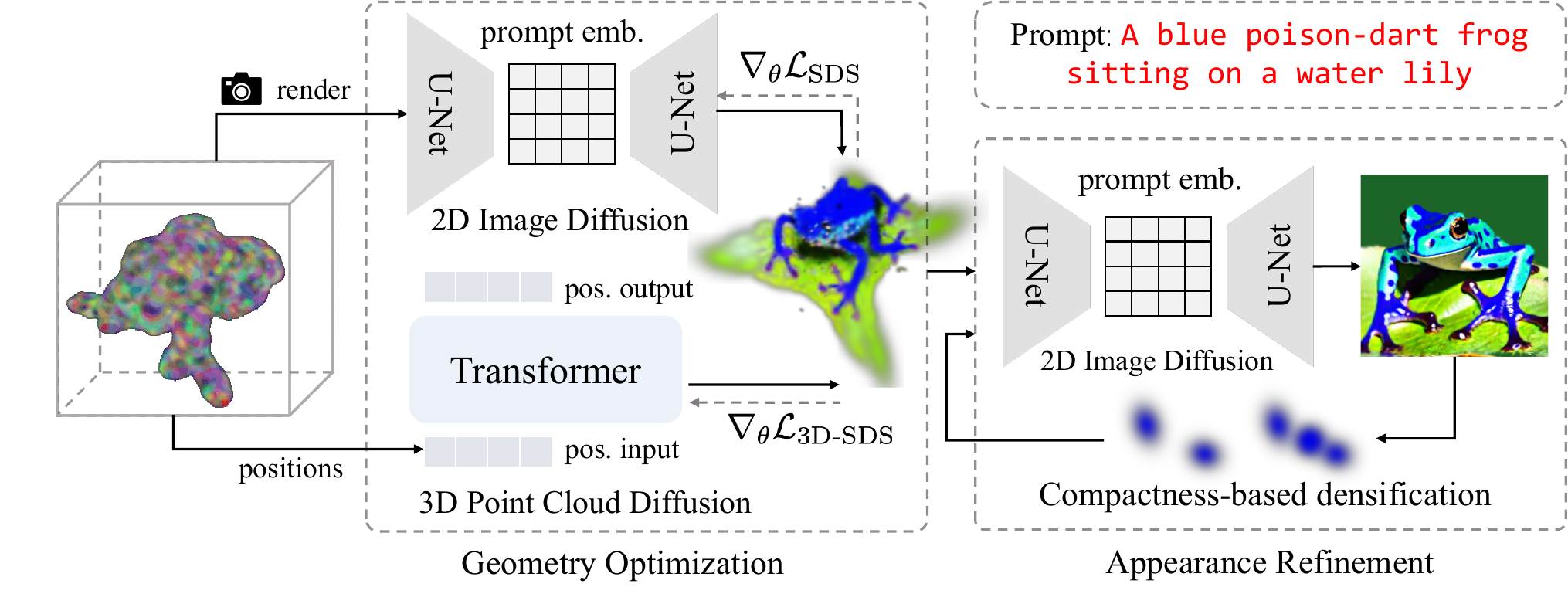}
    \caption{\textbf{Overview of the proposed \approach}. Our approach aims at generating 3D assets with accurate geometry and delicate appearance. \textsc{Gsgen} starts by utilizing Point-E to initialize the positions of the Gaussians (Sec \ref{sec:initialize}). The optimization is grouped into geometry optimization (Sec \ref{sec:geometrical_optimization}) and appearance refinement (Sec \ref{sec:appearance_refinement}) to meet a balance between coherent geometry structure and detailed texture. }
    \label{fig:method}
\end{figure*}

\subsection{Text-to-3D Generation}
Early efforts in text-to-3D generation, including CLIP-forge \citep{sanghi2021clip}, Dream Fields \citep{jain2021dreamfields}, Text2Mesh \citep{text2mesh}, TANGO \citep{TANGO}, CLIPNeRF \citep{clipnerf}, and CLIP-Mesh \citep{khalid2022clipmesh}, harness CLIP \citep{CLIP} guidance to create 3D assets. To leverage the stronger diffusion prior, DreamFusion \citep{dreamfusion} introduces score distillation sampling that optimizes the 3D content by minimizing the difference between rendered images and the diffusion prior. This development sparked a surge of interest in text-to-3D generation through image diffusion prior \citep{sjc, raj2023_dreambooth3d, lorraine2023_att3d, hifa, qian2023magic123, chen2023it3d}. Magic3D \citep{magic3d} employs a coarse-to-fine strategy, optimizing a NeRF with a low-resolution diffusion prior and then enhancing texture under latent diffusion prior with a \textsc{DMTet} initialized with the coarse NeRF.
Latent-NeRF \citep{latentnerf} trains a NeRF within the latent space of StableDiffusion and introduces the Sketch-Shape method to guide the generation process. Fantasia3D \citep{fantasia3d} disentangles the learning of geometry and material, harnessing physics-based rendering techniques to achieve high-fidelity mesh generation.
ProlificDreamer \citep{prolificdreamer} introduces variational score distillation to improve SDS and facilitate the generation of high-quality and diverse 3D assets. 
Our concurrent work DreamGaussian \citep{tang2023dreamgaussian} achieves fast image-to-3D by capitalizing on the rapid convergence of Gaussian Splatting, whose contribution is orthogonal to ours since we focus on incorporating 3D prior with more advanced representation.
Another line of work lies in training or fine-tuning diffusion models directly on 3D datasets (e.g. ShapeNet~\citep{shapenet} and Objaverse~\citep{objaverse, objaverse-xl}) to achieve more consistent results with advanced guidance~\citep{Wang2023rodin, shap_e, liu2023_zero1to3, cheng2023_sdfusion, zheng2023lasdiffusion, 3dgen, sdf_diffusion, ssdnerf, shi2023MVDream, sweetdreamer, liu2023syncdreamer}.
Our approach builds upon Point-E \citep{pointe}, a text-to-point-cloud diffusion model trained on millions of 3D models, which offers valuable 3D guidance and coarse initialization.


\section{Preliminary}
\subsection{Score Distillation Sampling}
Instead of directly generating 3D models, recent studies have achieved notable success by optimizing 3D representation with a 2D pre-trained image diffusion prior based on score distillation sampling, as proposed by \citet{dreamfusion}. In this paradigm, the scene is represented as a differentiable image parameterization (DIP) denoted as $\theta$, where the image can be differentiably rendered based on the given camera parameters through a transformation function $g$. The DIP $\theta$ is iteratively refined to ensure that, for any given camera pose, the rendered image $\mathbf{x}=g(\theta)$ closely resembles a plausible sample derived from the guidance diffusion model. DreamFusion achieves this by leveraging Imagen \citep{imagen} to provide a score estimation function denoted as $\epsilon_{\phi}(x_t;y,t)$, where $x_t$, $y$, and $t$ represent the noisy image, text embedding, and timestep, respectively. This estimated score plays a pivotal role in guiding the gradient update, as expressed by the following equation:
\begin{equation}
    \nabla_{\theta}\mathcal{L}_{\text{SDS}}=\mathbb{E}_{\epsilon, t}\left[w(t)(\epsilon_{\phi}(x_t;y,t)-\epsilon)\frac{\partial\mathbf{x}}{\partial\theta}\right]
\end{equation}
where $\epsilon$ is a Gaussian noise and $w(t)$ is a weighting function. 
Our approach combines score distillation sampling with 3D Gaussian Splatting at both 2D and 3D levels with different diffusion models to generate 3D assets with both detailed appearance and 3D-consistent geometry.

\subsection{3D Gaussian Splatting}
Gaussian Splatting, as introduced in \citet{kerbl3Dgaussians}, presents a pioneering method for novel view synthesis and 3D reconstruction from multi-view images. Unlike NeRF, 3D Gaussian Splatting adopts a distinctive approach, where the underlying scene is represented through a set of anisotropic 3D Gaussians parameterized by their positions, covariances, colors, and opacities. When rendering, the 3D Gaussians are projected onto the camera's imaging plane \citep{ewa}. Subsequently, the projected 2D Gaussians are assigned to individual tiles. The color of $\boldsymbol{p}$ on the image plane is rendered sequentially with point-based volume rendering technique \citep{ewa}:
\begin{equation}
    C(\boldsymbol{p})=\sum_{i \in \mathcal{N}} c_i \alpha_i \prod_{j=1}^{i-1}\left(1-\alpha_j\right) \quad
\end{equation}
where $\alpha_i=o_ie^{-\frac{1}{2}(\boldsymbol{p}-\mu_i)^T \Sigma_i^{-1}(\boldsymbol{p}-\mu_i)}$ refers to the opacity at point $\boldsymbol{p}$, $c_i$, $o_i$, $\mu_i$, and $\Sigma_i$ represent the color, opacity, position, and covariance of the $i$-th Gaussian respectively, $\mathcal{N}$ denotes the Gaussians in this tile.
To maximize the utilization of shared memory, Gaussian Splatting further designs a GPU-friendly rasterization process where each thread block is assigned to render an image tile. These advancements enable Gaussian Splatting to achieve more detailed scene reconstruction, significantly faster rendering speed, and reduction of memory usage during training compared to NeRF-based methods. In this study, we expand the application of Gaussian Splatting into text-to-3D generation and introduce a novel approach that leverages the explicit nature of Gaussian Splatting by integrating direct 3D diffusion priors, highlighting the potential of 3D Gaussians as a fundamental representation for generative tasks.


\section{Approach}
Our goal is to generate 3D content with accurate geometry and delicate detail. 
To accomplish this, \textsc{Gsgen} exploits the 3D Gaussians as representation due to its flexibility to incorporate geometry priors and capability to represent high-frequency details. Based on the observation that a point cloud can be seen as a set of isotropic Gaussians, we propose to integrate a 3D SDS loss with a pre-trained point cloud diffusion model to shape a 3D-consistent geometry. With this additional geometry prior, our approach could mitigate the Janus problem and generate more sensible geometry. 
Subsequently, in appearance refinement, the Gaussians undergo an iterative optimization to gradually improve fine-grained details with a compactness-based densification strategy, while preserving the fundamental geometric information. The detailed \textsc{Gsgen} methodology is presented as follows.

\subsection{Geometry Optimization}
\label{sec:geometrical_optimization}
Many text-to-3D methods encounter the significant challenge of overfitting to several views, resulting in assets with multiple faces and collapsed geometry \citep{dreamfusion, magic3d, fantasia3d}. This issue, known as the Janus problem \citep{perpneg, seo2023let}, has posed a persistent hurdle in the development of such approaches. In our early experiments, we faced a similar challenge that relying solely on 2D guidance frequently led to flawed results. However, we noticed that the geometry of 3D Gaussians can be directly rectified with a point cloud prior, which is not feasible for previous text-to-3D methods using NeRFs as their geometries are represented in implicit density functions. Recognizing this distinctive advantage, we introduce a geometry optimization process to shape a reasonable structure.
Concretely, in addition to the ordinary 2D image diffusion prior, we further optimize the positions of Gaussians using Point-E \citep{pointe} guidance, a pre-trained text-to-point-cloud diffusion model. Instead of directly aligning the Gaussians with a Point-E generated point cloud, we apply a 3D SDS loss to lead the positions inspired by image diffusion SDS, which avoids challenges including registration, scaling, and potential degeneration. 
We summarize the loss in the geometry optimization stage as the following equation:
\begin{equation}
\begin{split}
    \nabla_{\theta}\mathcal{L}_{\text{geometry}}&=\mathbb{E}_{\epsilon_{I}, t}\left[w_I(t)(\epsilon_{\phi}(x_t;y,t)-\epsilon_I)\frac{\partial\mathbf{x}}{\partial\theta}\right]\\
    &+\lambda_{\text{3D}}\cdot\mathbb{E}_{\epsilon_P, t}\left[w_P(t)(\epsilon_{\psi}(p_t;y,t)-\epsilon_P)\right],
\end{split}
\end{equation}
where $p_t$ and $x_t$ represent the noisy Gaussian positions and the rendered image, $w_*$ and $\epsilon_*$ refer to the corresponding weighting function and Gaussian noise.

\subsection{Appearance Refinement}
\label{sec:appearance_refinement}
While the introduction of 3D prior does help in learning a more reasonable geometry, we experimentally find it would also disturb the learning of appearance, resulting in insufficiently detailed assets. 
Based on this observation, \textsc{Gsgen} employs another appearance refinement stage that iteratively refines and densifies the Gaussians utilizing only the 2D image prior.

\begin{figure}[t]
    \centering
    \includegraphics[width=0.45\textwidth]{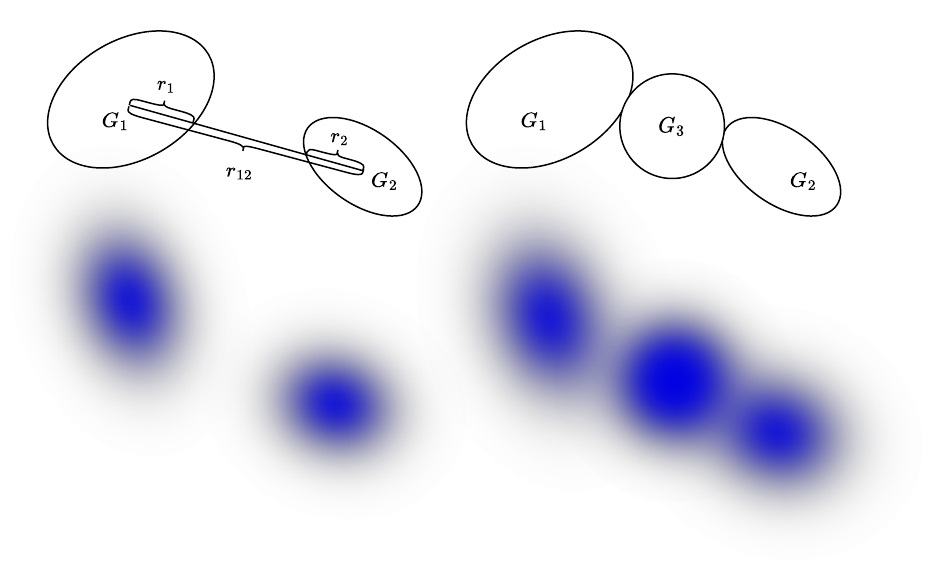}
    \vspace{-6mm}
    \caption{An illustration of the proposed compactness-based densification.}
    \vspace{-4mm}
    \label{fig:densification}
\end{figure}
To densify the Gaussians, \citet{kerbl3Dgaussians} propose to split Gaussians with a large view-space spatial gradient. However, we encountered challenges in determining the appropriate threshold for this spatial gradient under score distillation sampling. Due to the stochastic nature of SDS loss, employing a small threshold is prone to be misled by some stochastic large gradient thus generating an excessive number of Gaussians, whereas a large threshold will lead to a blurry appearance, as illustrated in Fig.~\ref{fig:ablation_densification}.

\begin{figure*}[t]
    \vspace{-10mm}
    \centering
    \begin{minipage}{0.32\textwidth}
        \centering
        \begin{subfigure}{\linewidth}
            \centering
            \caption*{\textit{Magic3D}}
            \includegraphics[width=1.0\linewidth]{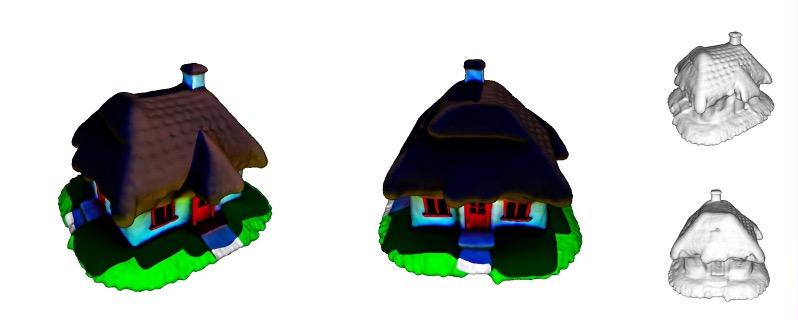} 
        \end{subfigure}
    \end{minipage}
    \hfill
    \begin{minipage}{0.32\textwidth}
        \centering
        \begin{subfigure}{\linewidth}
            \centering
            \caption*{\textit{Fantasia3D}}
            \includegraphics[width=1.0\linewidth]{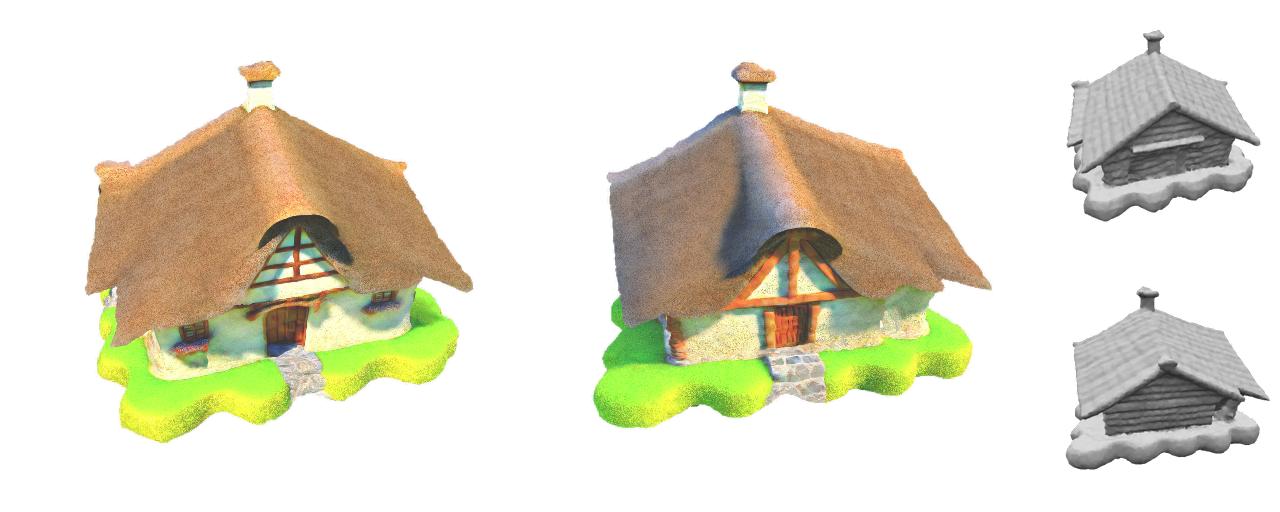} 
        \end{subfigure}
    \end{minipage}
    \hfill
    \begin{minipage}{0.32\textwidth}
        \centering
        \begin{subfigure}{\linewidth}
            \centering
            \caption*{\textit{\approach(ours)}}
            \includegraphics[width=1.0\linewidth]{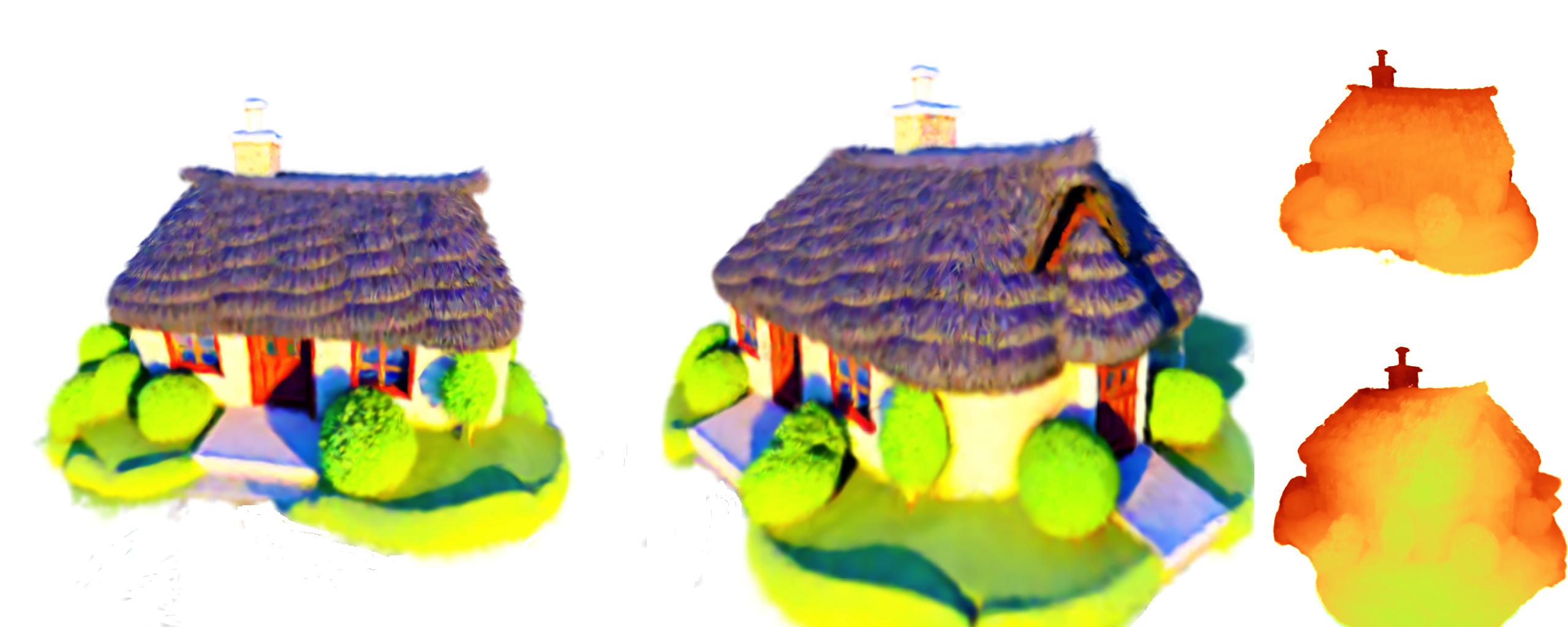} 
        \end{subfigure}
    \end{minipage}
    \\ \textit{\small A 3D model of an adorable cottage with a thatched roof.}\\
    \begin{minipage}{0.32\textwidth}
        \centering
        \begin{subfigure}{\linewidth}
            \centering
            \includegraphics[width=1.0\linewidth]{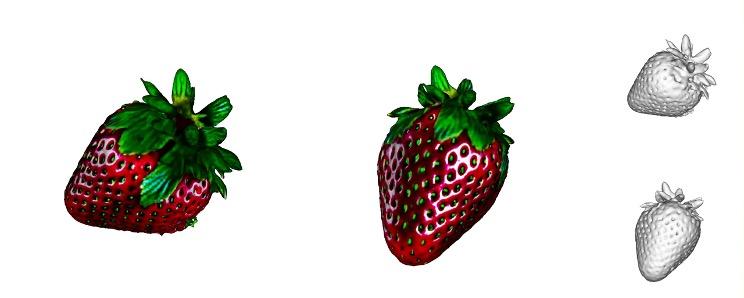} 
        \end{subfigure}
    \end{minipage}
    \hfill
    \begin{minipage}{0.32\textwidth}
        \centering
        \begin{subfigure}{\linewidth}
            \centering
            \includegraphics[width=1.0\linewidth]{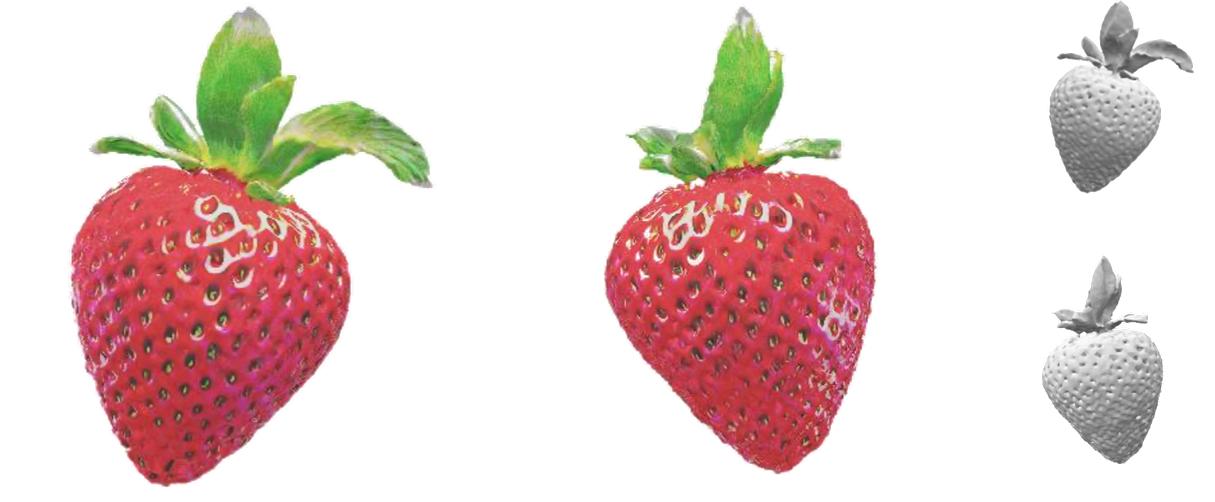} 
        \end{subfigure}
    \end{minipage}
    \hfill
    \begin{minipage}{0.32\textwidth}
        \centering
        \begin{subfigure}{\linewidth}
            \centering
            \includegraphics[width=1.0\linewidth]{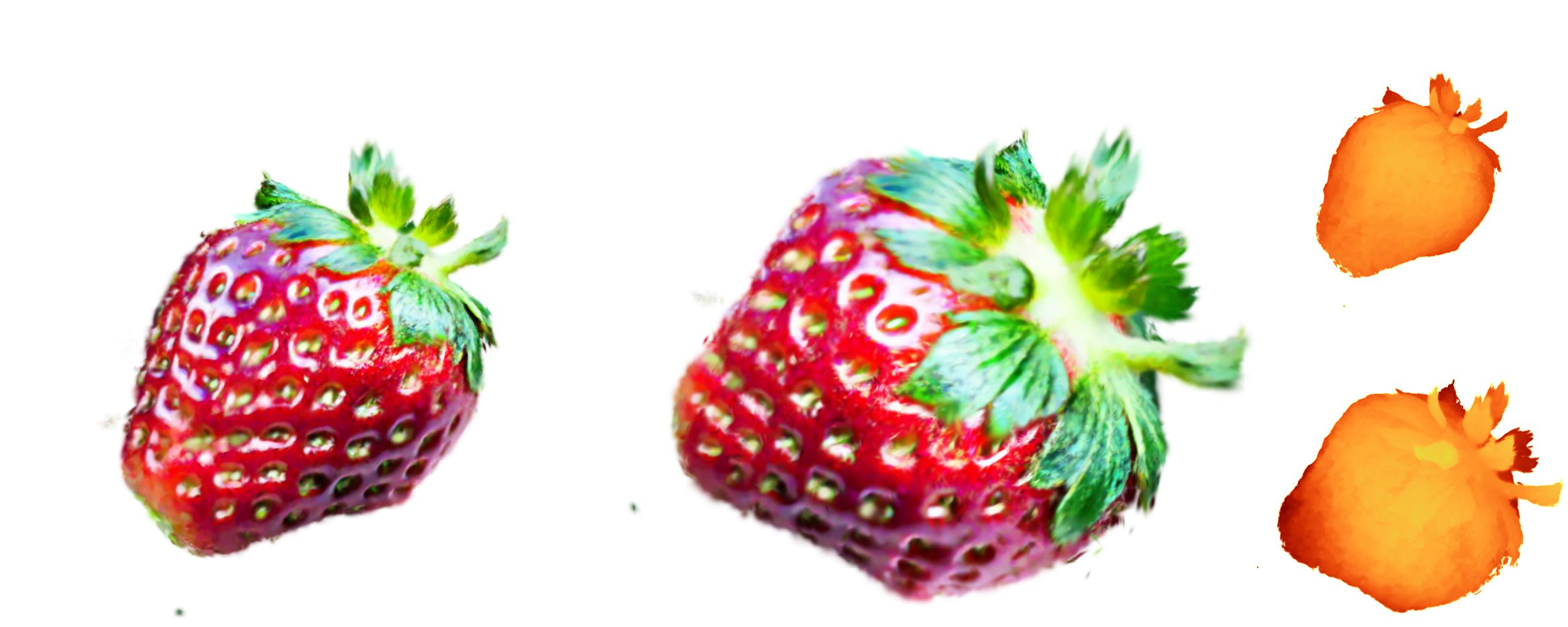} 
        \end{subfigure}
    \end{minipage}
    \\ \textit{\small A ripe strawberry} \\
    \begin{minipage}{0.32\textwidth}
        \centering
        \begin{subfigure}{\linewidth}
            \centering
            \includegraphics[width=1.0\linewidth]{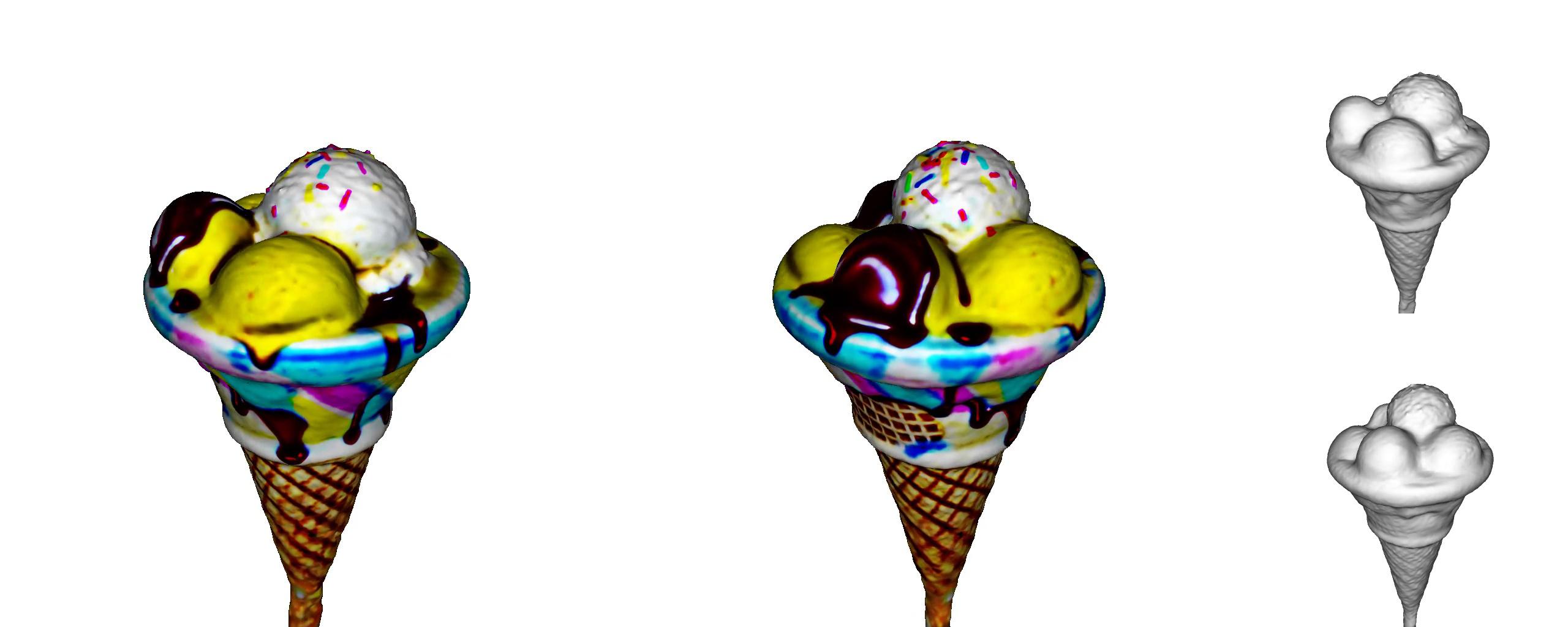} 
        \end{subfigure}
    \end{minipage}
    \hfill
    \begin{minipage}{0.32\textwidth}
        \centering
        \begin{subfigure}{\linewidth}
            \centering
            \includegraphics[width=1.0\linewidth]{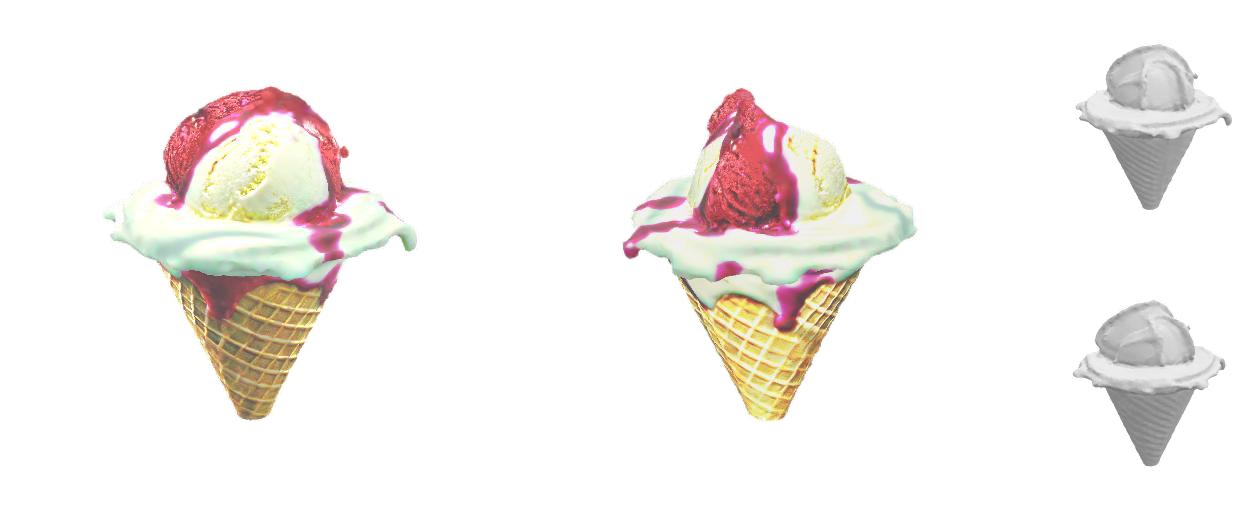} 
        \end{subfigure}
    \end{minipage}
    \hfill
    \begin{minipage}{0.32\textwidth}
        \centering
        \begin{subfigure}{\linewidth}
            \centering
            \includegraphics[width=1.0\linewidth]{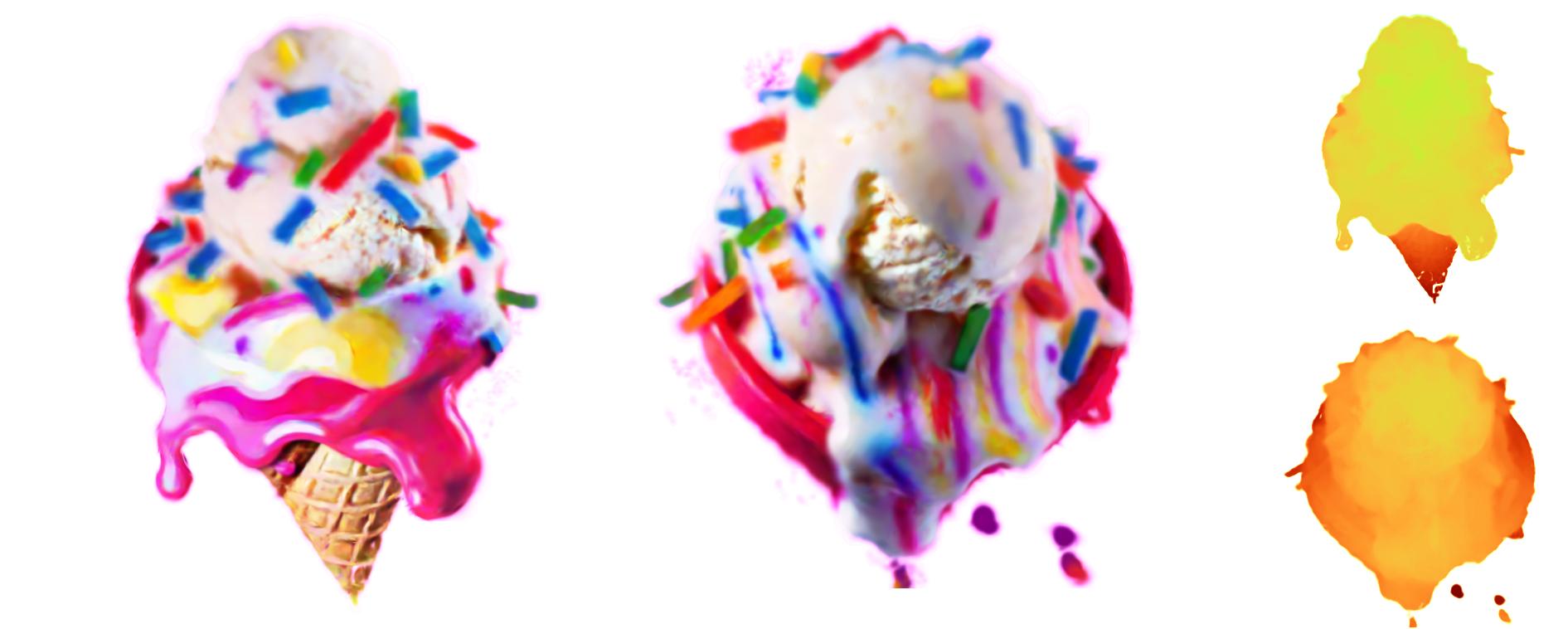} 
        \end{subfigure}
    \end{minipage}
    \\ \textit{\small A DSLR photo of an ice cream sundae} \\
    \begin{minipage}{0.32\textwidth}
        \centering
        \begin{subfigure}{\linewidth}
            \centering
            \includegraphics[width=1.0\linewidth]{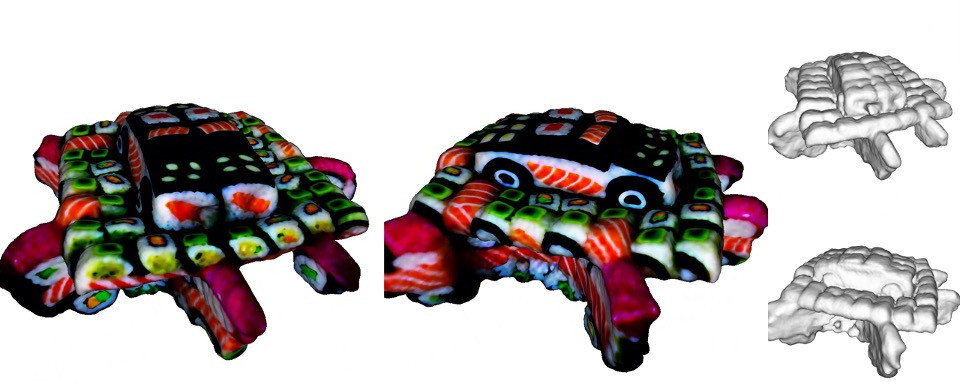} 
        \end{subfigure}
    \end{minipage}
    \hfill
    \begin{minipage}{0.32\textwidth}
        \centering
        \begin{subfigure}{\linewidth}
            \centering
            \includegraphics[width=1.0\linewidth]{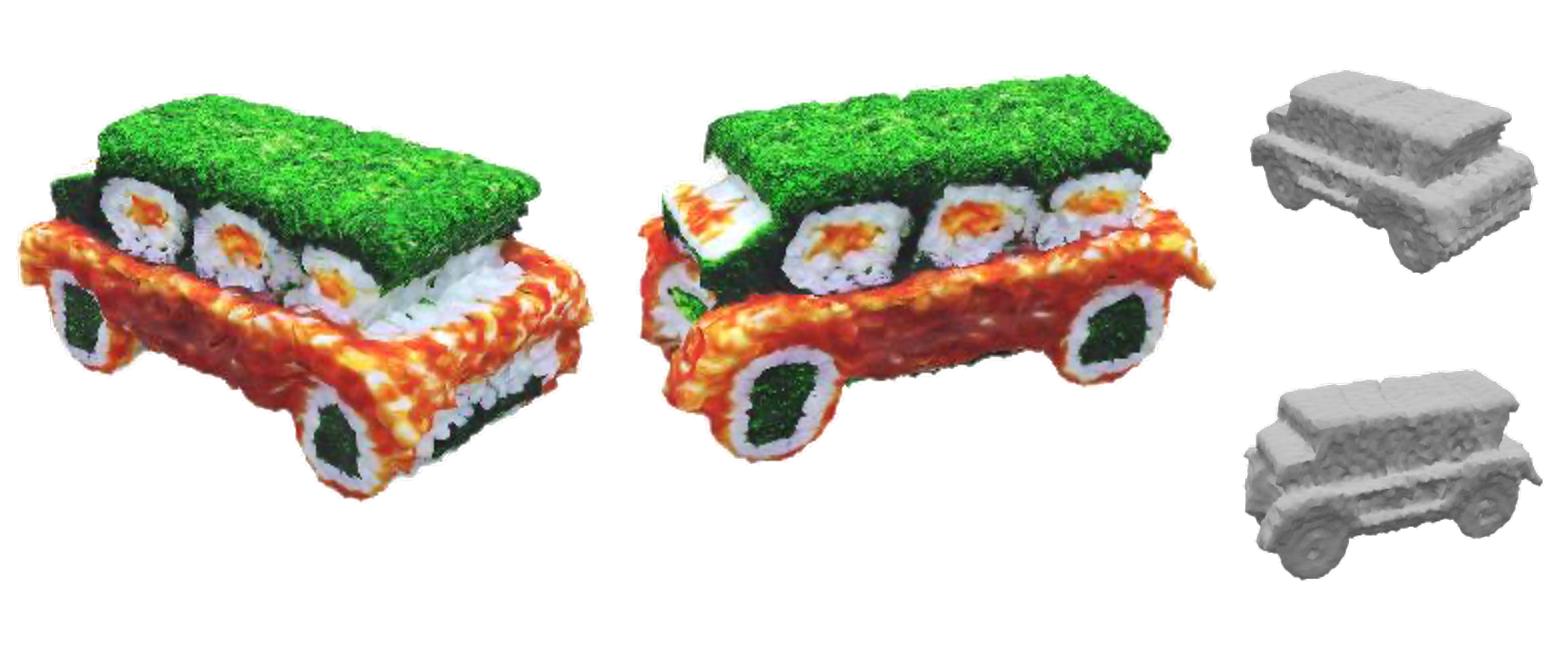} 
        \end{subfigure}
    \end{minipage}
    \hfill
    \begin{minipage}{0.32\textwidth}
        \centering
        \begin{subfigure}{\linewidth}
            \centering
            \includegraphics[width=1.0\linewidth]{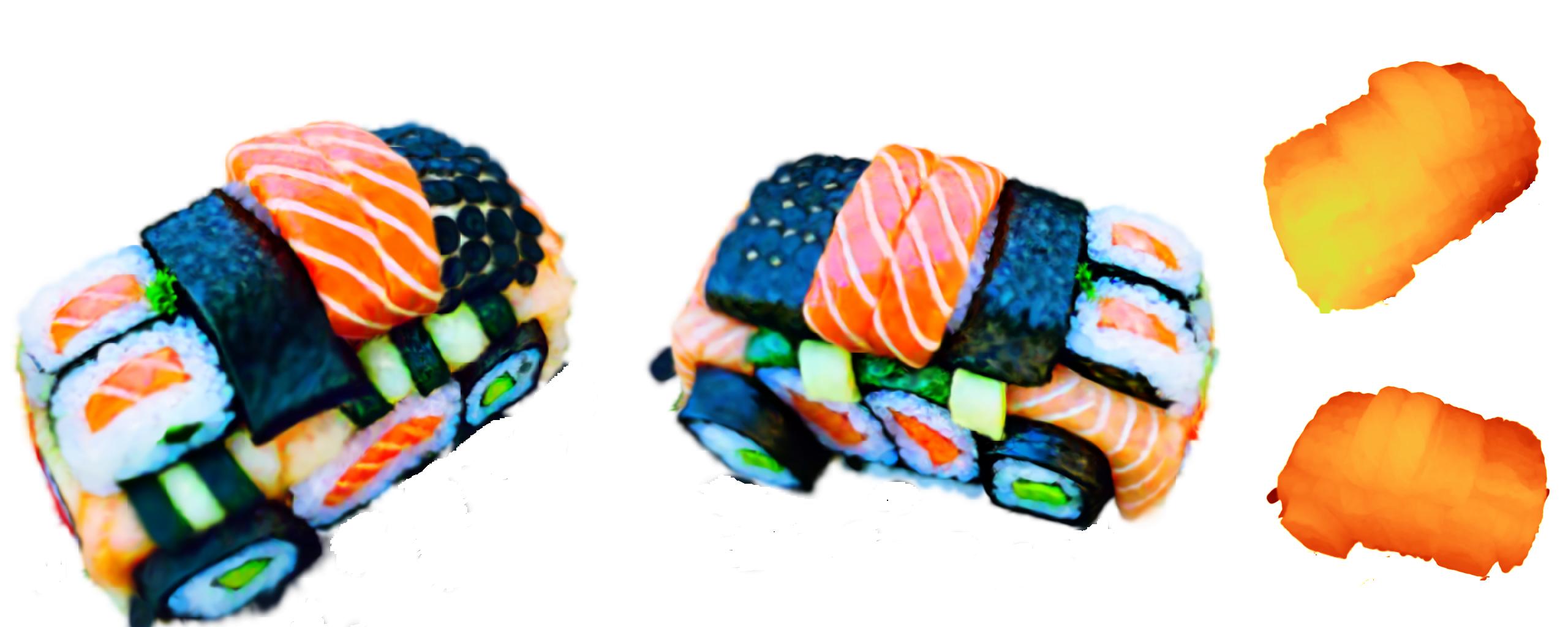} 
        \end{subfigure}
    \end{minipage}
    \\ \textit{\small A DSLR photo of car made out of sushi} \\

    \begin{minipage}{0.24\textwidth}
        \centering
        \begin{subfigure}{\linewidth}
            \centering
            \caption*{\textit{Magic3D}}
            \includegraphics[width=1.0\linewidth]{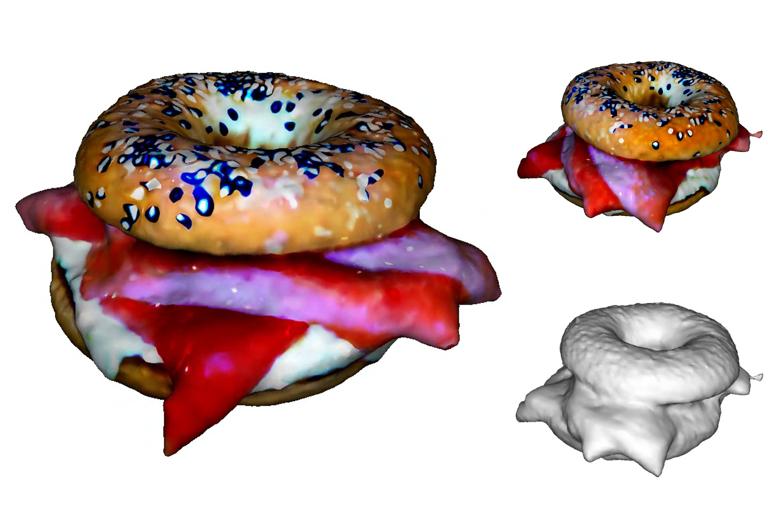} 
        \end{subfigure}
    \end{minipage}
    \hfill
    \begin{minipage}{0.24\textwidth}
        \centering
        \begin{subfigure}{\linewidth}
            \centering
            \caption*{\textit{GSGEN (Ours)}}
            \includegraphics[width=1.0\linewidth]{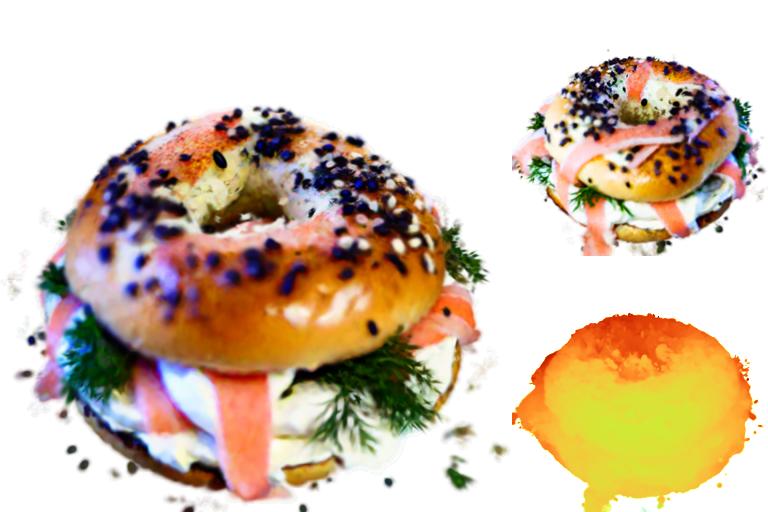} 
        \end{subfigure}
    \end{minipage}
    \hfill
    \begin{minipage}{0.24\textwidth}
        \centering
        \begin{subfigure}{\linewidth}
            \centering
            \caption*{\textit{ProlificDreamer}}
            \includegraphics[width=1.0\linewidth]{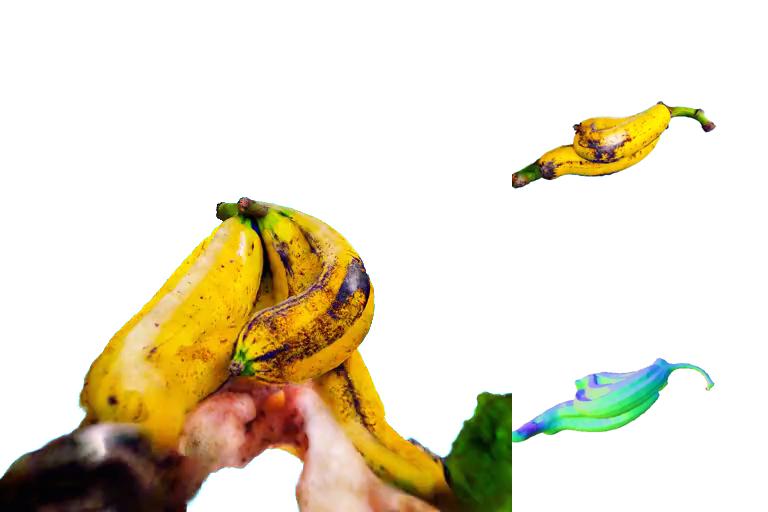} 
        \end{subfigure}
    \end{minipage}
    \hfill
    \begin{minipage}{0.24\textwidth}
        \centering
        \begin{subfigure}{\linewidth}
            \centering
            \caption*{\textit{GSGEN (Ours)}}
            \includegraphics[width=1.0\linewidth]{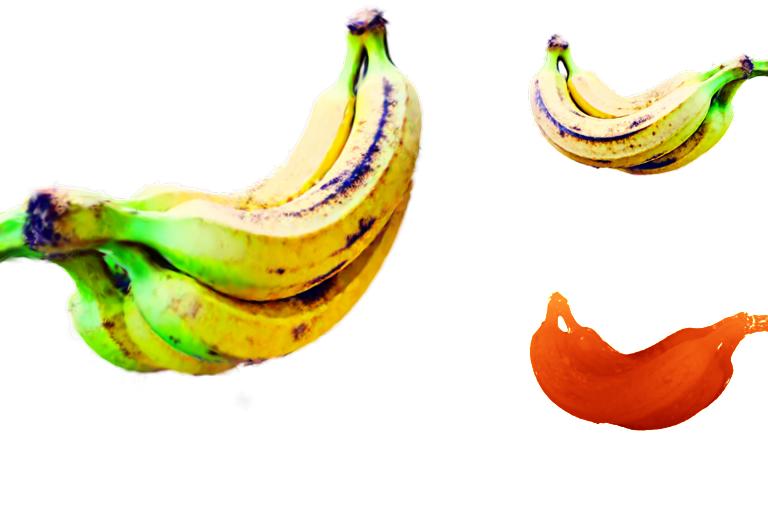} 
        \end{subfigure}
    \end{minipage}
    \begin{tabular}{>{\centering\arraybackslash}p{0.46\textwidth}>{\centering\arraybackslash}p{0.46\textwidth}}
        \textit{\small A bagel filled with cream cheese and lox} & \textit{\small A DSLR photo of banana} \\
    \end{tabular}

    \begin{minipage}{0.24\textwidth}
        \centering
        \begin{subfigure}{\linewidth}
            \centering
            \includegraphics[width=1.0\linewidth]{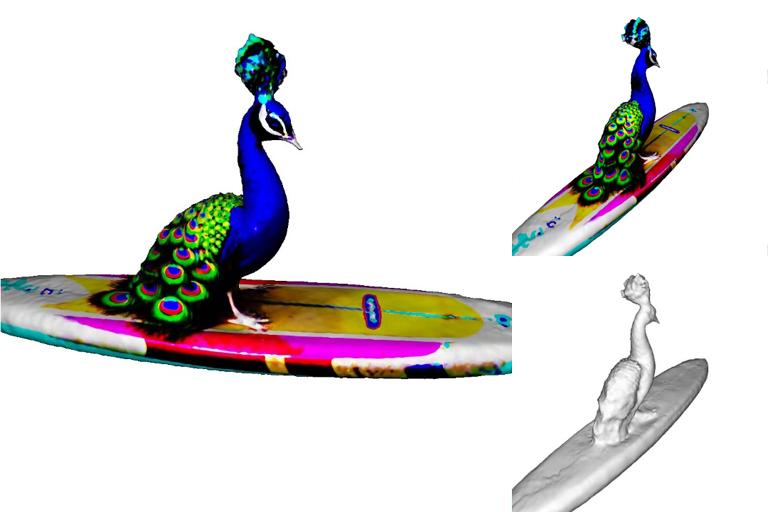} 
        \end{subfigure}
    \end{minipage}
    \hfill
    \begin{minipage}{0.24\textwidth}
        \centering
        \begin{subfigure}{\linewidth}
            \centering
            \includegraphics[width=1.0\linewidth]{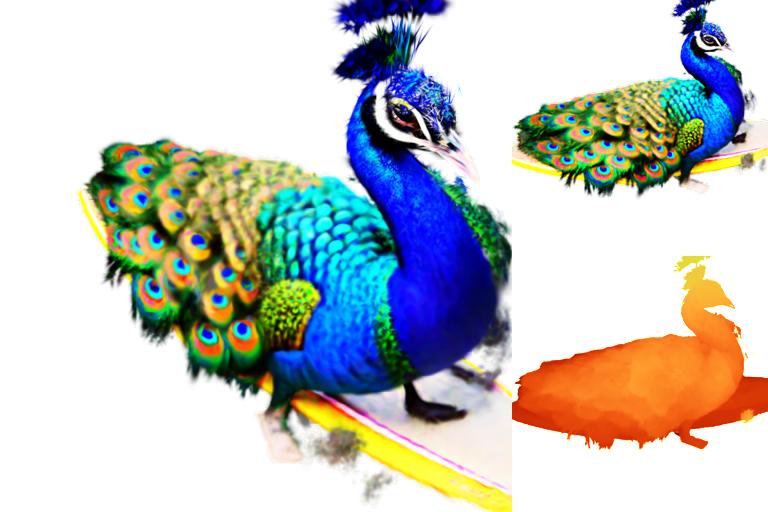} 
        \end{subfigure}
    \end{minipage}
    \hfill
    \begin{minipage}{0.24\textwidth}
        \centering
        \begin{subfigure}{\linewidth}
            \centering
            \includegraphics[width=1.0\linewidth]{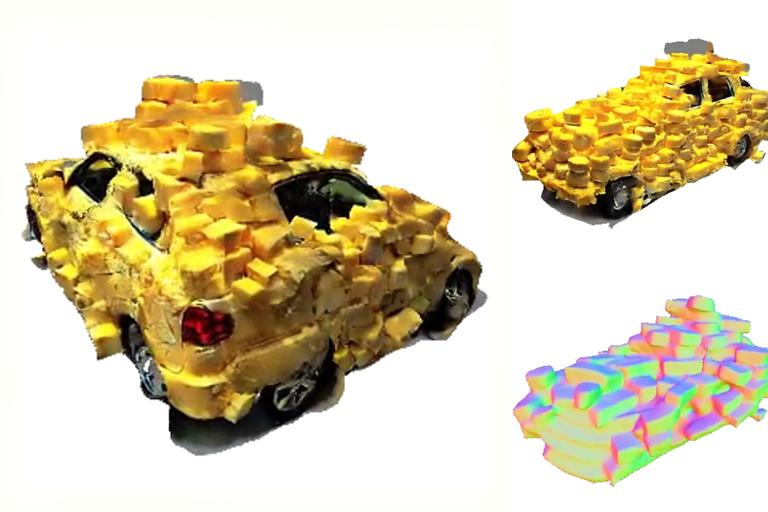} 
        \end{subfigure}
    \end{minipage}
    \hfill
    \begin{minipage}{0.24\textwidth}
        \centering
        \begin{subfigure}{\linewidth}
            \centering
            \includegraphics[width=1.0\linewidth]{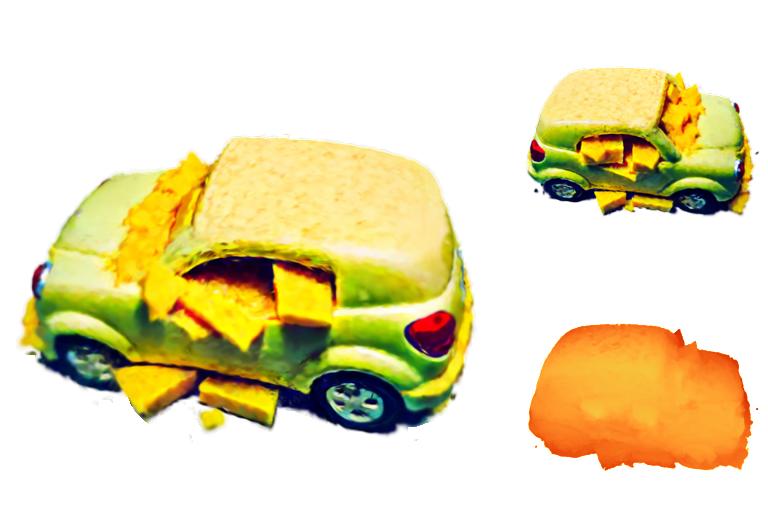} 
        \end{subfigure}
    \end{minipage}
    \begin{tabular}{>{\centering\arraybackslash}p{0.46\linewidth}>{\centering\arraybackslash}p{0.46\linewidth}}
        \textit{\small A peacock on a surfboard} & \textit{\small A car made out of cheese} \\
    \end{tabular}

    
    \vspace{-2mm}
    \caption{Qualitative comparison between the proposed \approach and state-of-the-art generation methods, including DreamFusion \citep{dreamfusion}, Magic3D \citep{magic3d}, Fantasia3D \citep{fantasia3d}, and ProlificDreamer~\citep{prolificdreamer}. For more qualitative comparison results, please refer to the appendix. Videos of these images are provided in the project page.}
    \label{fig:comparison}
    \vspace{-3mm}
\end{figure*}
To tackle this, we propose compactness-based densification as a supplement to positional gradient-based split with a large threshold. Specifically, for each Gaussian, we first obtain its K nearest neighbors with a KD-Tree. Then, for each of the neighbors, if the distance between the Gaussian and its neighbor is smaller than the sum of their radius, a Gaussian will be added between them with a radius equal to the residual. As illustrated in Fig.~\ref{fig:densification}, compactness-based densification could ``fill the holes", resulting in a more complete geometry structure. To prune unnecessary Gaussians, we add an extra loss to regularize opacity with a weight proportional to its distance to the center and remove Gaussians with opacity smaller than a threshold $\alpha_{min}$ periodically. Furthermore, we recognize the importance of ensuring the geometry consistency of the Gaussians throughout the refinement phase. With this concern, we penalize Gaussians which deviate significantly from their positions obtained during the preceding geometry optimization.
The loss function in the appearance refinement stage is summarized as the following:
\begin{equation}
\begin{split}
    &\nabla_\theta\mathcal{L}_{\text{refine}}=\lambda_{\text{SDS}}\mathbb{E}_{\epsilon_{I}, t}\left[w_I(t)(\epsilon_{\phi}(x_t;y,t)-\epsilon_I)\frac{\partial\mathbf{x}}{\partial\theta}\right] \\
    &+ \lambda_{\text{mean}}\nabla_\theta\sum_i||\mathbf{p}_i|| + \lambda_{\text{opacity}}\nabla_\theta\sum_i\mathtt{sg}(||\mathbf{p}_i||)\cdot o_i,
\end{split}
\end{equation}
where $\mathtt{sg}(\cdot)$ refers to the stop gradient operation, $\mathbf{p}_i$ and $o_i$ represents the position and opacity of the $i$-th Gaussian respectively. $\lambda_{\text{SDS}}$, $\lambda_{\text{mean}}$ and $\lambda_{\text{opacity}}$ are loss weights.

\begin{figure*}[ht]
    \vspace{-10mm}
    \centering
    \begin{minipage}{0.24\textwidth}
        \centering
        \begin{subfigure}{\linewidth}
            \centering
            \caption{\textit{w/o initialization}}
            \includegraphics[width=1.0\linewidth]{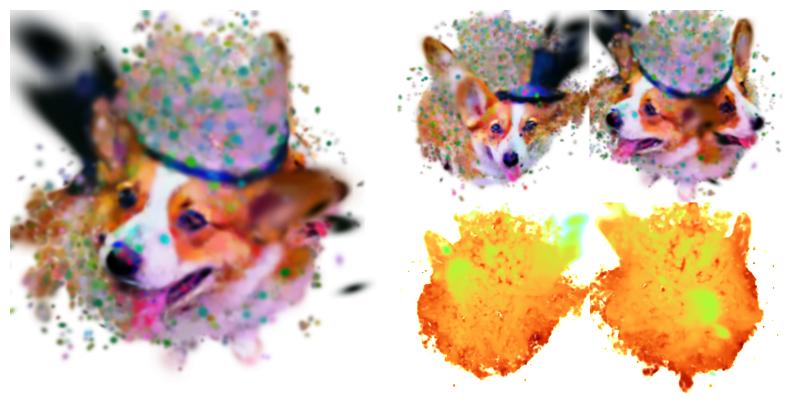}
            \label{fig:ablation_no_init}
        \end{subfigure}
    \end{minipage}
    \hfill
    \begin{minipage}{0.24\textwidth}
        \centering
        \begin{subfigure}{\linewidth}
            \centering
            \caption{\textit{w/o 3D guidance}}
            \includegraphics[width=1.0\linewidth]{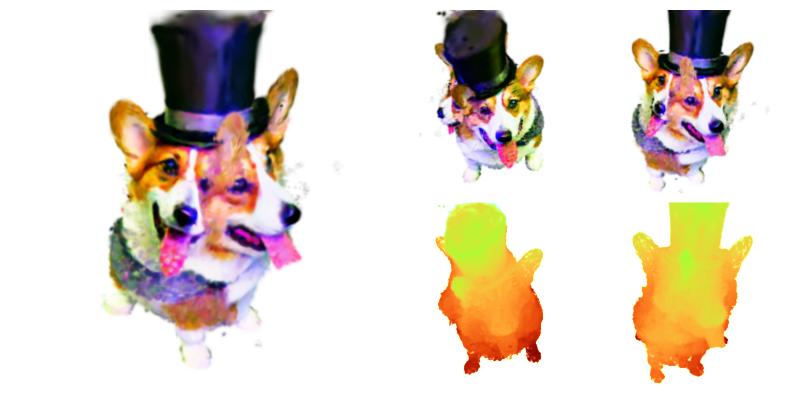}
            \label{fig:ablation_no_pointe}
        \end{subfigure}
    \end{minipage}
    \hfill
    \begin{minipage}{0.24\textwidth}
        \centering
        \begin{subfigure}{\linewidth}
            \centering
            \caption{\textit{Coarse Model}}
            \includegraphics[width=1.0\linewidth]{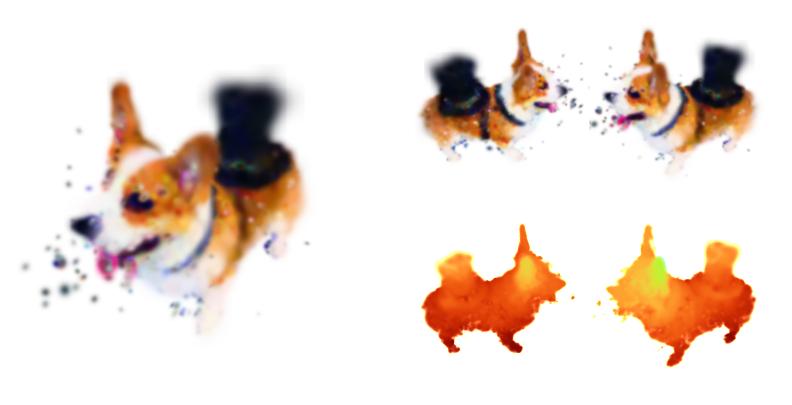}
            \label{fig:ablation_coarse}
        \end{subfigure}
    \end{minipage}
    \hfill
    \begin{minipage}{0.24\textwidth}
        \centering
        \begin{subfigure}{\linewidth}
            \centering
            \caption{\textit{Full}}
            \includegraphics[width=1.0\linewidth]{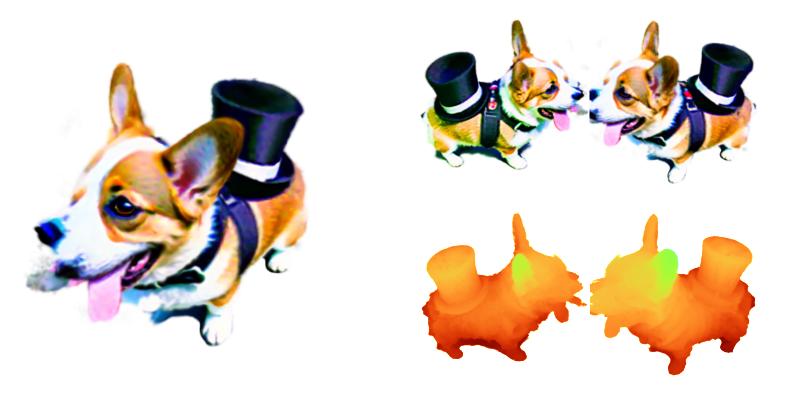}
            \label{fig:ablation_full}
        \end{subfigure}
    \end{minipage}
    \vspace{-5mm}
    \\ \textit{\smaller A zoomed out DSLR photo of a corgi wearing a top hat}

    \centering
    \begin{minipage}{0.24\textwidth}
        \centering
        \begin{subfigure}{\linewidth}
            \centering
            \includegraphics[width=1.0\linewidth]{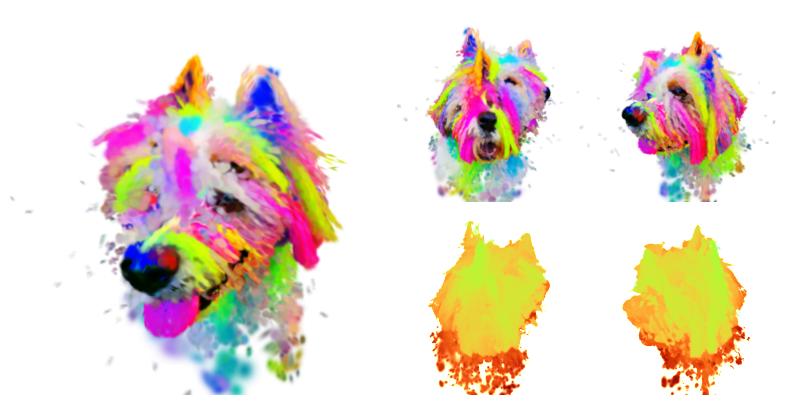}
        \end{subfigure}
    \end{minipage}
    \hfill
    \begin{minipage}{0.24\textwidth}
        \centering
        \begin{subfigure}{\linewidth}
            \centering
            \includegraphics[width=1.0\linewidth]{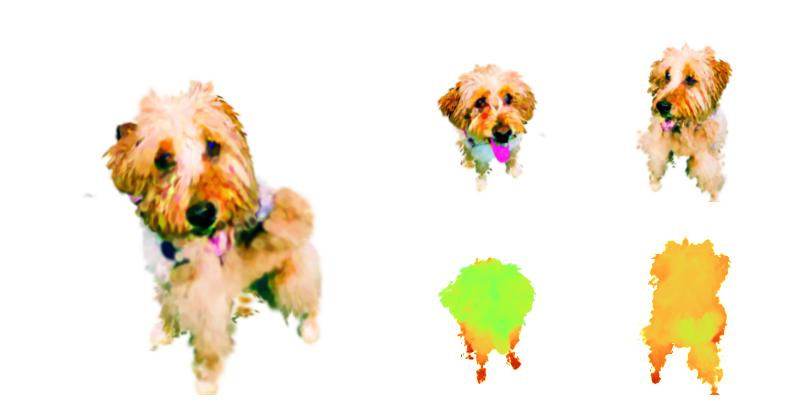}
        \end{subfigure}
    \end{minipage}
    \hfill
    \begin{minipage}{0.24\textwidth}
        \centering
        \begin{subfigure}{\linewidth}
            \centering
            \includegraphics[width=1.0\linewidth]{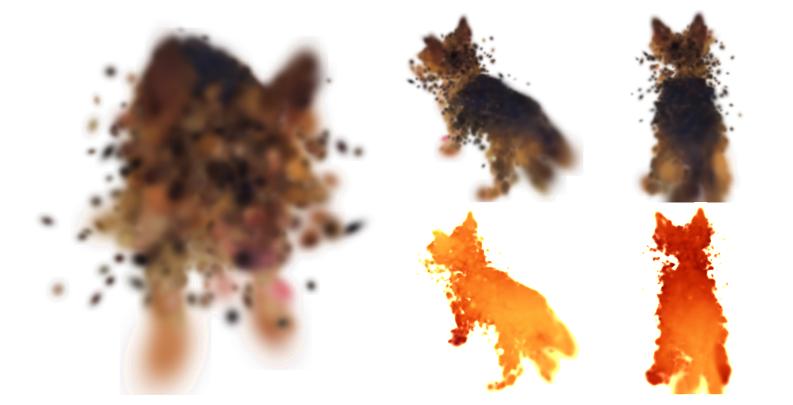}
        \end{subfigure}
    \end{minipage}
    \hfill
    \begin{minipage}{0.24\textwidth}
        \centering
        \begin{subfigure}{\linewidth}
            \centering
            \includegraphics[width=1.0\linewidth]{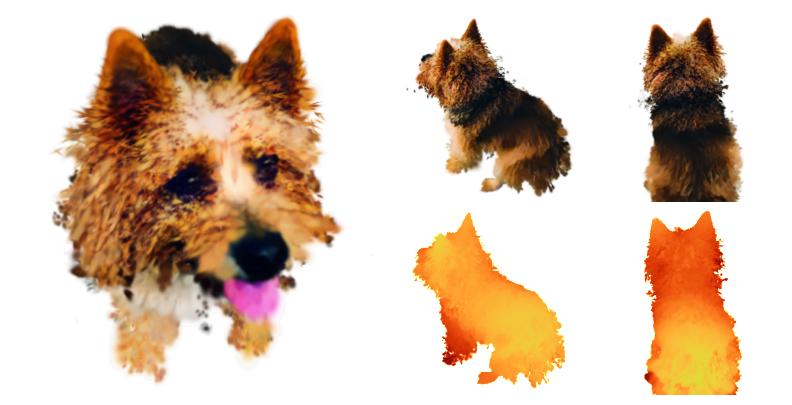}
        \end{subfigure}
    \end{minipage}
    \\ \textit{\smaller A high quality photo of a furry dog}

    \begin{minipage}{0.24\textwidth}
        \centering
        \begin{subfigure}{\linewidth}
            \centering
            \includegraphics[width=1.0\linewidth]{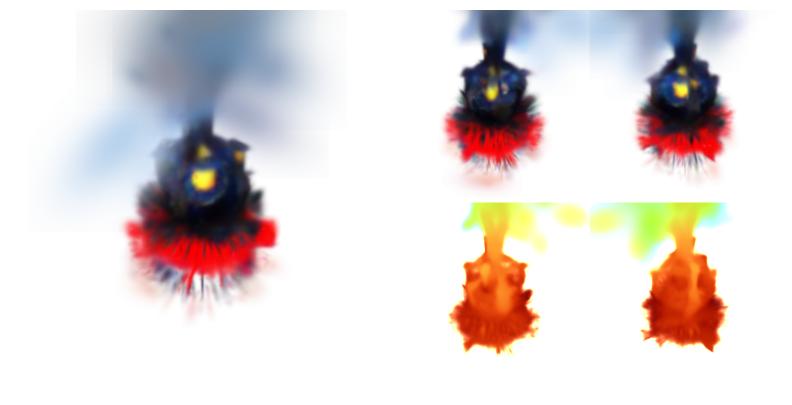}
        \end{subfigure}
    \end{minipage}
    \hfill
    \begin{minipage}{0.24\textwidth}
        \centering
        \begin{subfigure}{\linewidth}
            \centering
            \includegraphics[width=1.0\linewidth]{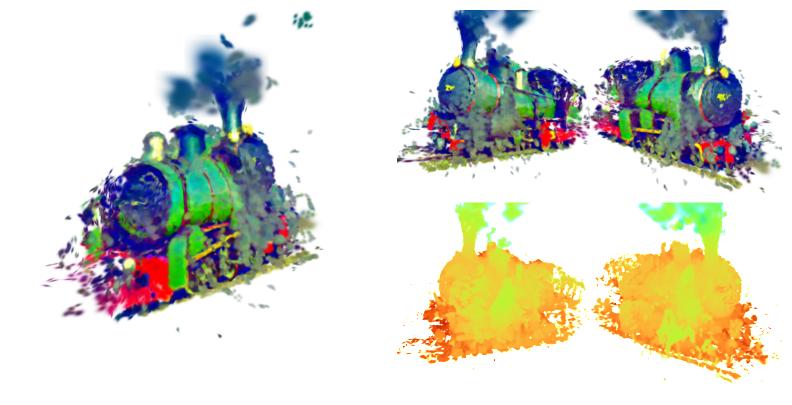}
        \end{subfigure}
    \end{minipage}
    \hfill
    \begin{minipage}{0.24\textwidth}
        \centering
        \begin{subfigure}{\linewidth}
            \centering
            \includegraphics[width=1.0\linewidth]{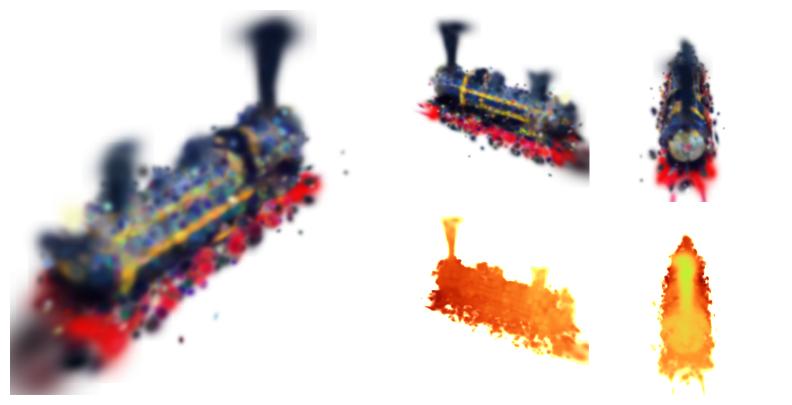}
        \end{subfigure}
    \end{minipage}
    \hfill
    \begin{minipage}{0.24\textwidth}
        \centering
        \begin{subfigure}{\linewidth}
            \centering
            \includegraphics[width=1.0\linewidth]{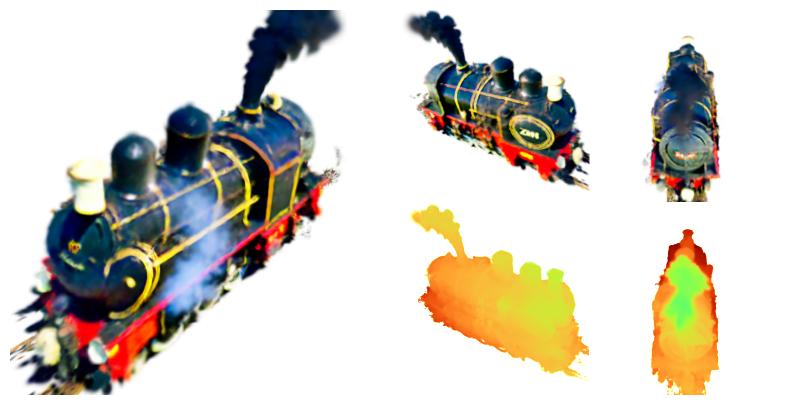}
        \end{subfigure}
    \end{minipage}

    \textit{\smaller A DSLR photo of a streaming engine train, high resolution}

    \begin{minipage}{0.24\textwidth}
        \centering
        \begin{subfigure}{\linewidth}
            \centering
            \includegraphics[width=1.0\linewidth]{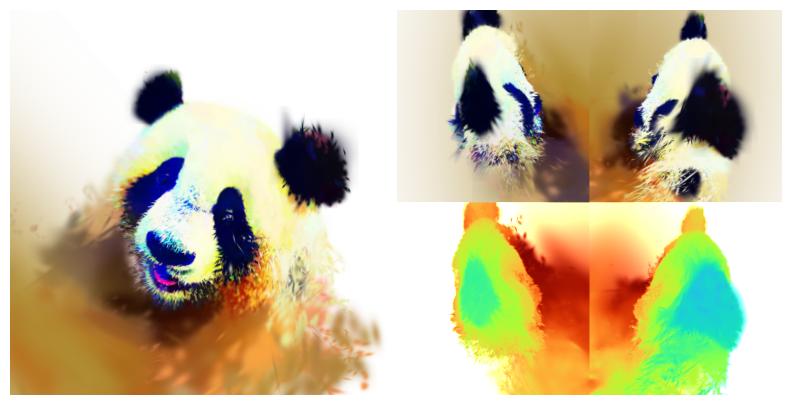}
        \end{subfigure}
    \end{minipage}
    \hfill
    \begin{minipage}{0.24\textwidth}
        \centering
        \begin{subfigure}{\linewidth}
            \centering
            \includegraphics[width=1.0\linewidth]{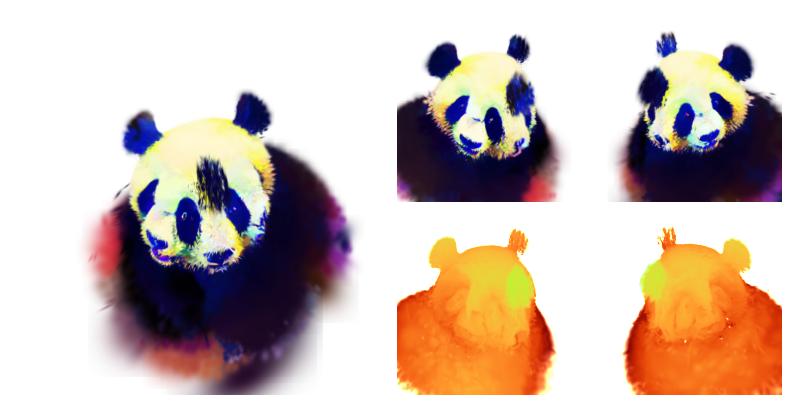}
        \end{subfigure}
    \end{minipage}
    \hfill
    \begin{minipage}{0.24\textwidth}
        \centering
        \begin{subfigure}{\linewidth}
            \centering
            \includegraphics[width=1.0\linewidth]{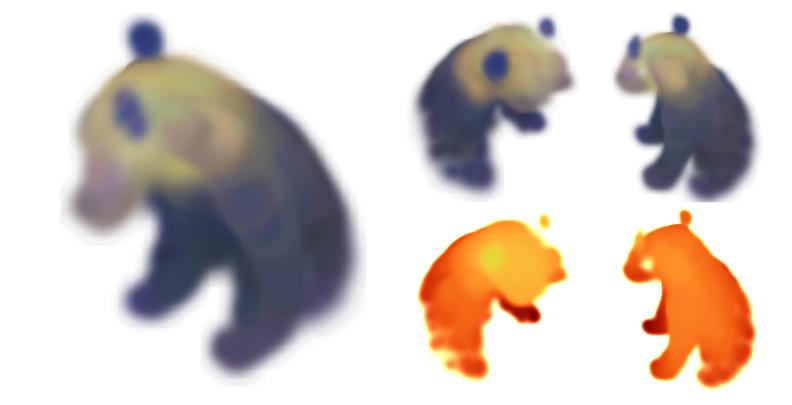}
        \end{subfigure}
    \end{minipage}
    \hfill
    \begin{minipage}{0.24\textwidth}
        \centering
        \begin{subfigure}{\linewidth}
            \centering
            \includegraphics[width=1.0\linewidth]{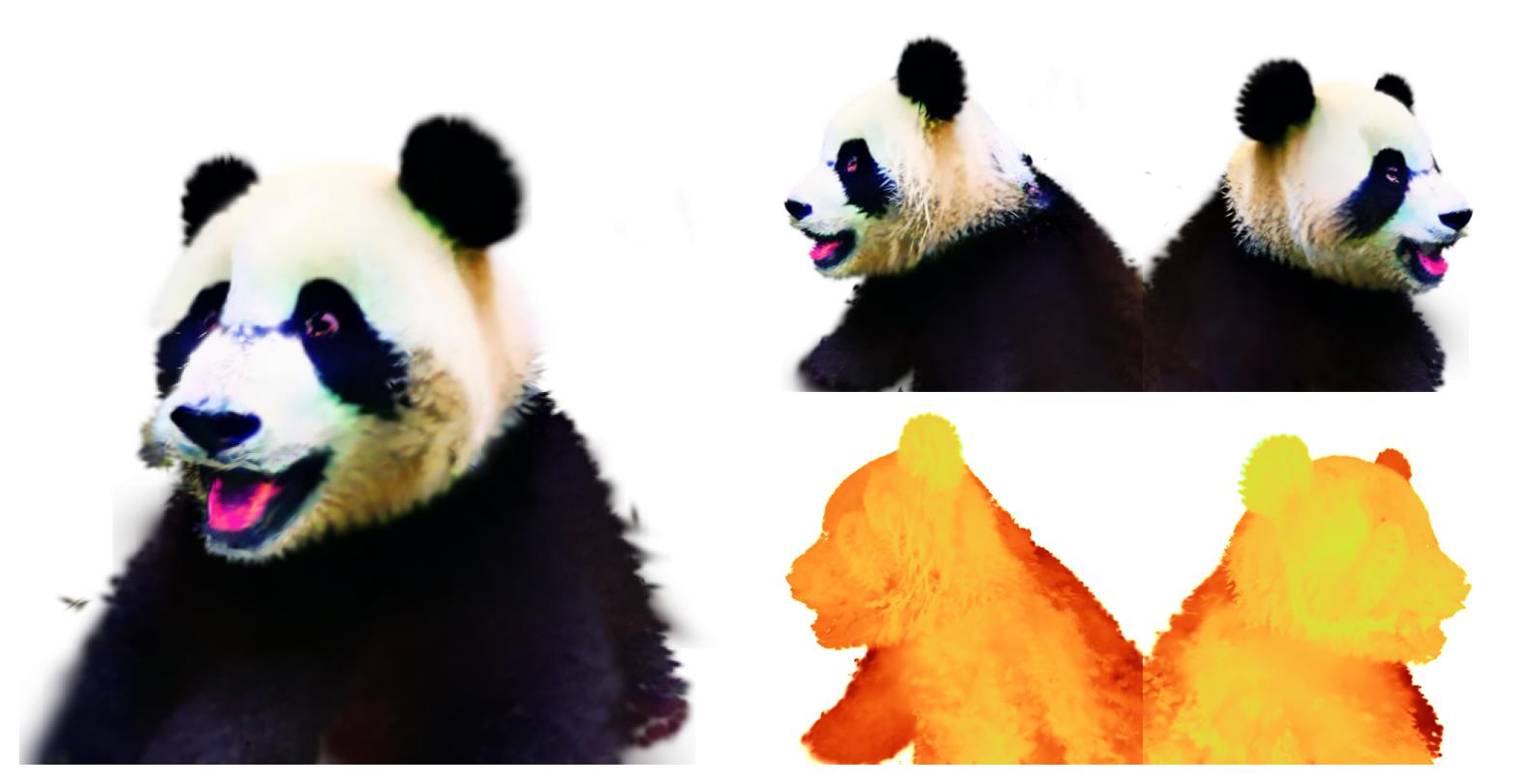}
        \end{subfigure}
    \end{minipage}

    \textit{\smaller A DSLR photo of a panda}
    \caption{Ablation study results on initialization and 3D prior.}
    \vspace{-3.8mm}
    \label{fig:ablation}
\end{figure*}
\subsection{Initialization with Geometry Prior}
\label{sec:initialize}
Previous studies \citep{fantasia3d, magic3d, latentnerf} have demonstrated the critical importance of starting with a reasonable geometry initialization. In our early experiments, we also found that initializing with a simple pattern could potentially lead to a degenerated 3D object. To overcome this, we opt for initializing the positions of the Gaussians either with a generated point cloud or with a 3D shape provided by the users (either a mesh or a point cloud).
In the context of general text-to-3D generation, we employ a text-to-point-cloud diffusion model, \textit{Point-E} \citep{pointe}, to generate a rough geometry according to the text prompt. While Point-E can produce colored point clouds, we opt for random color initialization based on empirical observations, as direct utilization of the generated colors has been found to have detrimental effects in early experiments 
(See the appendix for visualization). 
The scales and opacities of the Gaussians are assigned with fixed values, and the rotation matrix is set to the identity matrix. For user-guided generation, we convert the preferred shape to a point cloud. To avoid too many vertices in the provided shape, we use farthest point sampling \citep{fps} for point clouds and uniform surface sampling for meshes to extract a subset of the original shape instead of directly using all the vertices or points.
\section{Experiments}

In this section, we present our experiments on validating the effectiveness of the proposed approach. Specifically, we compare \textsc{Gsgen} with previous state-of-the-art methods in general text-to-3D generation. Additionally, we conduct several ablation studies to evaluate the importance of initialization, 3D guidance, and densification strategy. The detailed results are shown as follows.
\subsection{Implementation Details}
\textbf{Guidance model setup.}
We implement the guidance model based on the publicly available diffusion model, StableDiffusion \citep{stablediffusion, von-platen-etal-2022-diffusers}. For the guidance scale, we adopt 100 for \textit{StableDiffusion} as suggested in DreamFusion and other works. We also exploit the view-dependent prompt technique proposed by DreamFusion. All the assets demonstrated in this section are obtained with StableDiffusion checkpoint \textit{runwayml/stable-diffusion-v1-5}.
\\
\textbf{3D Gaussian Splatting setup.} We implement the 3D Gaussian Splatting rendering pipeline with a PyTorch CUDA extension, and further add learnable background support to facilitate our application. For densification, we split the Gaussians by view-space position gradient every 500 iterations with a threshold $T_{pos}=0.02$
and perform compactness-based densification every 1000 iterations which we empirically found effective for achieving a complete geometry. For pruning, we remove Gaussians with opacity lower than $\alpha_{min}=0.05$, and excessively large world-space or view-space radius every 200 iterations.
\\
\textbf{Traning setup.} We use the same focal length, elevation, and azimuth range as those of DreamFusion \citep{dreamfusion}. To sample more uniformly in the camera position, we employ a stratified sampling on azimuth. We choose the loss weight hyperparameters $\lambda_{\text{SDS}}=0.1$ and $\lambda_{\text{3D}}=0.01$ in geometry optimization stage, and $\lambda_{\text{SDS}}=0.1$, $\lambda_{\text{mean}}=1.0$ and $\lambda_{\text{opacity}}=100.0$ in appearance refinement.

\subsection{Text-to-3D Generation}
We evaluate the performance of the proposed \textsc{Gsgen} in the context of general text-to-3D generation and present qualitative comparison against state-of-the-art methods. As illustrated in Fig.~\ref{fig:janus}, our approach produces delicate 3D assets with more accurate geometry and intricate details. In contrast, previous state-of-the-art methods under SDS guidance \citep{stable-dreamfusion, dreamfusion, magic3d, threestudio2023, fantasia3d} struggle in generating collapsed geometry under the same guidance and prompt, which underscores the effectiveness of our approach. While the VSD guidance proposed by ProlificDreamer~\citep{prolificdreamer} significantly improves the appearance of generated assets, it is still susceptible to the Janus problem, resulting in flawed geometry. 
We present more qualitative comparison results in Fig.~\ref{fig:comparison}, where our approach showcases notable enhancements in preserving high-frequency details such as the intricate patterns on sushi, the feathers of the peacock, and the thatched roof. 
In contrast, Magic3D and Fantasia3D yield over-smoothed geometry due to the limitation of mesh-based methods while ProlificDreamer is prone to the multi-face problem, making the generated assets less realistic. 
Furthermore, our \approach stands out for its efficiency, generating 3D assets in about 40 minutes, on par with Magic3D and Fantasia3D, but with improved fidelity and richer details.
For more qualitative comparisons and the performance of \approach under more advanced guidance including MVDream~\citep{shi2023MVDream} and DeepFloyd IF~\citep{Alex2023deep}, please refer to the appendix.


    
\subsection{Ablation Study}
\textbf{Initialization.}
To assess the impact of initialization, we introduce a variant that initiates the positions of the Gaussians with an origin-centered Gaussian distribution which emulates the initialization adopted in DreamFusion \citep{dreamfusion}. The qualitative comparisons are shown in Fig.~\ref{fig:ablation_no_init}. It is evident that assets generated with DreamFusion-like initialization encounter severe degeneration issues, especially for prompts depicting asymmetric scenes, resulting in collapsed geometry. In contrast, Point-E initialization breaks the symmetry by providing an anisotropic geometry prior, leading to the creation of more 3D-consistent objects.
\\
\textbf{3D prior.}
We evaluate the necessity of incorporating 3D prior by generating assets without point cloud guidance during geometry optimization. The qualitative comparisons are visualized in Fig.~\ref{fig:ablation_no_pointe}. 
Although achieved better geometry consistency compared to random initialization, relying solely on image diffusion prior still suffers from the Janus problem, which is particularly evident in cases with asymmetric geometries, such as the dog and the panda. In contrast, our approach effectively addresses this issue with the introduction of 3D prior, rectifying potentially collapsed structures in the geometry optimization stage and resulting in a 3D-consistent rough shape. 
Notably, we show in the appendix that \approach maintains great performance even when Point-E behaves sub-optimally. We attribute this to direct 3D prior provided by Point-E assisting in geometrical consistency by correcting major shape deviations in the early stage, without the need to guide fine-grained geometric details. For a comprehensive analysis, please refer to the appendix.
\begin{figure}[h]
    \centering
    \begin{minipage}{0.15\textwidth}
        \centering
        \begin{subfigure}{\linewidth}
            \centering
            \includegraphics[width=1.0\linewidth]{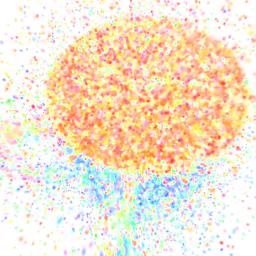} 
            \caption*{$T_{pos}=0.0002$}
        \end{subfigure}
    \end{minipage}
    \hfill
    \begin{minipage}{0.15\textwidth}
        \centering
        \begin{subfigure}{\linewidth}
            \centering
            \includegraphics[width=1.0\linewidth]{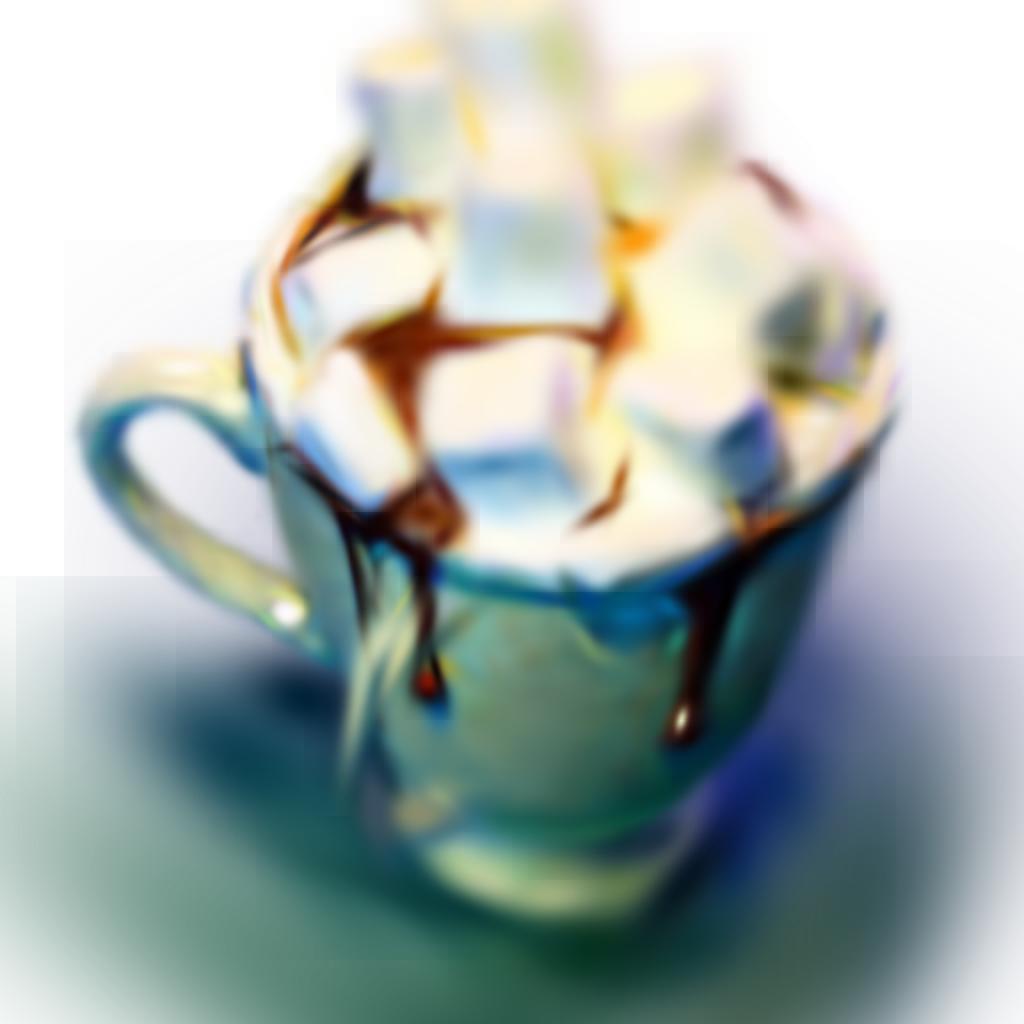} 
            \caption*{$T_{pos}=0.02$}
        \end{subfigure}
    \end{minipage}
    \hfill
    \begin{minipage}{0.15\textwidth}
        \centering
        \begin{subfigure}{\linewidth}
            \centering
            \includegraphics[width=1.0\linewidth]{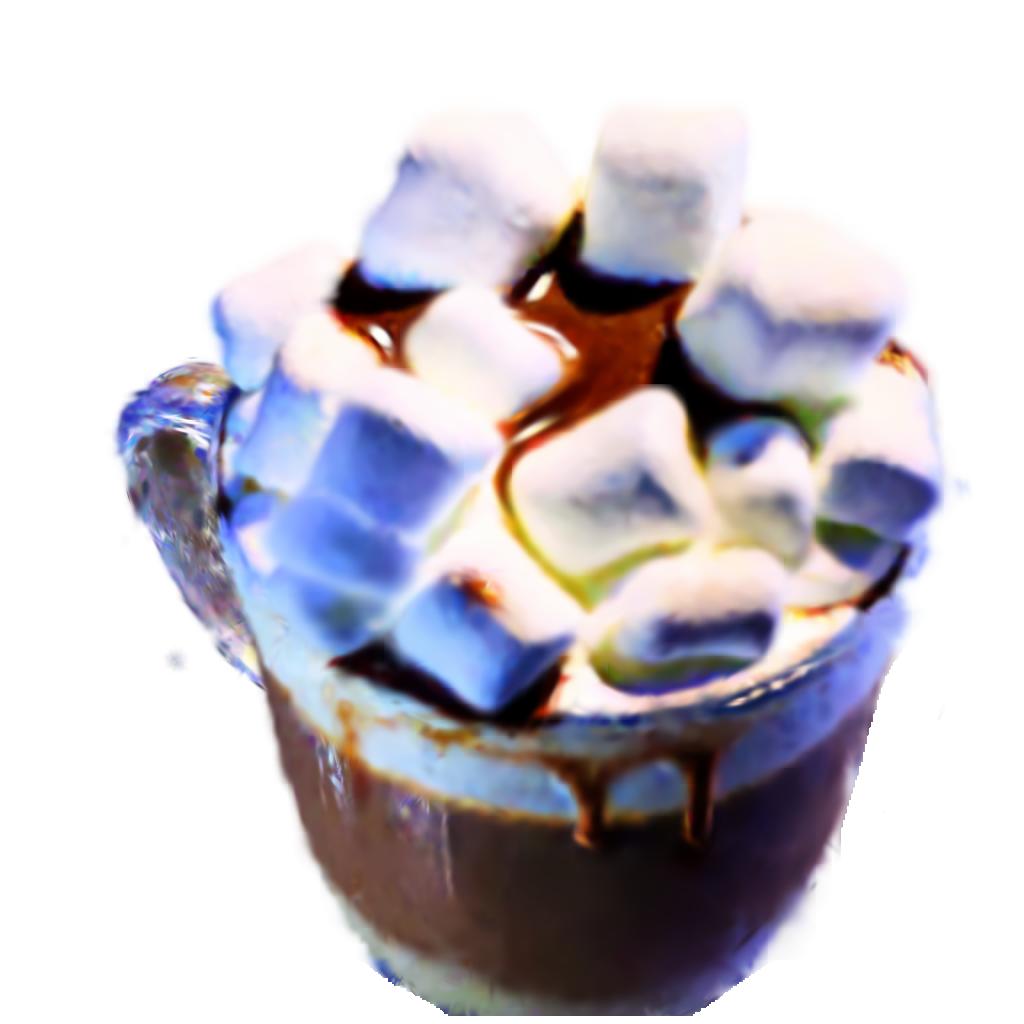} 
            \caption*{$T_{pos}=0.02$\\+compactness\centering}
        \end{subfigure}
    \end{minipage}
    \vspace{-3mm}
    \caption{Ablation study on densification strategy.}
    \vspace{-6mm}
    \label{fig:ablation_densification}
\end{figure}
\\
\textbf{Densification strategy.}
To validate the effectiveness of the proposed densification strategy, we propose two variants for comparison: (1) The original densification strategy that split Gaussians with an average view-space gradient larger than $T_{pos}=0.0002$. (2) With larger $T_{pos}=0.02$ that avoids too many new Gaussians. While effective in 3D reconstruction, the original densification strategy that relies only on view-space gradient encounters a dilemma in the context of score distillation sampling: within limited times of densification, a large threshold tends to generate an over-smoothed appearance while a small threshold is easily affected by unstable gradients. As shown in Fig.~\ref{fig:ablation_densification}, the proposed compactness-based densification is an effective supplement to the original densification strategy under SDS guidance.

\section{Limitations and Conclusion}
\textbf{Limitations.} 
    \approach tends to generate unsatisfying results when the provided text prompt contains a complex description of the scene or with complicated logic due to the limited language understanding ability of Point-E and the CLIP text encoder used in \textit{StableDiffusion}. Moreover, although incorporating 3D prior mitigates the Janus problem, it is far from eliminating the potential degenerations, especially when the textual prompt is extremely biased in the guidance diffusion models. Concrete failure cases and corresponding analyses are illustrated in the appendix.
\\
\textbf{Conclusion.}
In this paper, we propose \approach, a novel method for generating highly detailed and 3D consistent assets using Gaussian Splatting. In particular, we adopt a two-stage optimization strategy including geometry optimization and appearance refinement. In the geometry optimization stage, a rough shape is established under the joint guidance of a point cloud diffusion prior along with the common image SDS loss. In appearance refinement, the Gaussians are further optimized to enrich details and densified to achieve better continuity and fidelity with compactness-based densification. We conduct comprehensive experiments to validate the effectiveness of the proposed method, demonstrating its ability to generate 3D consistent assets and superior performance in capturing high-frequency components. We hope our method can serve as an efficient and powerful approach for high-quality text-to-3D generation and could pave the way for more extensive applications of Gaussians Splatting and direct incorporation of 3D prior.

\section{Acknowledgement}
This work was supported in part by the National Natural Science Fund for Distinguished Young Scholars under Grant 62025304.

{
    \small
    \bibliographystyle{ieeenat_fullname}
    \bibliography{main}
}

\clearpage
\appendix
\section{Implementation Details}
\textbf{3D Gaussian Splatting Details.}
Instead of directly using the official 3D Gaussian Splatting code provided by \citet{kerbl3Dgaussians}, we reimplement this algorithm by ourselves due to the need to support learnable MLP background. The official 3D Gaussian Splatting implementation propagates the gradients of the Gaussians in an inverse order, i.e., the Gaussians rendered last get gradient first. Our implementation follows a plenoxel \citep{yu2021plenoxels} style back propagation that calculates the gradient in the rendering order, which we found much easier to incorporate a per-pixel background.

The depth maps are rendered using the view-space depth of the centers of the Gaussians, which we claim is accurate enough due to the tiny scale of the Gaussians \citep{ewa}. Besides, we implement a z-variance renderer to support z-var loss proposed by \citep{hifa}. However, we found that z-var loss seems to have a limited impact on the generated 3D asset, mainly due to the sparsity of Gaussians naturally enforcing a relatively thin surface.
During rendering and optimizing, we follow the original 3D Gaussian Splatting to clamp the opacity of the Gaussians into $[0.004, 0.99]$ to ensure a stable gradient and prevent potential overflows or underflows.

\textbf{Guidance Details.}
All the guidance of 2D image diffusion models we used in this paper is provided by huggingface diffusers \citep{von-platen-etal-2022-diffusers}. For StableDiffusion guidance, we opt for the \textit{runwayml/stable-diffusion-v1-5} checkpoint for all the experiments conducted in this paper. We also test the performance of \approach under other checkpoints, including \textit{stabilityai/stable-diffusion-2-base} and \textit{stabilityai/stable-diffusion-2-1-base}, but no improvements are observed.

For Point-E diffusion model and its checkpoints, we directly adopted their official implementation.

\textbf{Training Details.}
All the assets we demonstrate in this paper and the supplemental video are trained on a single NVIDIA 3090 GPU with a batch size of 4 and take about 40 minutes to optimize for one prompt. The 3D assets we showcase in this paper and supplemental video are obtained under the same hyper-parameter setting since we found our parameters robust toward the input prompt. The number of Gaussians after densification is around $[1e^5, 1e^6]$.

\textbf{Open-Sourced Resources and Corresponding Licenses.}
We summarize open-sourced code and resources with corresponding licenses used in our experiments in the following table.
\begin{table}[!ht]
    \centering
    \label{tab:the_only_one}
    \caption{Open-sourced resources used in the experiment.}
    \begin{tabular}{ccc}
    \toprule[2pt]
        Resource & License \\ \hline
        \href{https://github.com/ashawkey/stable-dreamfusion}{Stable DreamFusion} \citep{stable-dreamfusion} & Apache License 2.0 \\
        \href{https://fantasia3d.github.io/}{Fantasia3D} \citep{fantasia3d} & Apache License 2.0 \\
        \href{https://github.com/threestudio-project/threestudio}{threestudio} \citep{threestudio2023} & Apache License 2.0 \\
        \href{https://github.com/Stability-AI/stablediffusion}{StableDiffusion} \citep{stablediffusion} & MIT License \\
        \href{https://www.deepfloyd.ai/}{DeepFloyd IF} \citep{Alex2023deep} & DeepFloyd IF License \\
        \href{https://github.com/huggingface/diffusers}{HuggingFace Diffusers} & Apache License 2.0\\
        \href{https://github.com/openai/point-e}{OpenAI Point-E}~\citep{pointe} & MIT License\\
        \href{https://github.com/salesforce/ULIP}{ULIP}~\citep{xue2022ulip, xue2023ulip2} & BSD 3-Clause License\\
    \bottomrule[2pt]
    \end{tabular}
\end{table}

We use Stable DreamFusion and threestudio to obtain the results of DreamFusion, Magic3D, and ProlificDreamer under StableDiffusion and on the prompts that are not included in their papers and project pages since the original implementation has not been open-sourced due to the usage of private diffusion models. The results of Fatansia3D are obtained by running their official implementation with their parameter setting for dog-like shapes.

\section{Additional Results}
\label{app:addtional_results}
\subsection{User-Guided Generation}

\begin{figure}[ht]
    \centering
    \includegraphics[width=0.48\textwidth]{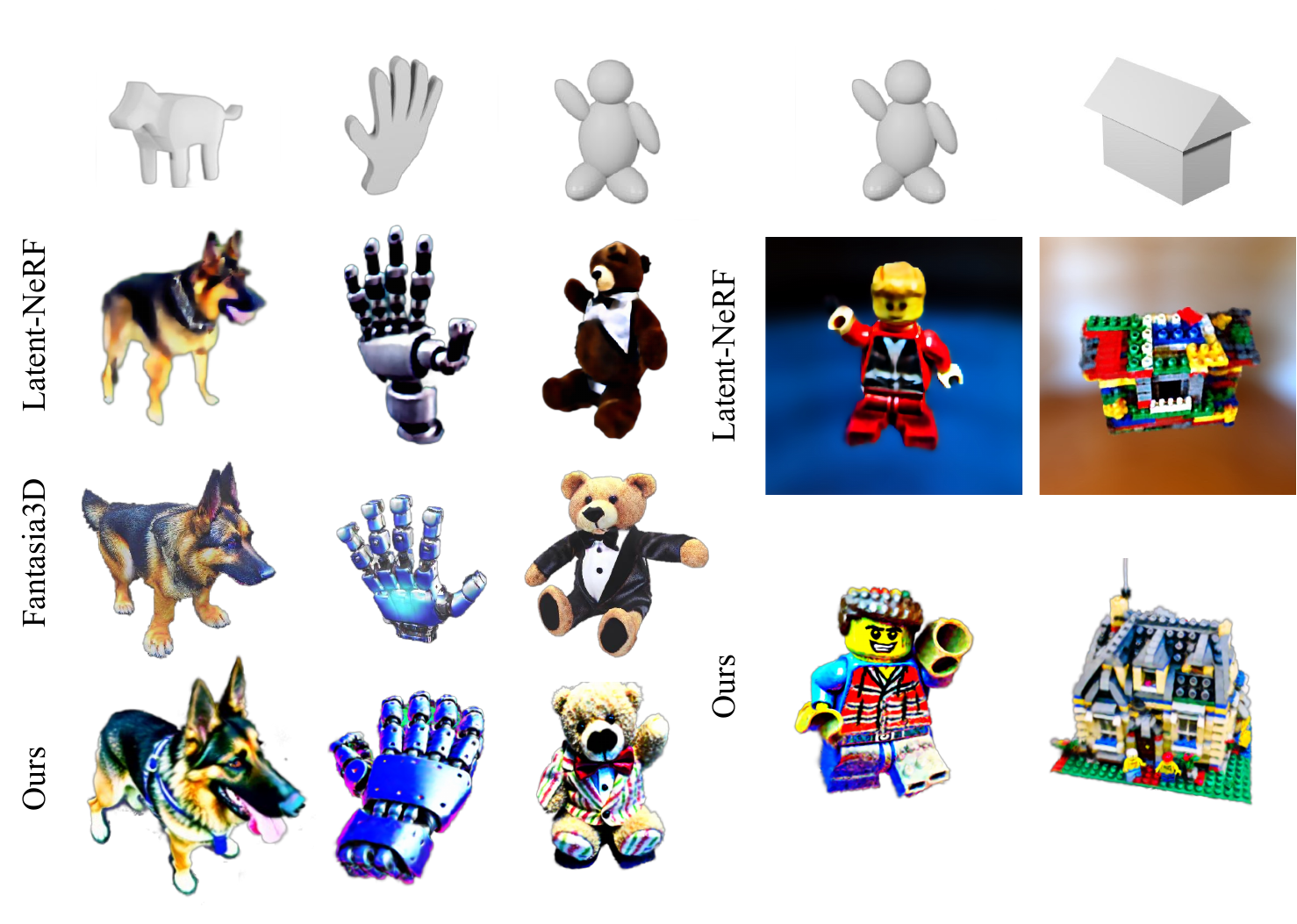}
    \caption{Qualitative comparison results on user-guided generation. The prompts from left to right are \textit{(1)~A German Shepherd; (2)~A robot hand, realistic; (3)~A teddy bear in a tuxedo; (4)~a lego man; (5)~a house made of lego}.}
    \label{fig:user_guided}
\end{figure}

Initialization is straightforward for 3D Gaussian Splatting due to its explicit nature, thereby automatically supporting user-guided generation. We evaluate the proposed \textsc{Gsgen} on user-guided generation with shapes provided in Latent-NeRF \citep{latentnerf}. In this experiment, the initial points are generated by uniformly sampling points on the mesh surface. To better preserve the user's desired shape, we opt for a relatively small learning rate for positions. We compare the 3D content generated by \approach with that generated by the state-of-the-art user-guided generation methods, Latent-NeRF \citep{latentnerf} and Fantasia3D \citep{fantasia3d}, as shown in Fig.~\ref{fig:user_guided}. Our proposed \textsc{Gsgen} achieves the best results among all alternatives in both geometry and textures and mostly keeps the geometrical prior given by the users. 

\subsection{More Text-To-3D Results}
\label{app:more_results}
\begin{figure*}[t]
    \vspace{-8mm}
    \centering
    \begin{minipage}{0.45\textwidth}
        \centering
        \begin{subfigure}{\linewidth}
            \centering
            \includegraphics[width=1.0\linewidth]{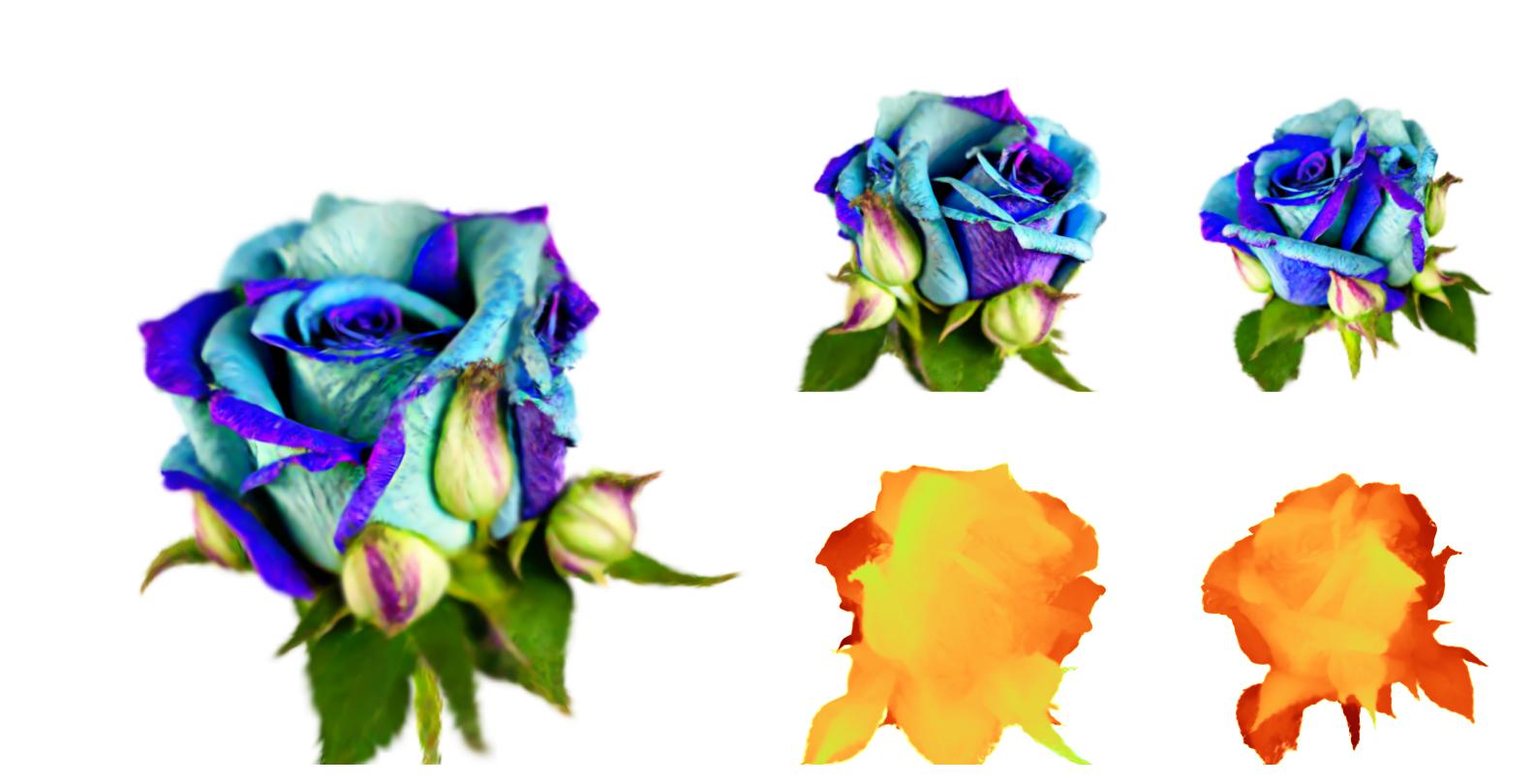} 
            \caption*{\textit{A bunch of blue rose, highly detailed}}
        \end{subfigure}
    \end{minipage}
    \begin{minipage}{0.45\textwidth}
        \centering
        \begin{subfigure}{\linewidth}
            \centering
            \includegraphics[width=1.0\linewidth]{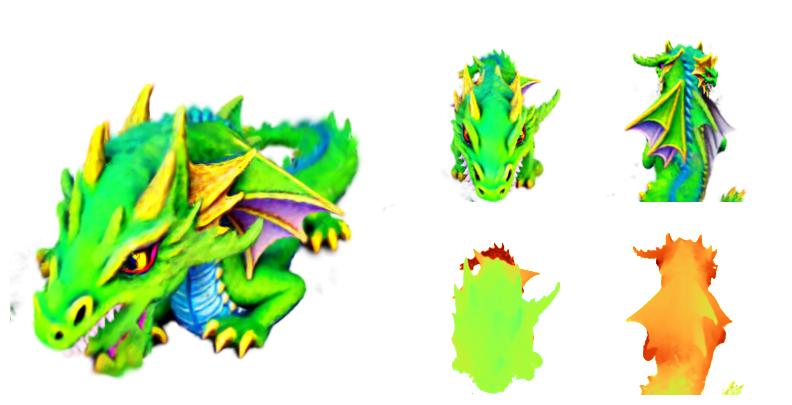} 
            \caption*{\textit{A high quality photo of a dragon}}
        \end{subfigure}
    \end{minipage}

    \begin{minipage}{0.45\textwidth}
        \centering
        \begin{subfigure}{\linewidth}
            \centering
            \includegraphics[width=1.0\linewidth]{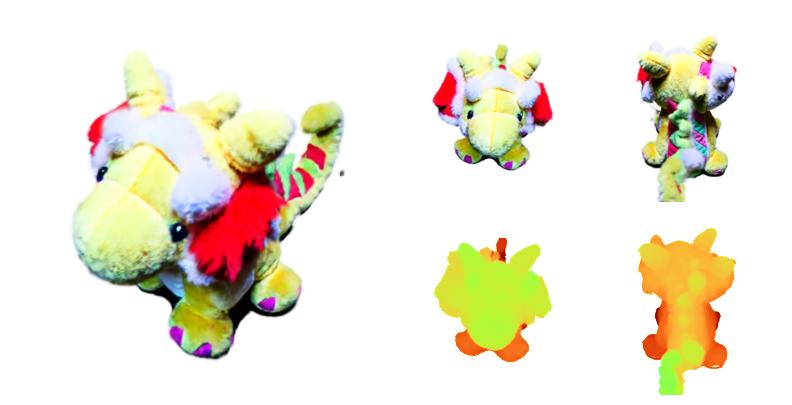} 
            \caption*{\textit{A plush dragon toy}}
        \end{subfigure}
    \end{minipage}
    \begin{minipage}{0.45\textwidth}
        \centering
        \begin{subfigure}{\linewidth}
            \centering
            \includegraphics[width=1.0\linewidth]{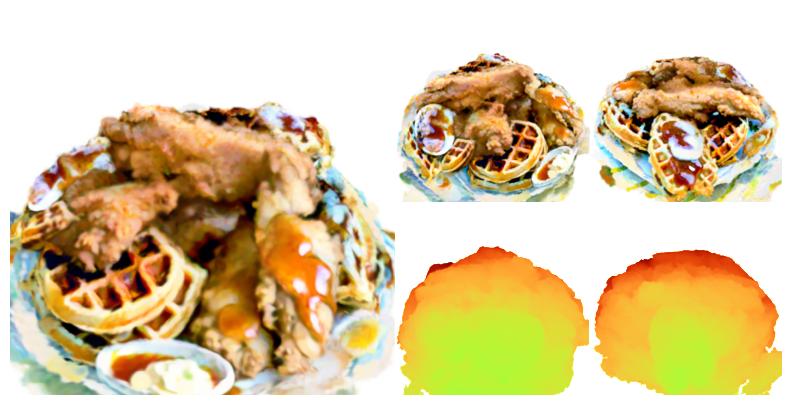} 
            \caption*{\textit{A zoomed out DSLR photo of a plate of fried chicken and waffles with maple syrup on them}}
        \end{subfigure}
    \end{minipage}

    \begin{minipage}{0.45\textwidth}
        \centering
        \begin{subfigure}{\linewidth}
            \centering
            \includegraphics[width=1.0\linewidth]{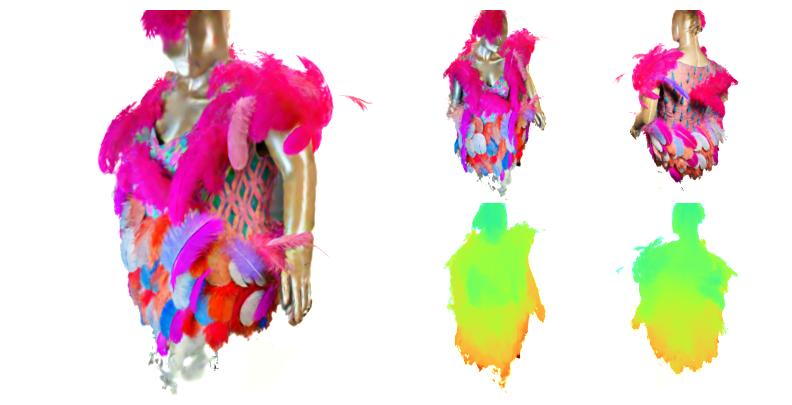} 
            \caption*{\textit{A beautiful dress made of feathers, on a mannequin}}
        \end{subfigure}
    \end{minipage}
    \begin{minipage}{0.45\textwidth}
        \centering
        \begin{subfigure}{\linewidth}
            \centering
            \includegraphics[width=1.0\linewidth]{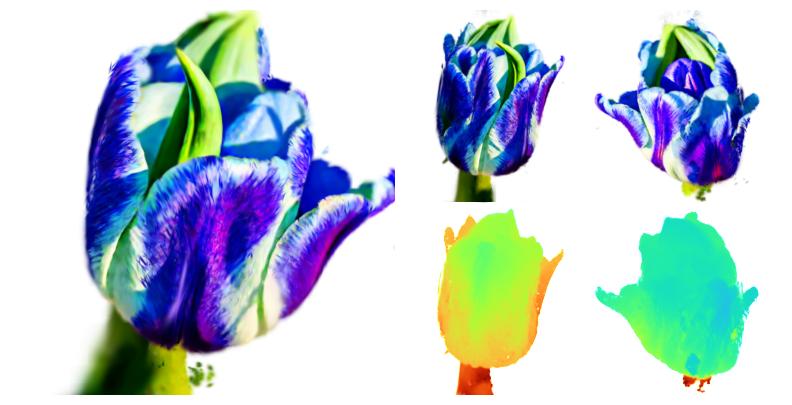} 
            \caption*{\textit{A high quality photo of a blue tulip}}
        \end{subfigure}
    \end{minipage}

    \begin{minipage}{0.45\textwidth}
        \centering
        \begin{subfigure}{\linewidth}
            \centering
            \includegraphics[width=1.0\linewidth]{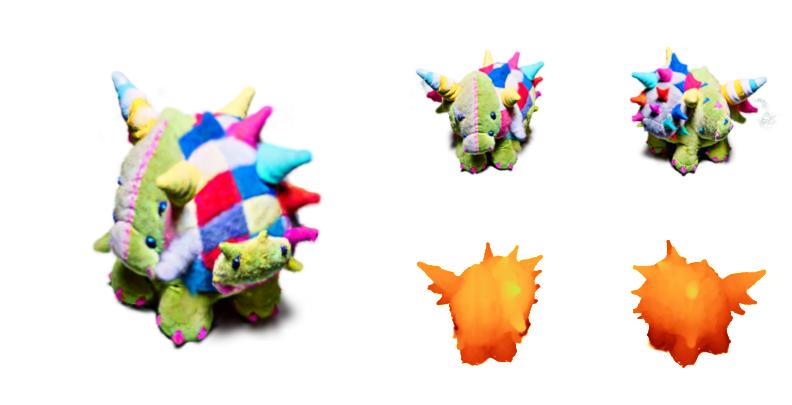} 
            \caption*{\textit{A DSLR photo of a plush triceratops toy, studio lighting, hight resolution}}
        \end{subfigure}
    \end{minipage}
    \begin{minipage}{0.45\textwidth}
        \centering
        \begin{subfigure}{\linewidth}
            \centering
            \includegraphics[width=1.0\linewidth]{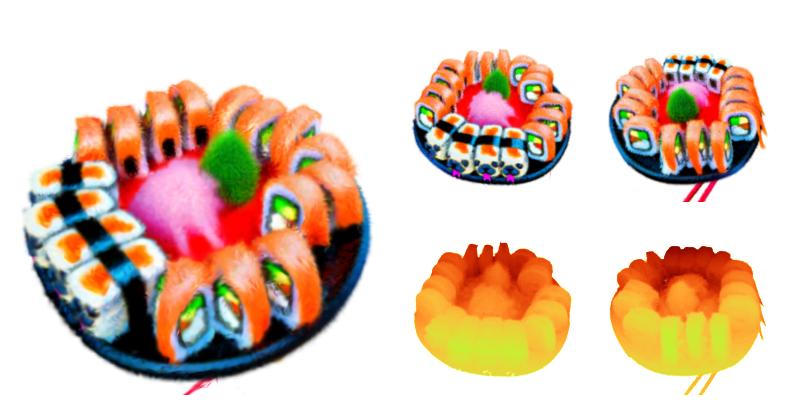} 
            \caption*{\textit{A DSLR photo of a tray of Sushi containing pugs}}
        \end{subfigure}
    \end{minipage}

    \begin{minipage}{0.45\textwidth}
        \centering
        \begin{subfigure}{\linewidth}
            \centering
            \includegraphics[width=1.0\linewidth]{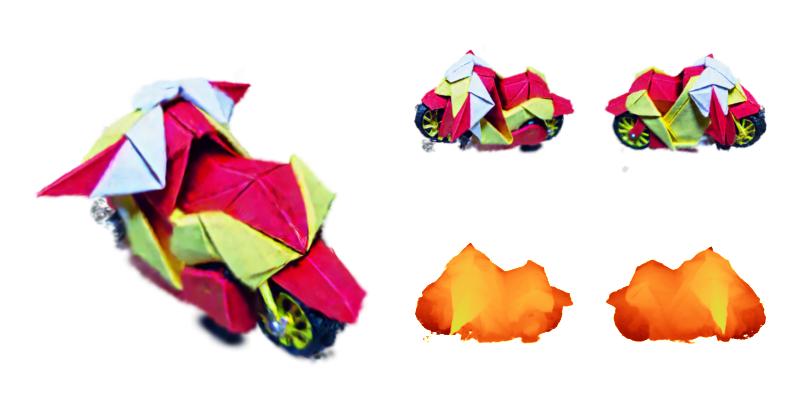} 
            \caption*{\textit{A DSLR photo of an origami motorcycle}}
        \end{subfigure}
    \end{minipage}
    \begin{minipage}{0.45\textwidth}
        \centering
        \begin{subfigure}{\linewidth}
            \centering
            \includegraphics[width=1.0\linewidth]{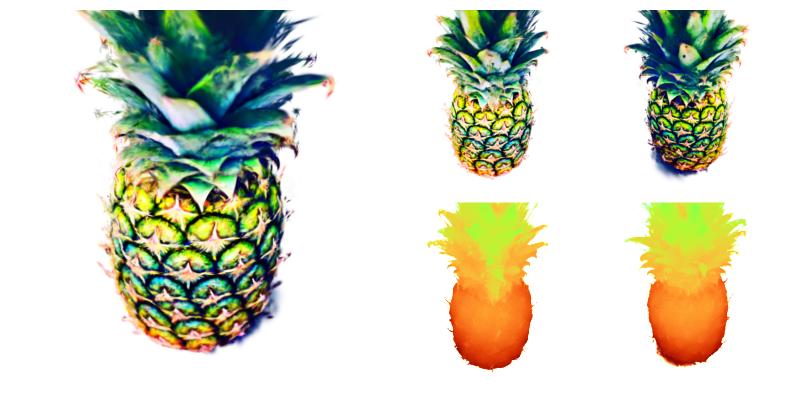} 
            \caption*{\textit{A DSLR photo of a pineapple}}
        \end{subfigure}
    \end{minipage}

    \caption{More 3D assets generated with \approach.}
    \label{fig:more_results1}
\end{figure*}

\begin{figure*}[t]
    \vspace{-1cm}
    \centering
    \begin{minipage}{0.45\textwidth}
        \centering
        \begin{subfigure}{\linewidth}
            \centering
            \includegraphics[width=1.0\linewidth]{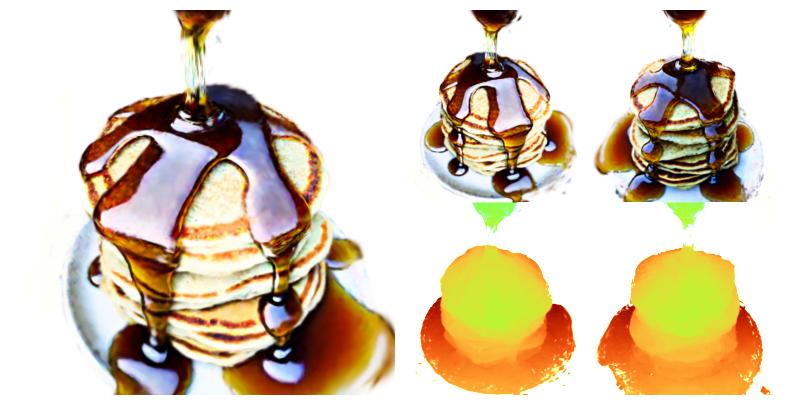} 
            \caption*{\textit{A high quality photo of a stack of pancakes covered in maple syrup}}
        \end{subfigure}
    \end{minipage}
    \begin{minipage}{0.45\textwidth}
        \centering
        \begin{subfigure}{\linewidth}
            \centering
            \includegraphics[width=1.0\linewidth]{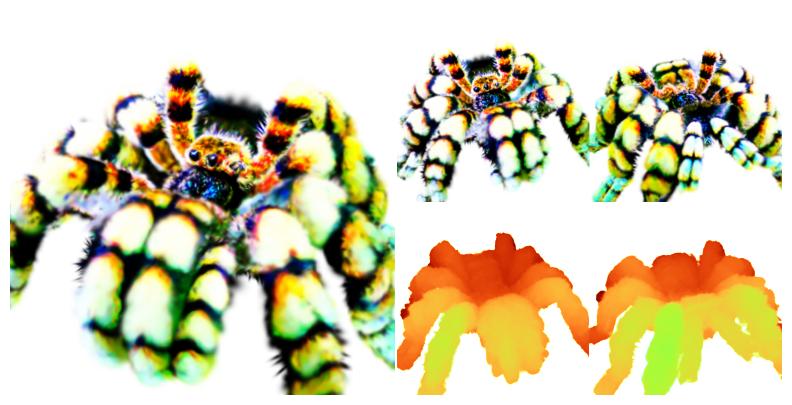} 
            \caption*{\textit{A tarantula, highly detailed}}
        \end{subfigure}
    \end{minipage}

    \begin{minipage}{0.45\textwidth}
        \centering
        \begin{subfigure}{\linewidth}
            \centering
            \includegraphics[width=1.0\linewidth]{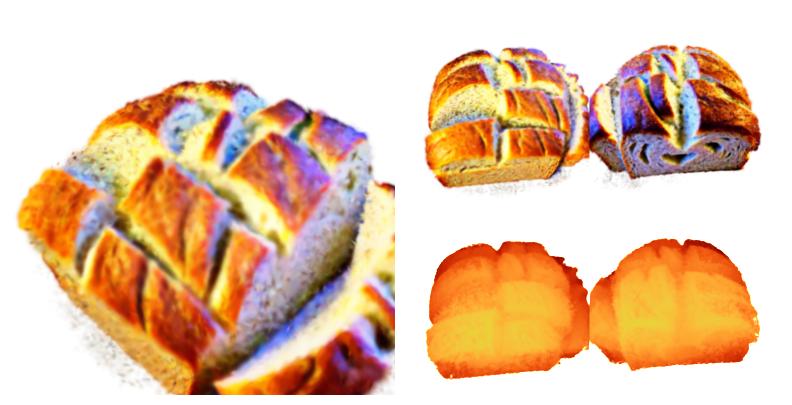} 
            \caption*{\textit{A sliced loaf of fresh bread}}
        \end{subfigure}
    \end{minipage}
    \begin{minipage}{0.45\textwidth}
        \centering
        \begin{subfigure}{\linewidth}
            \centering
            \includegraphics[width=1.0\linewidth]{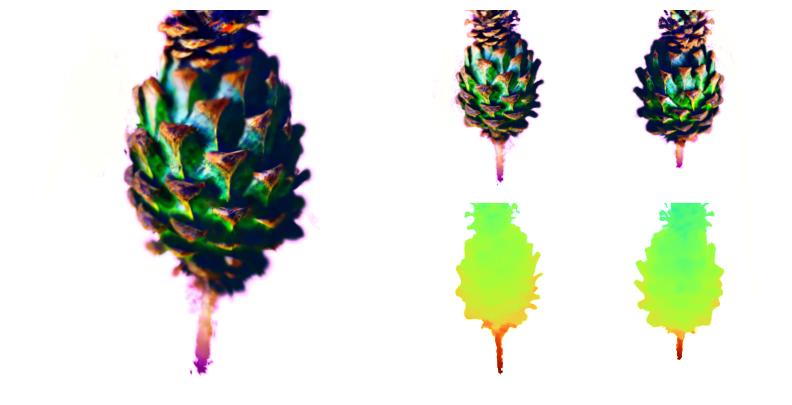} 
            \caption*{\textit{A high quality photo of a pinecone}}
        \end{subfigure}
    \end{minipage}

    \begin{minipage}{0.45\textwidth}
        \centering
        \begin{subfigure}{\linewidth}
            \centering
            \includegraphics[width=1.0\linewidth]{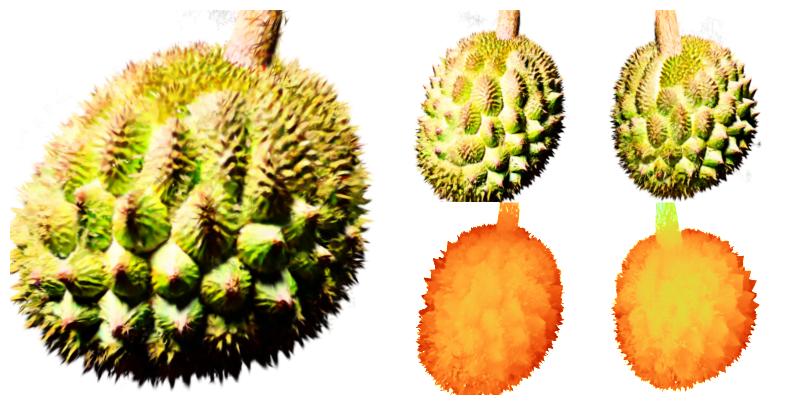} 
            \caption*{\textit{A high quality photo of a durian}}
        \end{subfigure}
    \end{minipage}
    \begin{minipage}{0.45\textwidth}
        \centering
        \begin{subfigure}{\linewidth}
            \centering
            \includegraphics[width=1.0\linewidth]{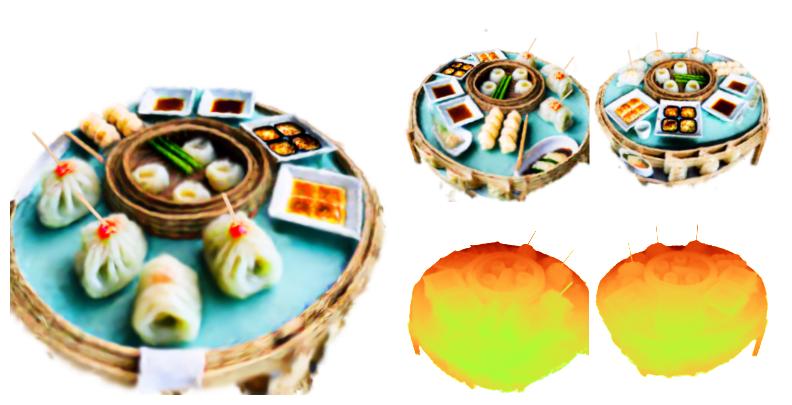} 
            \caption*{\textit{A zoomed out DSLR photo of a table with dim sum on it}}
        \end{subfigure}
    \end{minipage}

    \begin{minipage}{0.45\textwidth}
        \centering
        \begin{subfigure}{\linewidth}
            \centering
            \includegraphics[width=1.0\linewidth]{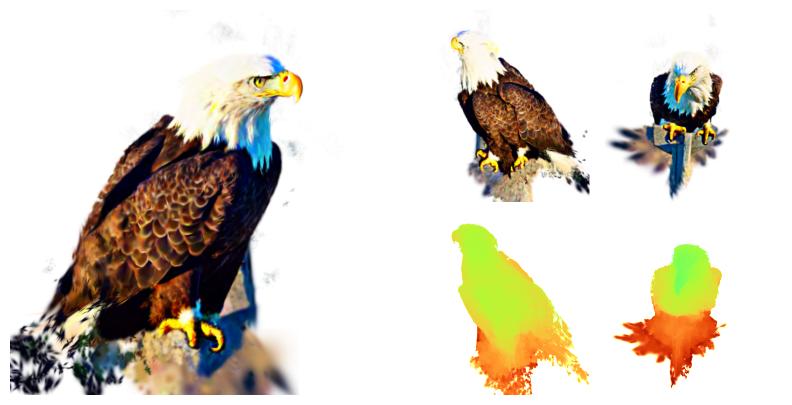} 
            \caption*{\textit{A DSLR photo of a bald eagle}}
        \end{subfigure}
    \end{minipage}
    \begin{minipage}{0.45\textwidth}
        \centering
        \begin{subfigure}{\linewidth}
            \centering
            \includegraphics[width=1.0\linewidth]{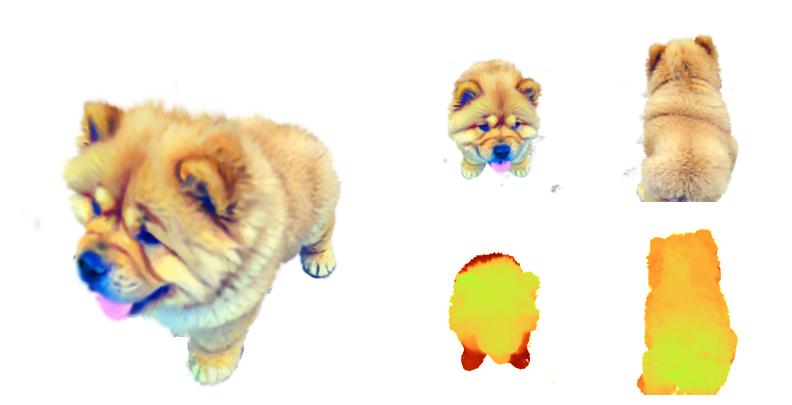} 
            \caption*{\textit{A high quality photo of a chow chow puppy}}
        \end{subfigure}
    \end{minipage}

    \begin{minipage}{0.45\textwidth}
        \centering
        \begin{subfigure}{\linewidth}
            \centering
            \includegraphics[width=1.0\linewidth]{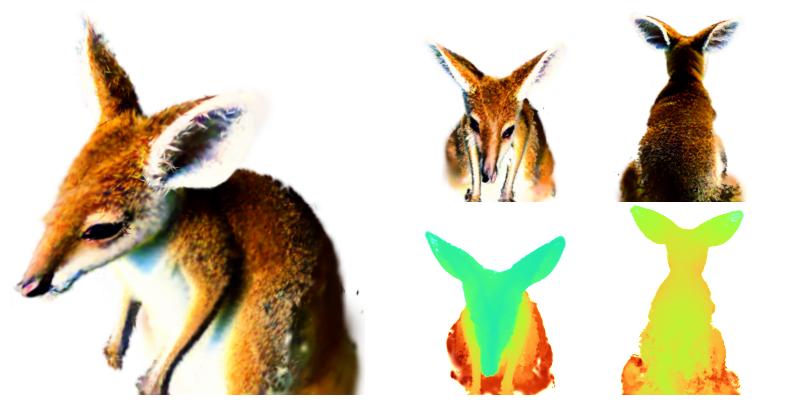} 
            \caption*{\textit{A high quality photo of a kangaroo}}
        \end{subfigure}
    \end{minipage}
    \begin{minipage}{0.45\textwidth}
        \centering
        \begin{subfigure}{\linewidth}
            \centering
            \includegraphics[width=1.0\linewidth]{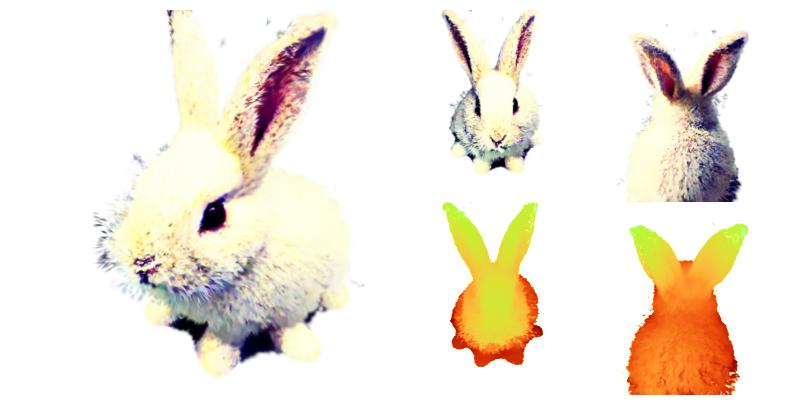} 
            \caption*{\textit{A high quality photo of a furry rabbit}}
        \end{subfigure}
    \end{minipage}

    \vspace{-3mm}
    \caption{More 3D assets generated with \approach.}
    \label{fig:more_results2}
\end{figure*}

We present more general text-to-3D generation results of \approach in Fig.~\ref{fig:more_results1} and Fig.~\ref{fig:more_results2}. Our approach can generate 3D assets with accurate geometry and improved fidelity.

For more delicate assets generated with \approach and the corresponding videos, please watch our supplemental video.

\subsection{More Qualitative Comparisons}
\label{app:comparison}
\begin{figure*}[t]
    \vspace{-8mm}
    \centering
    \begin{minipage}{0.43\textwidth}
        \centering
        \begin{subfigure}{\linewidth}
            \centering
            \includegraphics[width=1.0\linewidth]{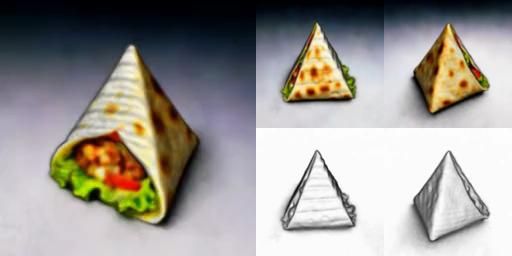} 
            \caption*{\textit{DreamFusion}}
        \end{subfigure}
    \end{minipage}
    \begin{minipage}{0.43\textwidth}
        \centering
        \begin{subfigure}{\linewidth}
            \centering
            \includegraphics[width=1.0\linewidth]{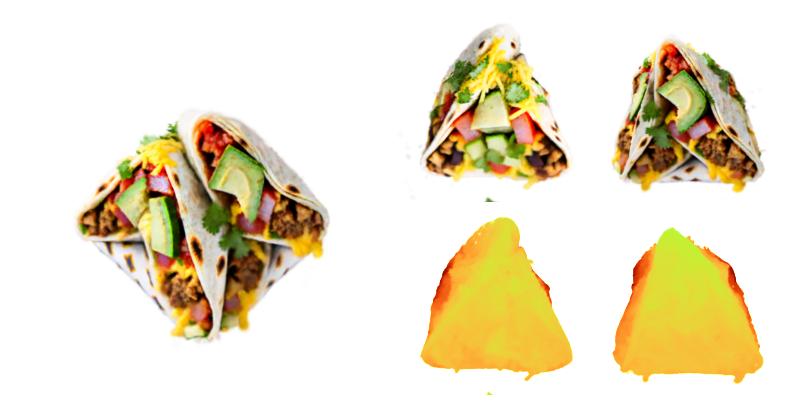} 
            \caption*{\textit{GSGEN}}
        \end{subfigure}
    \end{minipage}
    \\ \textit{A DSLR photo of pyramid shaped burrito with a slice cut out of it}

    \begin{minipage}{0.43\textwidth}
        \centering
        \begin{subfigure}{\linewidth}
            \centering
            \includegraphics[width=1.0\linewidth]{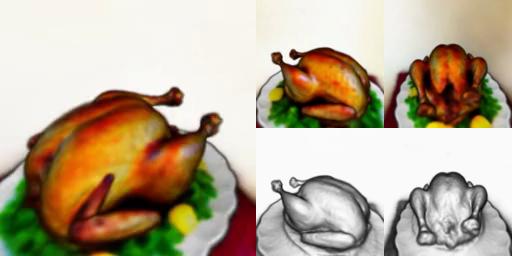} 
            \caption*{\textit{DreamFusion}}
        \end{subfigure}
    \end{minipage}
    \begin{minipage}{0.43\textwidth}
        \centering
        \begin{subfigure}{\linewidth}
            \centering
            \includegraphics[width=1.0\linewidth]{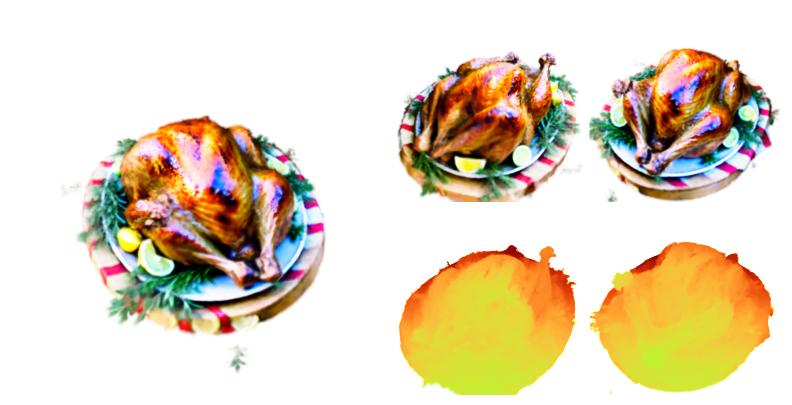} 
            \caption*{\textit{GSGEN}}
        \end{subfigure}
    \end{minipage}
    \\ \textit{A DSLR photo of a roast turkey on a platter}

    \begin{minipage}{0.43\textwidth}
        \centering
        \begin{subfigure}{\linewidth}
            \centering
            \includegraphics[width=1.0\linewidth]{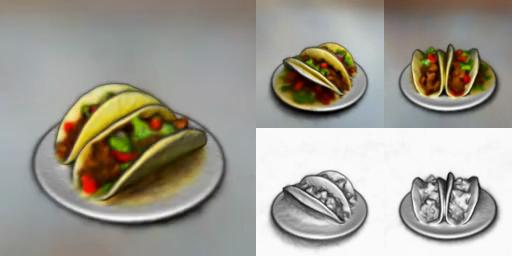} 
            \caption*{\textit{DreamFusion}}
        \end{subfigure}
    \end{minipage}
    \begin{minipage}{0.43\textwidth}
        \centering
        \begin{subfigure}{\linewidth}
            \centering
            \includegraphics[width=1.0\linewidth]{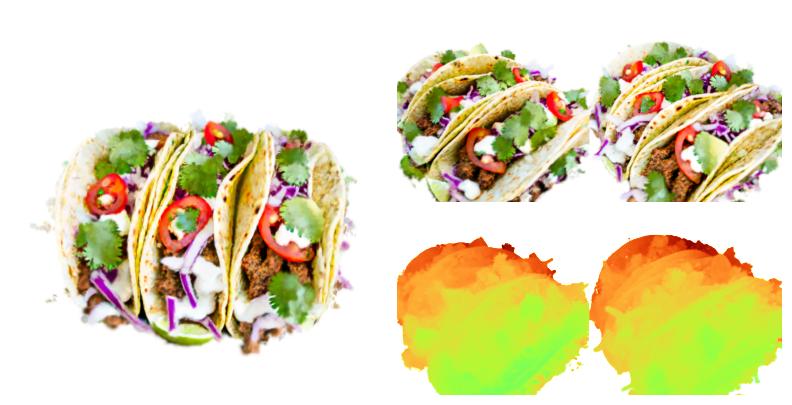} 
            \caption*{\textit{GSGEN}}
        \end{subfigure}
    \end{minipage}
    \\ \textit{A plate of delicious tacos}

    \begin{minipage}{0.43\textwidth}
        \centering
        \begin{subfigure}{\linewidth}
            \centering
            \includegraphics[width=1.0\linewidth]{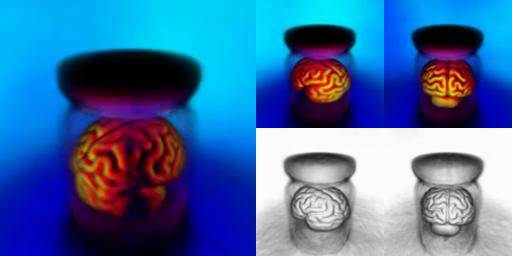} 
            \caption*{\textit{DreamFusion}}
        \end{subfigure}
    \end{minipage}
    \begin{minipage}{0.43\textwidth}
        \centering
        \begin{subfigure}{\linewidth}
            \centering
            \includegraphics[width=1.0\linewidth]{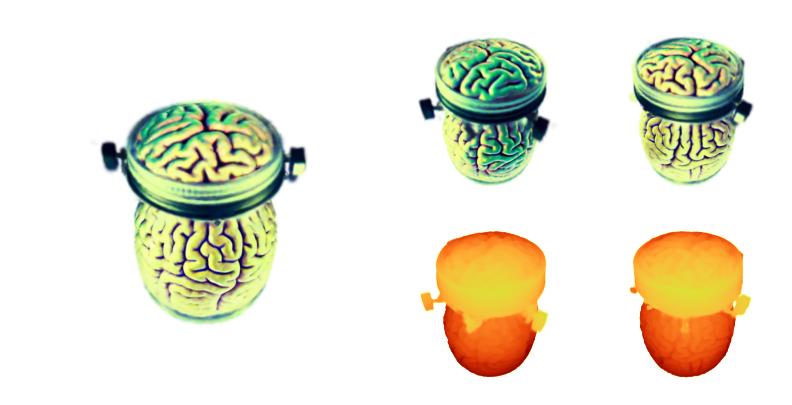} 
            \caption*{\textit{GSGEN}}
        \end{subfigure}
    \end{minipage}
    \\ \textit{A zoomed out DSLR photo of a brain in a jar}

    \begin{minipage}{0.43\textwidth}
        \centering
        \begin{subfigure}{\linewidth}
            \centering
            \includegraphics[width=1.0\linewidth]{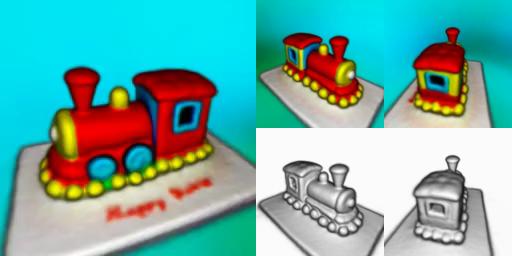} 
            \caption*{\textit{DreamFusion}}
        \end{subfigure}
    \end{minipage}
    \begin{minipage}{0.43\textwidth}
        \centering
        \begin{subfigure}{\linewidth}
            \centering
            \includegraphics[width=1.0\linewidth]{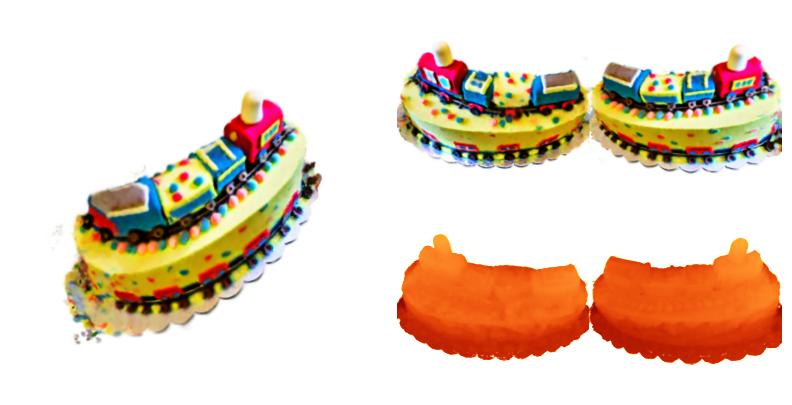} 
            \caption*{\textit{GSGEN}}
        \end{subfigure}
    \end{minipage}
    \\ \textit{A zoomed out DSLR photo of a cake in the shape of a train}
    \caption{More comparison results with DreamFusion.}
    \label{fig:more_comparison_df_1}
\end{figure*}

\begin{figure*}[t]
    \vspace{-8mm}
    \centering
    \begin{minipage}{0.43\textwidth}
        \centering
        \begin{subfigure}{\linewidth}
            \centering
            \includegraphics[width=1.0\linewidth]{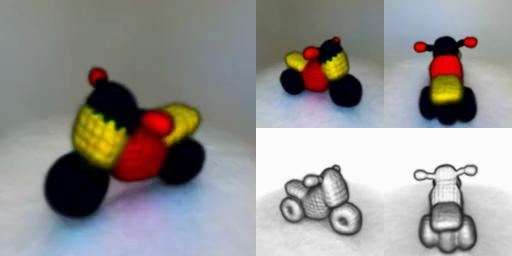} 
            \caption*{\textit{DreamFusion}}
        \end{subfigure}
    \end{minipage}
    \begin{minipage}{0.43\textwidth}
        \centering
        \begin{subfigure}{\linewidth}
            \centering
            \includegraphics[width=1.0\linewidth]{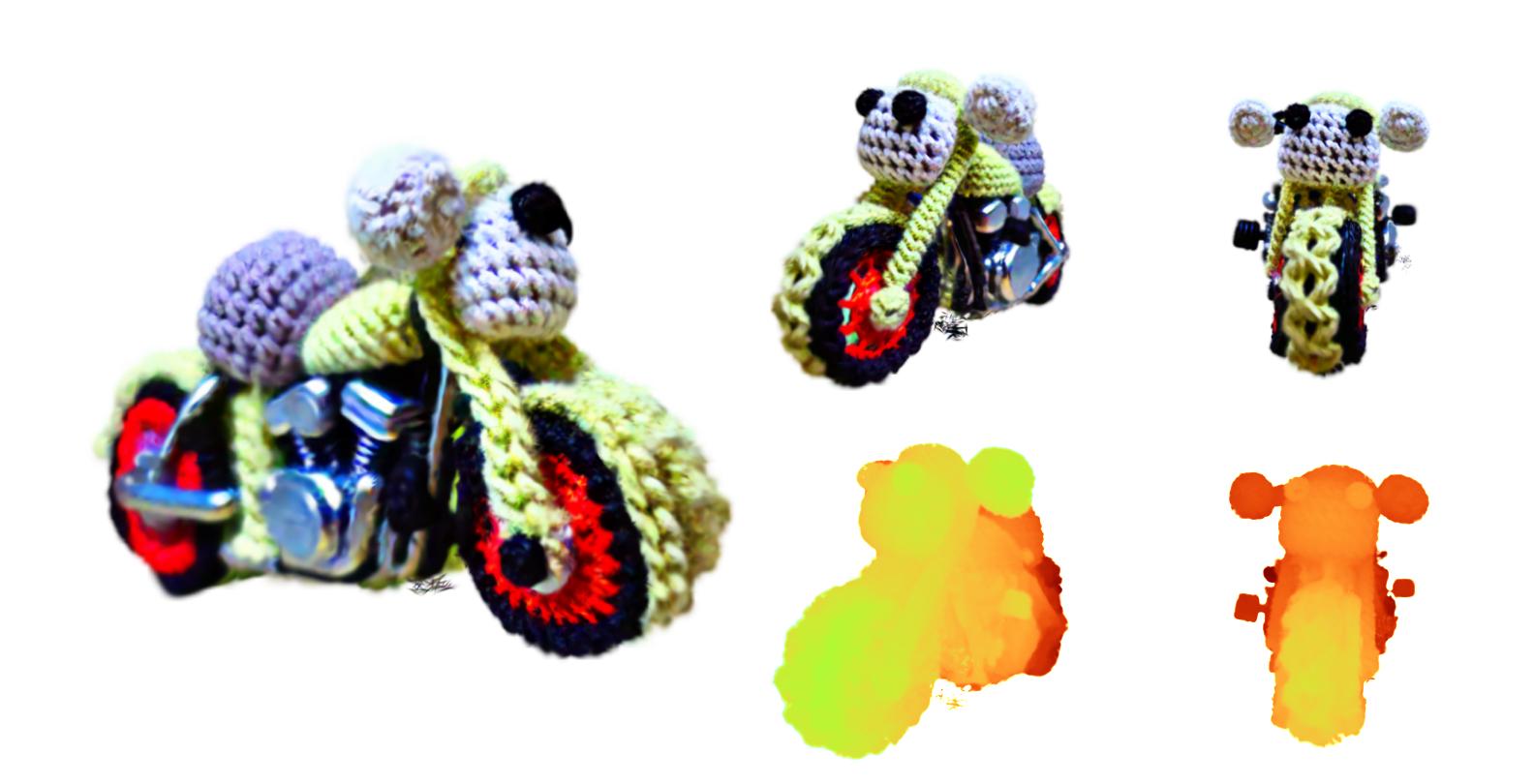} 
            \caption*{\textit{GSGEN}}
        \end{subfigure}
    \end{minipage}
    \\ \textit{A zoomed out DSLR photo of an amigurumi motorcycle}

    \begin{minipage}{0.43\textwidth}
        \centering
        \begin{subfigure}{\linewidth}
            \centering
            \includegraphics[width=1.0\linewidth]{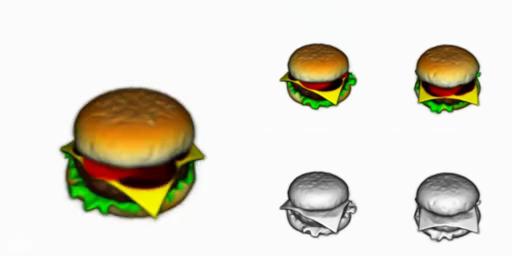} 
            \caption*{\textit{DreamFusion}}
        \end{subfigure}
    \end{minipage}
    \begin{minipage}{0.43\textwidth}
        \centering
        \begin{subfigure}{\linewidth}
            \centering
            \includegraphics[width=1.0\linewidth]{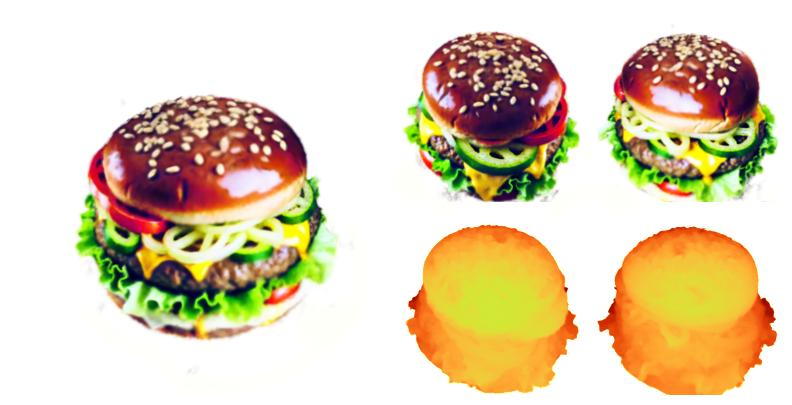} 
            \caption*{\textit{GSGEN}}
        \end{subfigure}
    \end{minipage}
    \\ \textit{A delicious hamburger}

    \begin{minipage}{0.43\textwidth}
        \centering
        \begin{subfigure}{\linewidth}
            \centering
            \includegraphics[width=1.0\linewidth]{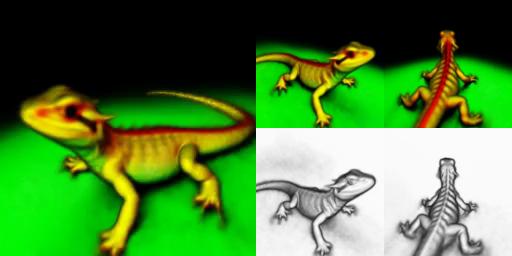} 
            \caption*{\textit{DreamFusion}}
        \end{subfigure}
    \end{minipage}
    \begin{minipage}{0.43\textwidth}
        \centering
        \begin{subfigure}{\linewidth}
            \centering
            \includegraphics[width=1.0\linewidth]{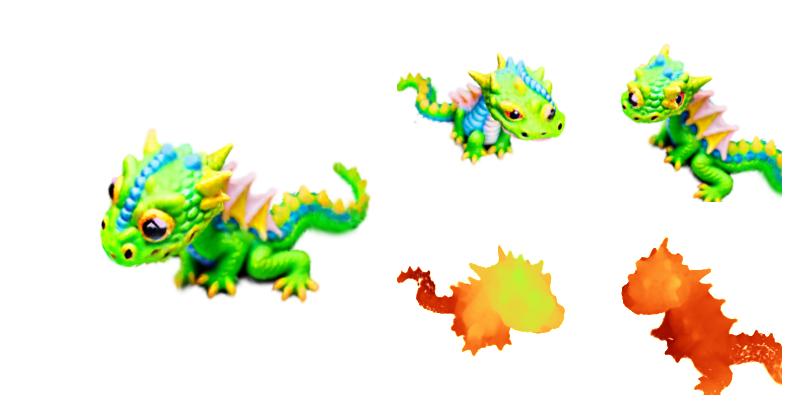} 
            \caption*{\textit{GSGEN}}
        \end{subfigure}
    \end{minipage}
    \\ \textit{A zoomed out DSLR photo of a baby dragon}

    \begin{minipage}{0.43\textwidth}
        \centering
        \begin{subfigure}{\linewidth}
            \centering
            \includegraphics[width=1.0\linewidth]{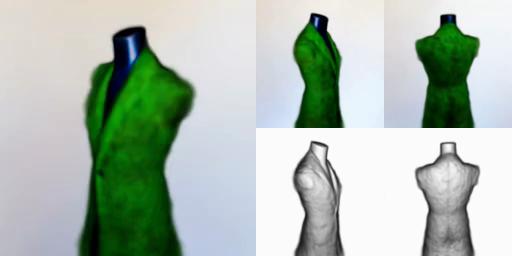} 
            \caption*{\textit{DreamFusion}}
        \end{subfigure}
    \end{minipage}
    \begin{minipage}{0.43\textwidth}
        \centering
        \begin{subfigure}{\linewidth}
            \centering
            \includegraphics[width=1.0\linewidth]{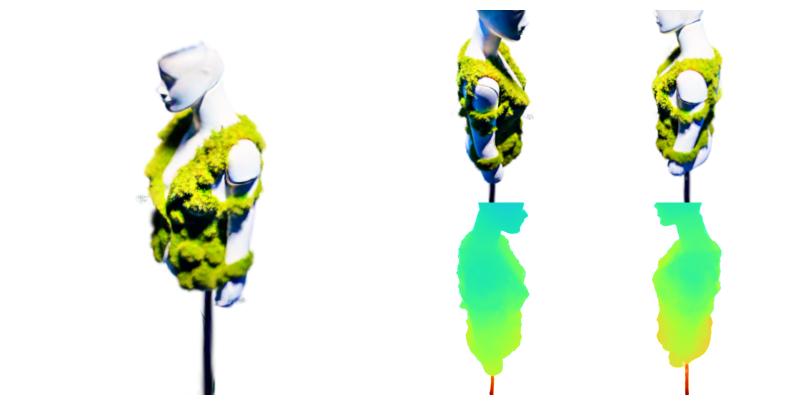} 
            \caption*{\textit{GSGEN}}
        \end{subfigure}
    \end{minipage}
    \\ \textit{A zoomed out DSLR photo of a beautiful suit made out of moss, on a mannequin. Studio lighting, high quality, high resolution}

    \begin{minipage}{0.43\textwidth}
        \centering
        \begin{subfigure}{\linewidth}
            \centering
            \includegraphics[width=1.0\linewidth]{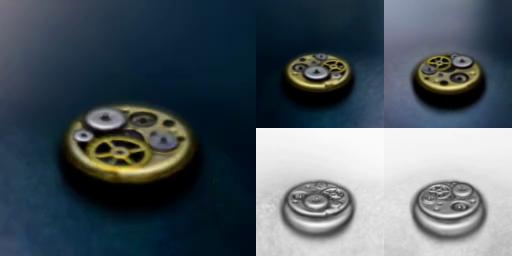} 
            \caption*{\textit{DreamFusion}}
        \end{subfigure}
    \end{minipage}
    \begin{minipage}{0.43\textwidth}
        \centering
        \begin{subfigure}{\linewidth}
            \centering
            \includegraphics[width=1.0\linewidth]{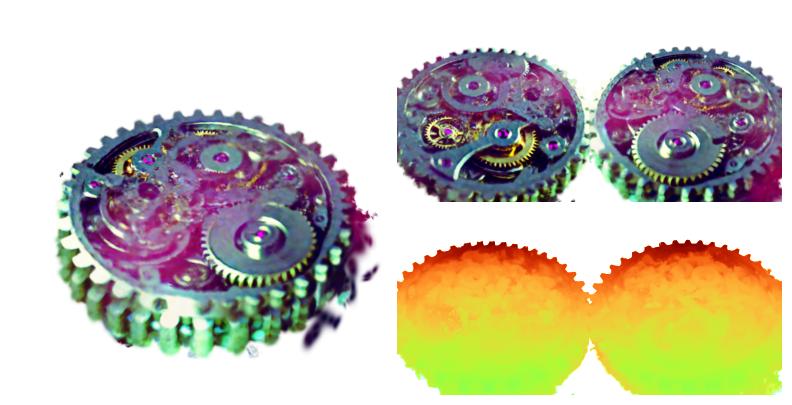} 
            \caption*{\textit{GSGEN}}
        \end{subfigure}
    \end{minipage}
    \\ \textit{A zoomed out DSLR photo of a complex movement from an expensive watch with many shiny gears, sitting on a table}

    
    \caption{More comparison results with DreamFusion.}
    \label{fig:more_comparison_df_2}
\end{figure*}

\begin{figure*}[t]
    \vspace{-1cm}
    \centering
    \begin{minipage}{0.49\textwidth}
        \centering
        \begin{subfigure}{\linewidth}
            \centering
            \includegraphics[width=1.0\linewidth]{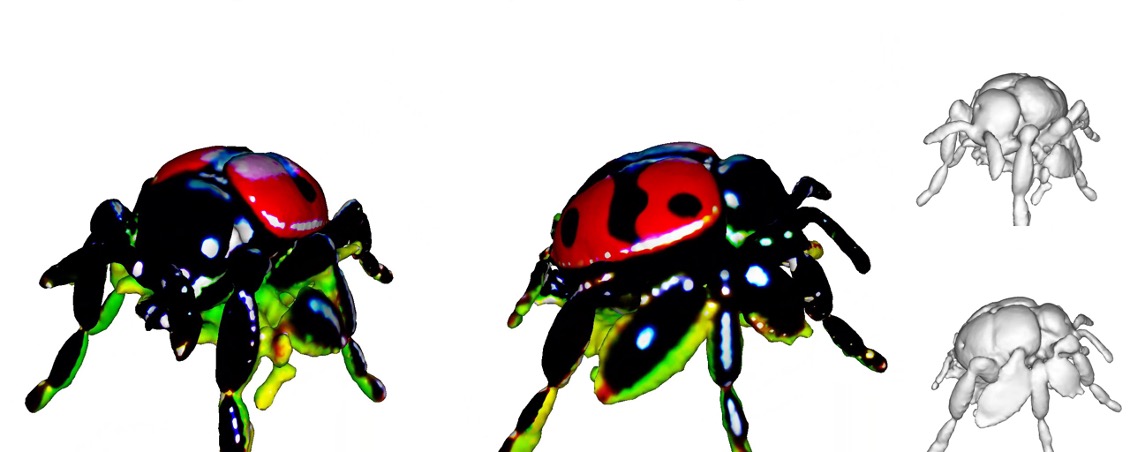} 
            \caption*{\textit{Magic3D}}
        \end{subfigure}
    \end{minipage}
    \begin{minipage}{0.49\textwidth}
        \centering
        \begin{subfigure}{\linewidth}
            \centering
            \includegraphics[width=1.0\linewidth]{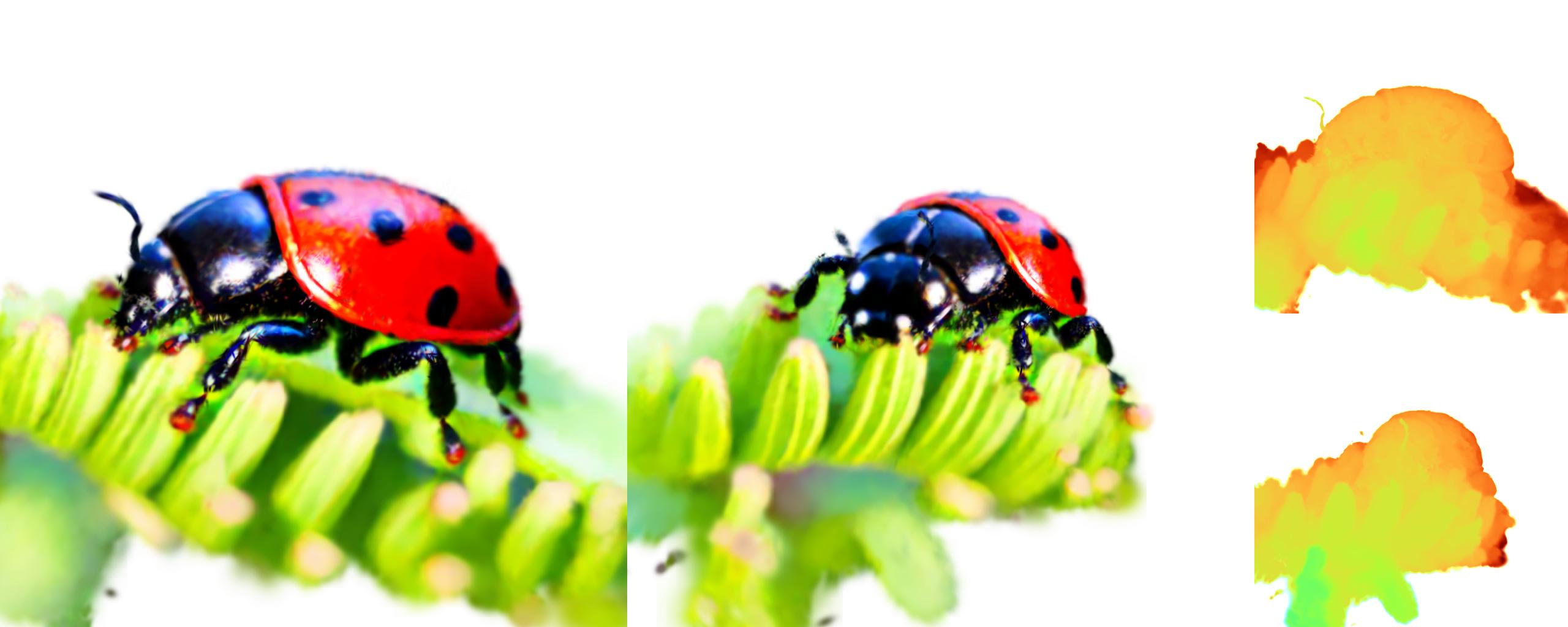} 
            \caption*{\textit{GSGEN}}
        \end{subfigure}
    \end{minipage}
    \\ \textit{A zoomed out DSLR photo of a ladybug}

    \begin{minipage}{0.49\textwidth}
        \centering
        \begin{subfigure}{\linewidth}
            \centering
            \includegraphics[width=1.0\linewidth]{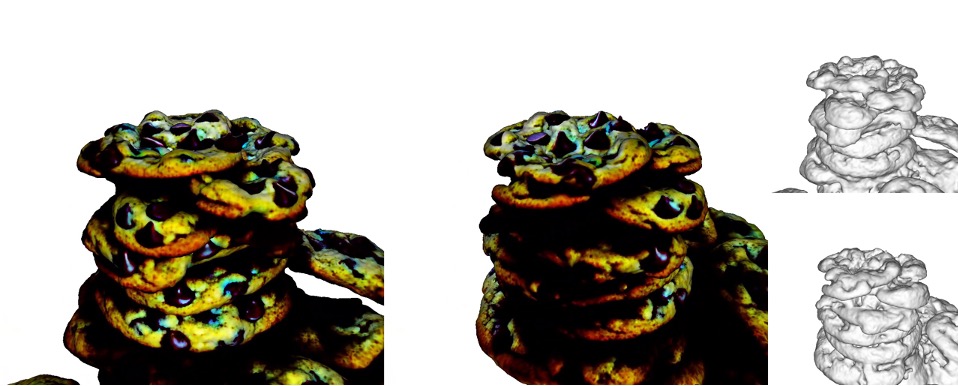} 
            \caption*{\textit{Magic3D}}
        \end{subfigure}
    \end{minipage}
    \begin{minipage}{0.49\textwidth}
        \centering
        \begin{subfigure}{\linewidth}
            \centering
            \includegraphics[width=1.0\linewidth]{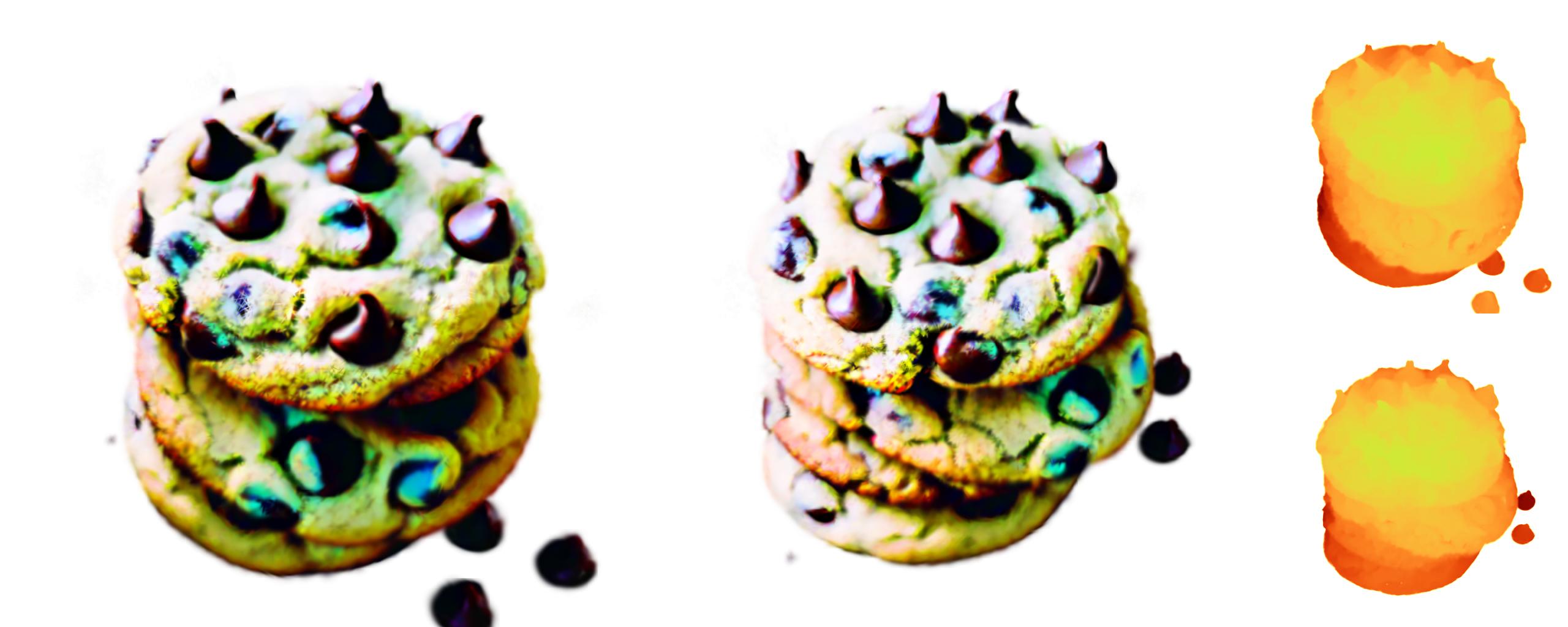} 
            \caption*{\textit{GSGEN}}
        \end{subfigure}
    \end{minipage}
    \\ \textit{A zoomed out DSLR photo of a plate piled high with chocolate chip cookies}

    \begin{minipage}{0.49\textwidth}
        \centering
        \begin{subfigure}{\linewidth}
            \centering
            \includegraphics[width=1.0\linewidth]{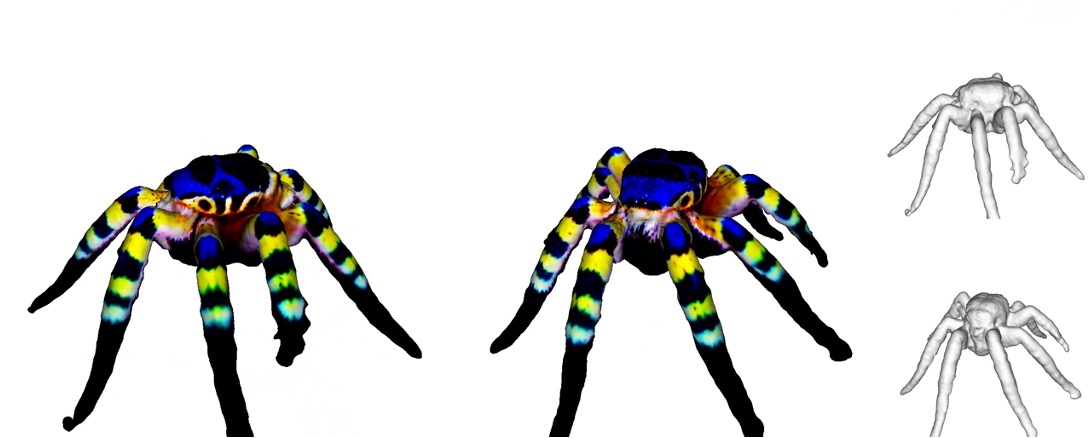} 
            \caption*{\textit{Magic3D}}
        \end{subfigure}
    \end{minipage}
    \begin{minipage}{0.49\textwidth}
        \centering
        \begin{subfigure}{\linewidth}
            \centering
            \includegraphics[width=1.0\linewidth]{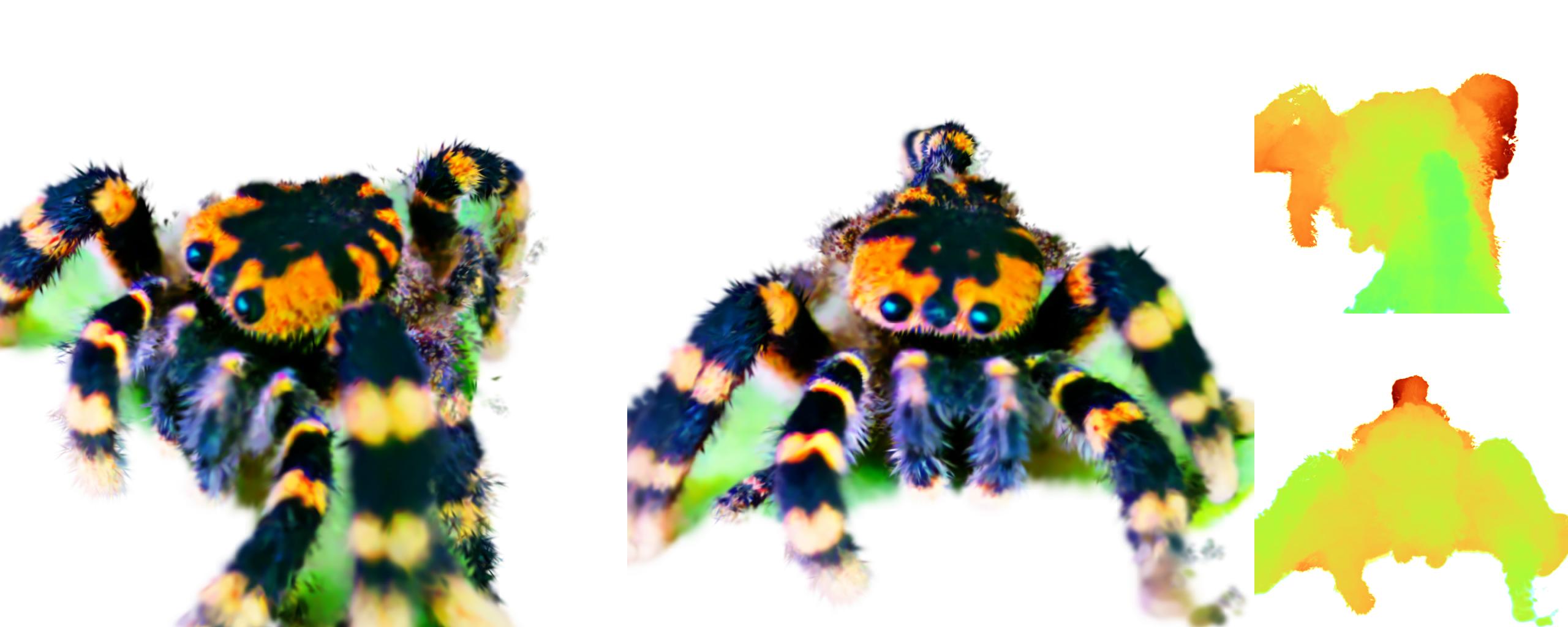} 
            \caption*{\textit{GSGEN}}
        \end{subfigure}
    \end{minipage}
    \\ \textit{A DSLR photo of a tarantula, highly detailed}

    \begin{minipage}{0.49\textwidth}
        \centering
        \begin{subfigure}{\linewidth}
            \centering
            \includegraphics[width=1.0\linewidth]{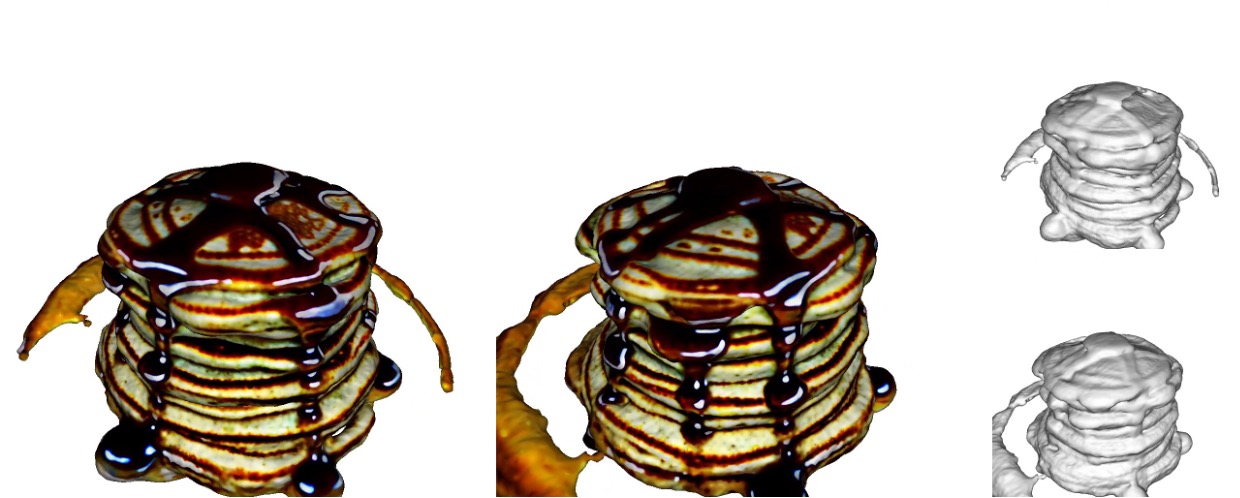} 
            \caption*{\textit{Magic3D}}
        \end{subfigure}
    \end{minipage}
    \begin{minipage}{0.49\textwidth}
        \centering
        \begin{subfigure}{\linewidth}
            \centering
            \includegraphics[width=1.0\linewidth]{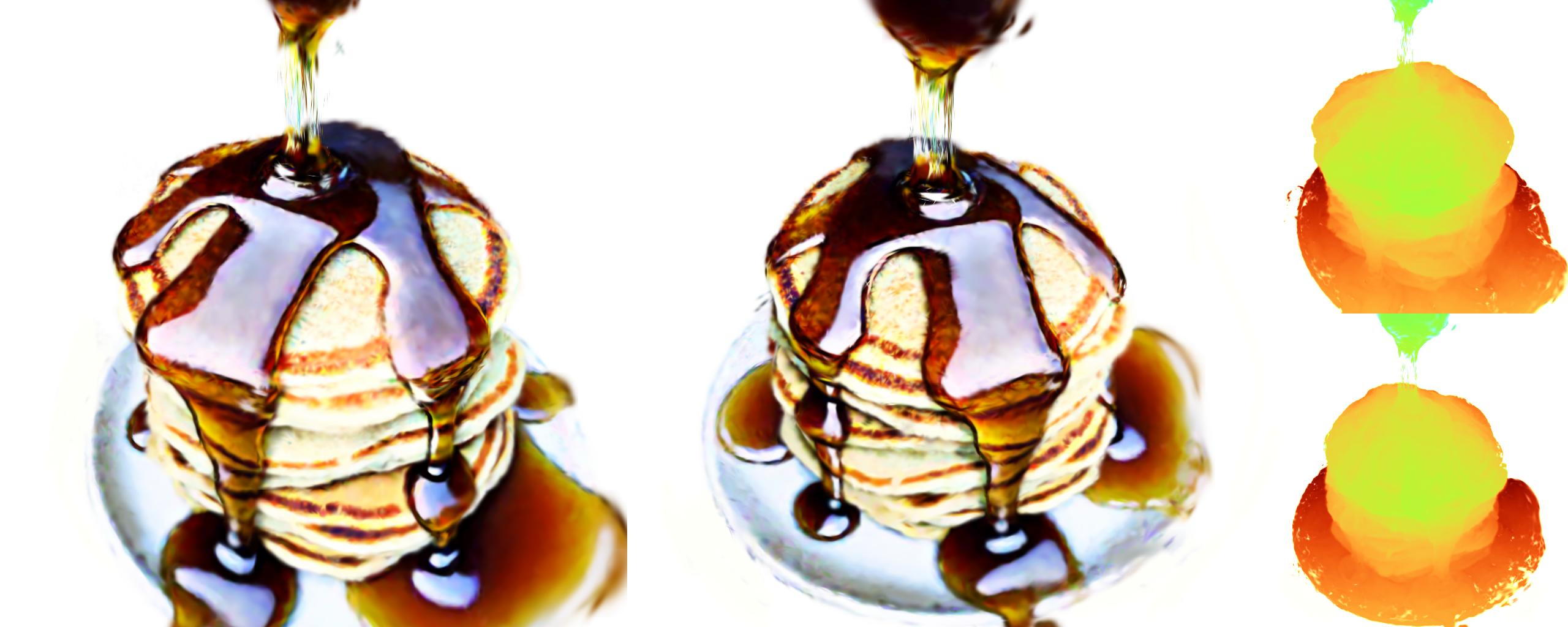} 
            \caption*{\textit{GSGEN}}
        \end{subfigure}
    \end{minipage}
    \\ \textit{A DSLR photo of a stack of pancakes covered in maple syrup}

    \begin{minipage}{0.49\textwidth}
        \centering
        \begin{subfigure}{\linewidth}
            \centering
            \includegraphics[width=1.0\linewidth]{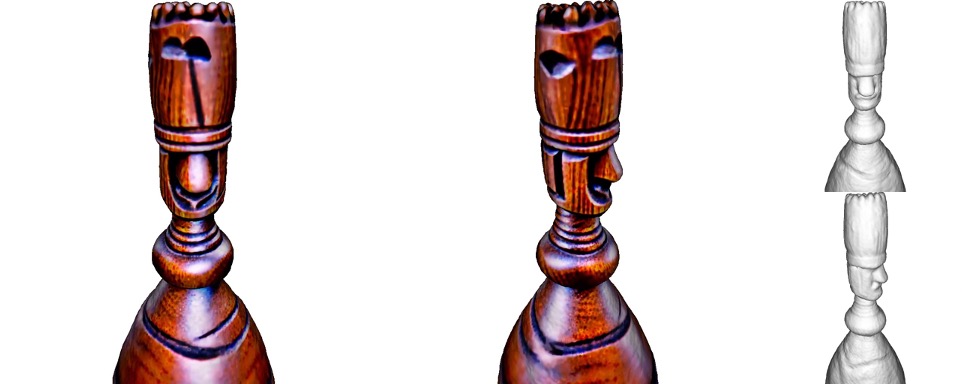} 
            \caption*{\textit{Magic3D}}
        \end{subfigure}
    \end{minipage}
    \begin{minipage}{0.49\textwidth}
        \centering
        \begin{subfigure}{\linewidth}
            \centering
            \includegraphics[width=1.0\linewidth]{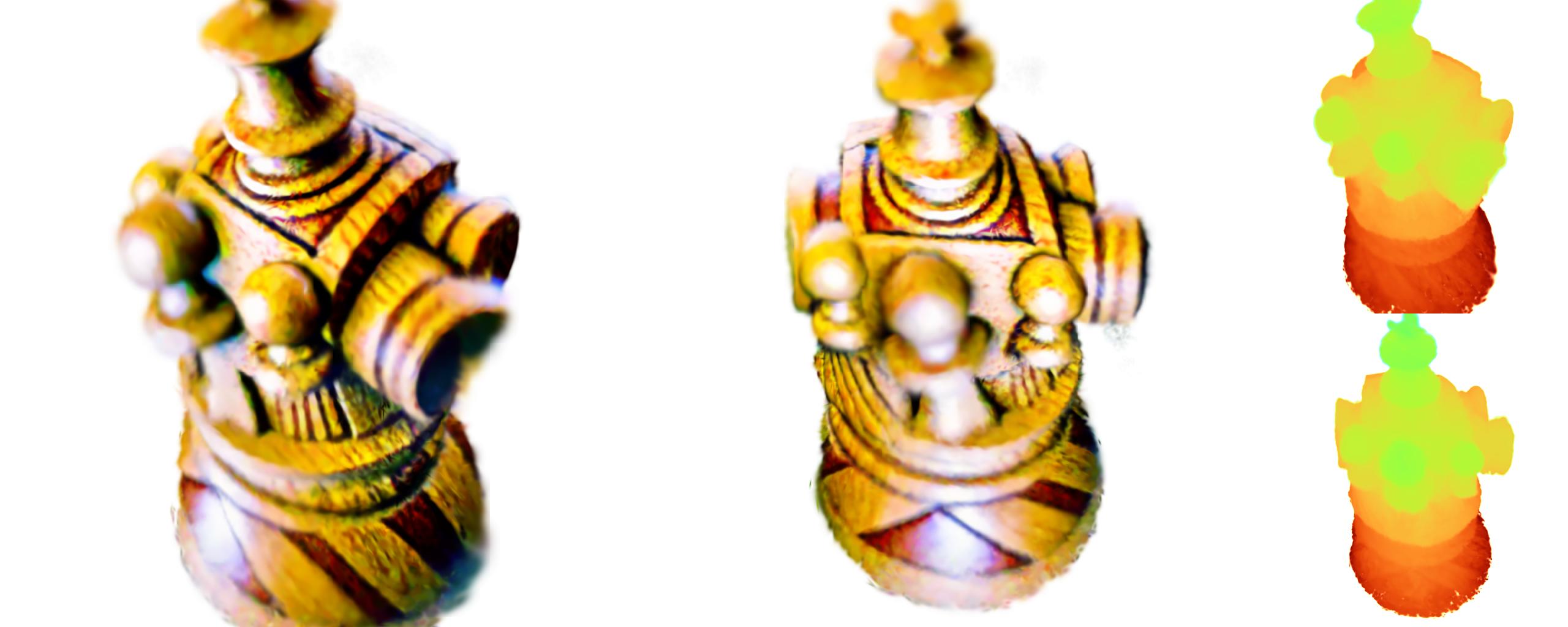} 
            \caption*{\textit{GSGEN}}
        \end{subfigure}
    \end{minipage}
    \\ \textit{A zoomed out DSLR photo of a beautifully carved wooden knight chess piece}
    \caption{More comparison results with Magic3D.}
    \label{fig:more_comparison_magic3d}
\end{figure*}

\begin{figure*}[t]
    \vspace{-1cm}
    \centering
    \begin{minipage}{0.49\textwidth}
        \centering
        \begin{subfigure}{\linewidth}
            \centering
            \includegraphics[width=1.0\linewidth]{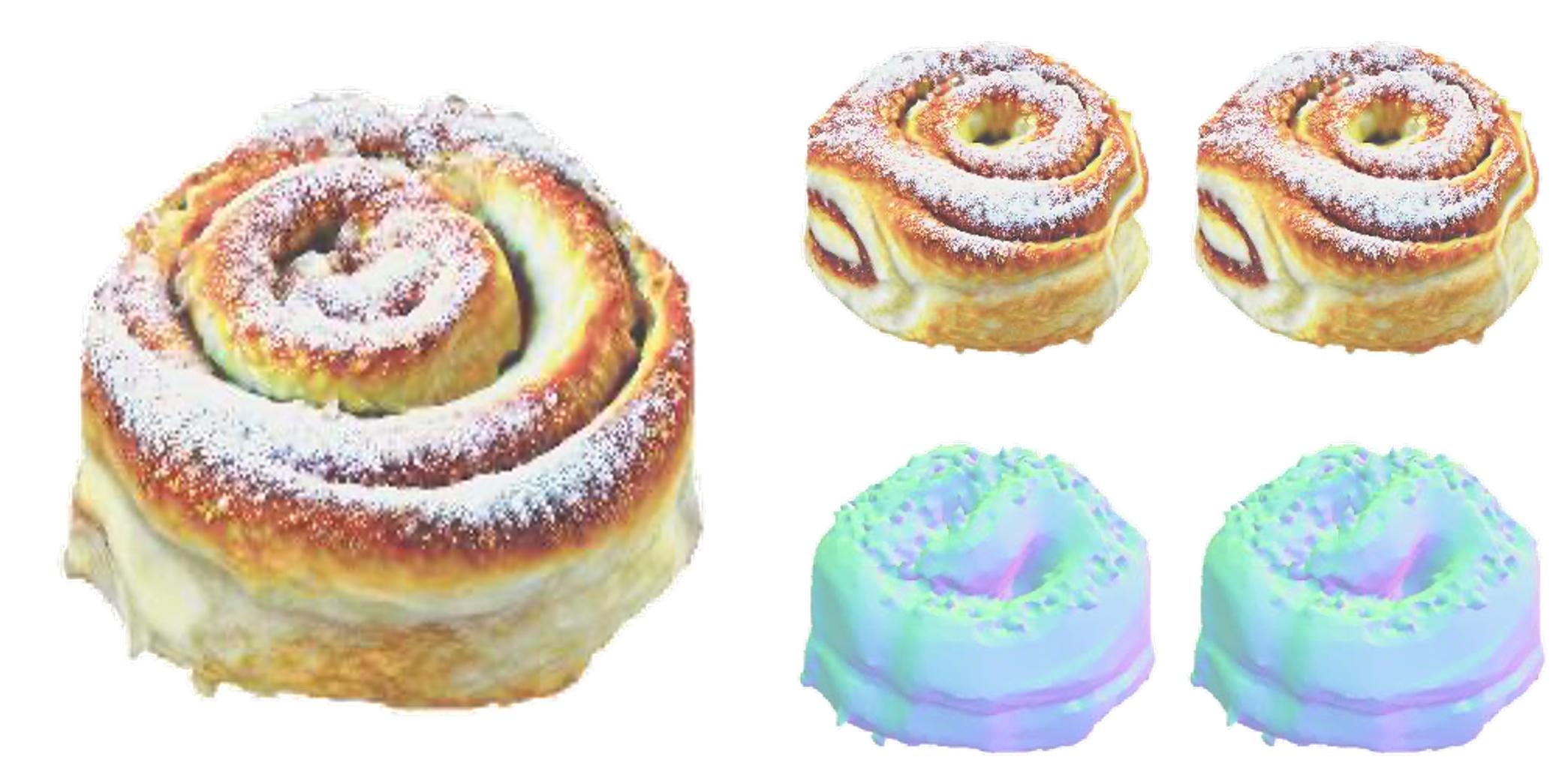} 
            \caption*{\textit{Fantasia3D}}
        \end{subfigure}
    \end{minipage}
    \begin{minipage}{0.49\textwidth}
        \centering
        \begin{subfigure}{\linewidth}
            \centering
            \includegraphics[width=1.0\linewidth]{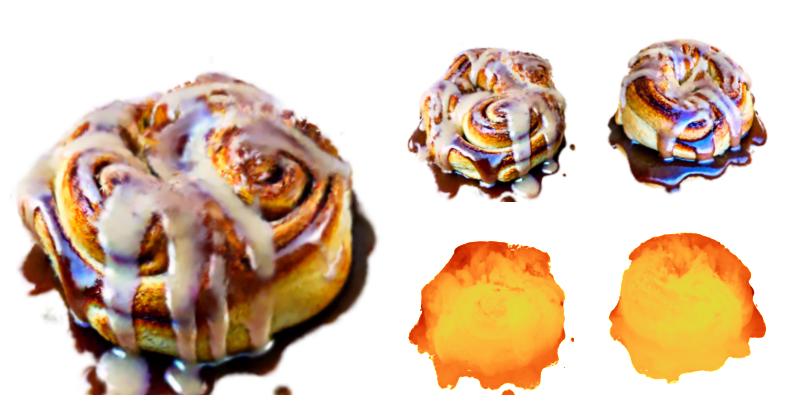} 
            \caption*{\textit{GSGEN}}
        \end{subfigure}
    \end{minipage}
    \\ \textit{A fresh cinnamon roll covered in glaze, high resolution}

    \begin{minipage}{0.49\textwidth}
        \centering
        \begin{subfigure}{\linewidth}
            \centering
            \includegraphics[width=1.0\linewidth]{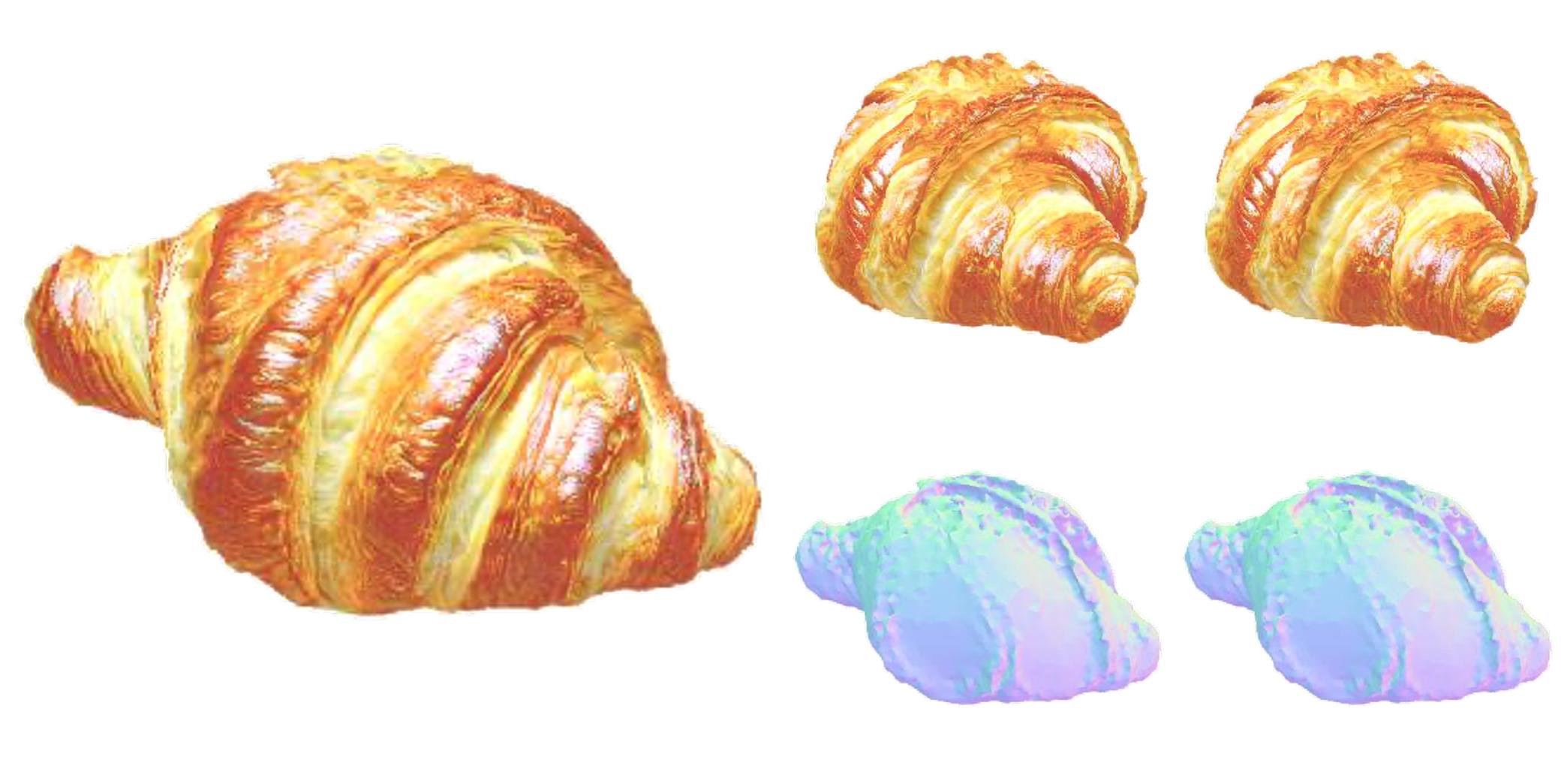} 
            \caption*{\textit{Fantasia3D}}
        \end{subfigure}
    \end{minipage}
    \begin{minipage}{0.49\textwidth}
        \centering
        \begin{subfigure}{\linewidth}
            \centering
            \includegraphics[width=1.0\linewidth]{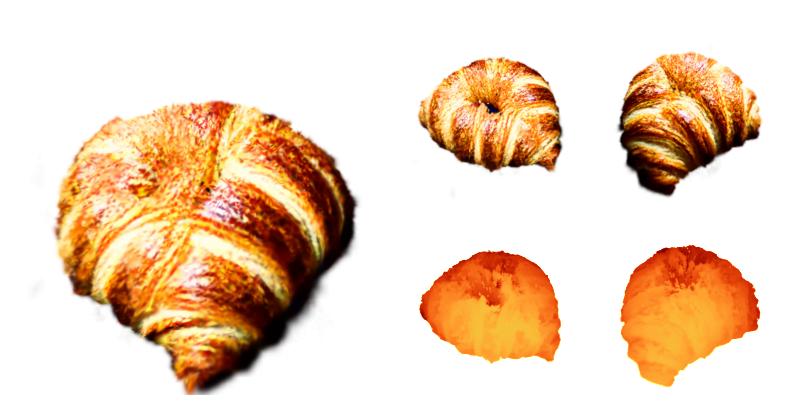} 
            \caption*{\textit{GSGEN}}
        \end{subfigure}
    \end{minipage}
    \\ \textit{A delicious croissant}

    \begin{minipage}{0.49\textwidth}
        \centering
        \begin{subfigure}{\linewidth}
            \centering
            \includegraphics[width=1.0\linewidth]{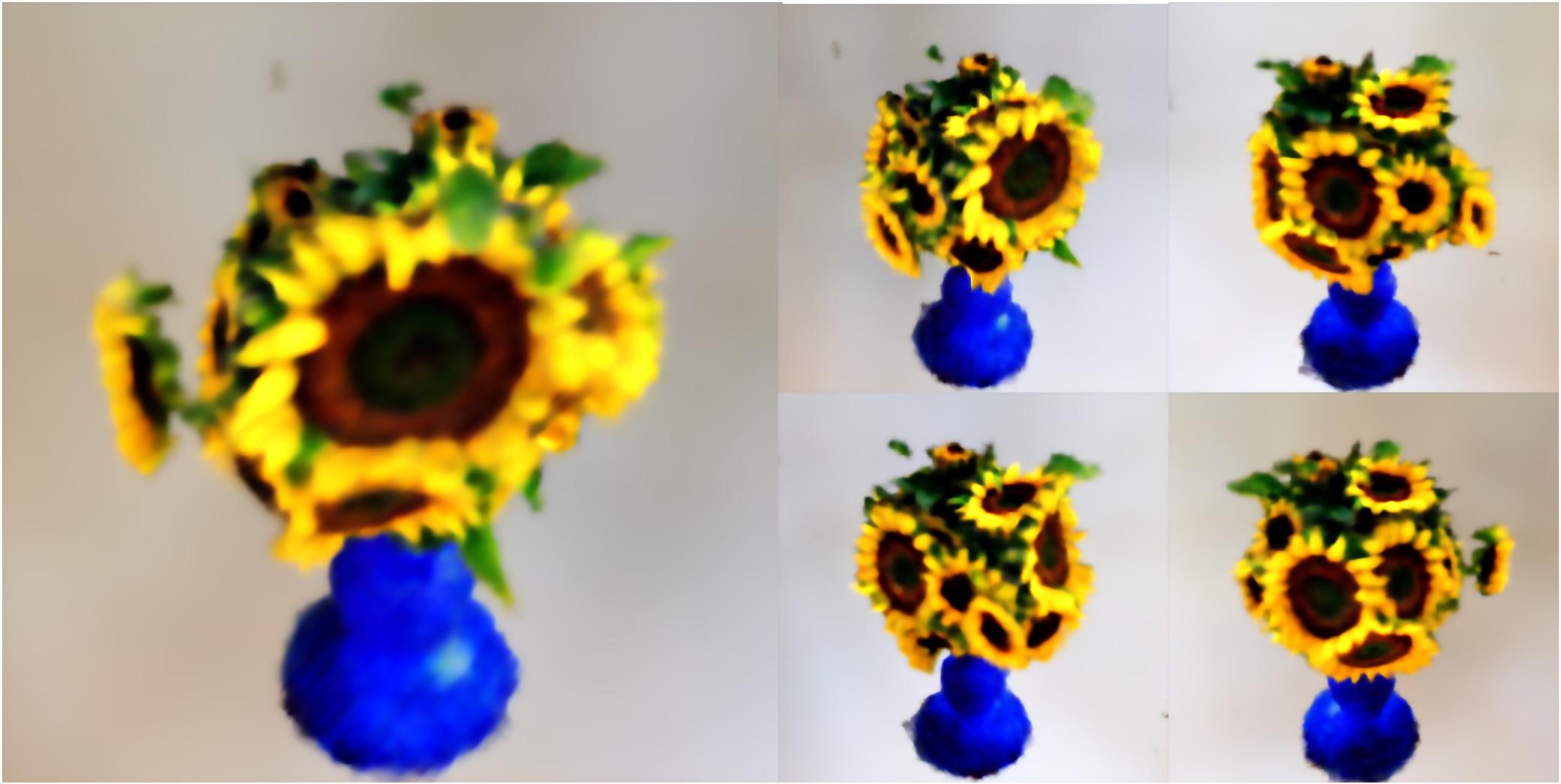} 
            \caption*{\textit{LatentNeRF}}
        \end{subfigure}
    \end{minipage}
    \begin{minipage}{0.49\textwidth}
        \centering
        \begin{subfigure}{\linewidth}
            \centering
            \includegraphics[width=1.0\linewidth]{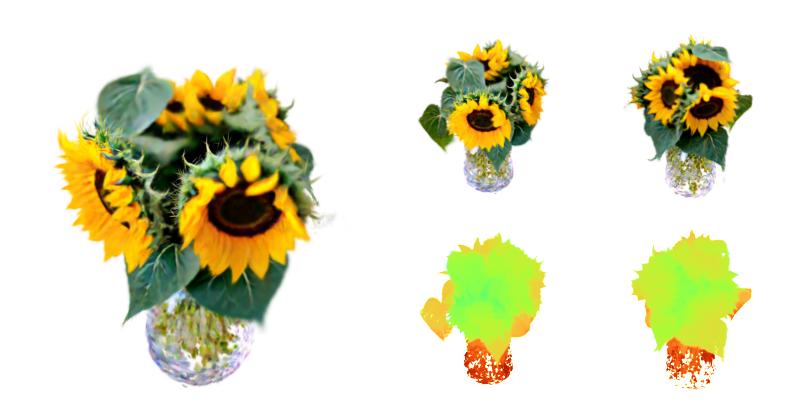} 
            \caption*{\textit{GSGEN}}
        \end{subfigure}
    \end{minipage}
    \\ \textit{A photo of a vase with sunﬂowers}


    \begin{minipage}{0.49\textwidth}
        \centering
        \begin{subfigure}{\linewidth}
            \centering
            \includegraphics[width=1.0\linewidth]{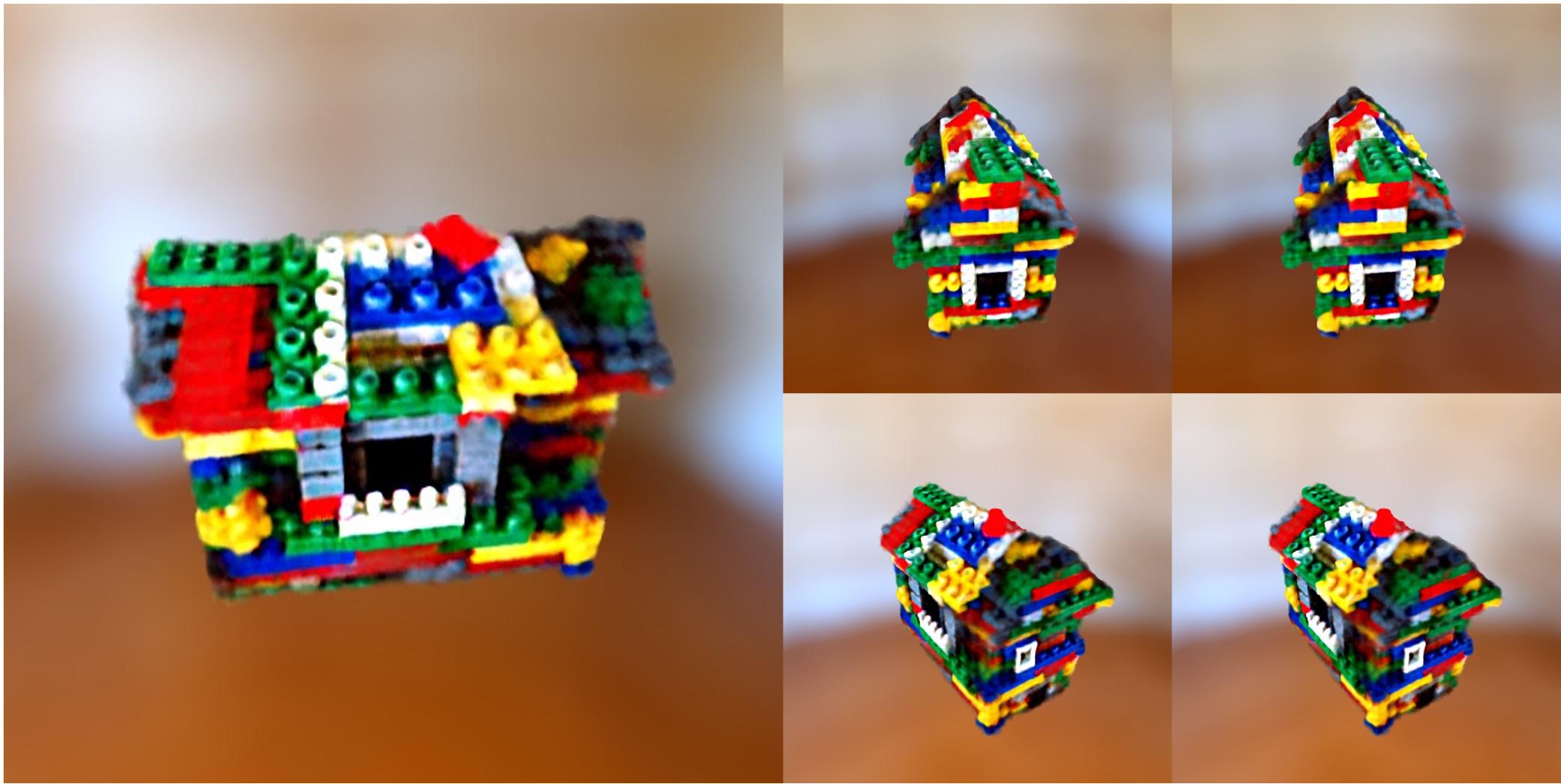} 
            \caption*{\textit{LatentNeRF}}
        \end{subfigure}
    \end{minipage}
    \begin{minipage}{0.49\textwidth}
        \centering
        \begin{subfigure}{\linewidth}
            \centering
            \includegraphics[width=1.0\linewidth]{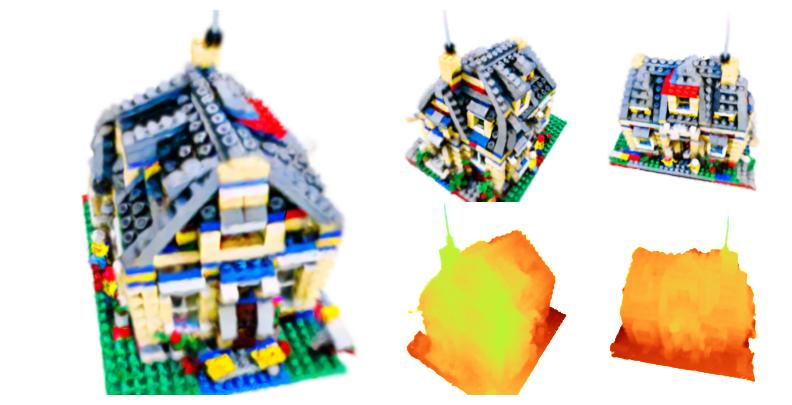} 
            \caption*{\textit{GSGEN}}
        \end{subfigure}
    \end{minipage}
    \\ \textit{A house made of lego}
    \caption{More comparison results with LatentNeRF and Fantasia3D.}
    \label{fig:more_comparison_others}
\end{figure*}

In addition to the qualitative comparison in the main text, we provide more comparisons with DreamFusion \citep{dreamfusion} in Fig.~\ref{fig:more_comparison_df_1} and Fig.~\ref{fig:more_comparison_df_2}, Magic3D \citep{magic3d} in Fig.~\ref{fig:more_comparison_magic3d}, Fantasia3D \citep{fantasia3d} and LatentNeRF \citep{latentnerf} in Fig.~\ref{fig:more_comparison_others}. In order to make a fair comparison, the images of these methods are directly copied from their papers or project pages. Video comparisons are presented in the supplemental video.

\begin{figure}
    \centering
    \begin{minipage}{0.235\textwidth}
        \centering
        \begin{subfigure}{\linewidth}
            \centering
            \includegraphics[width=1.0\linewidth]{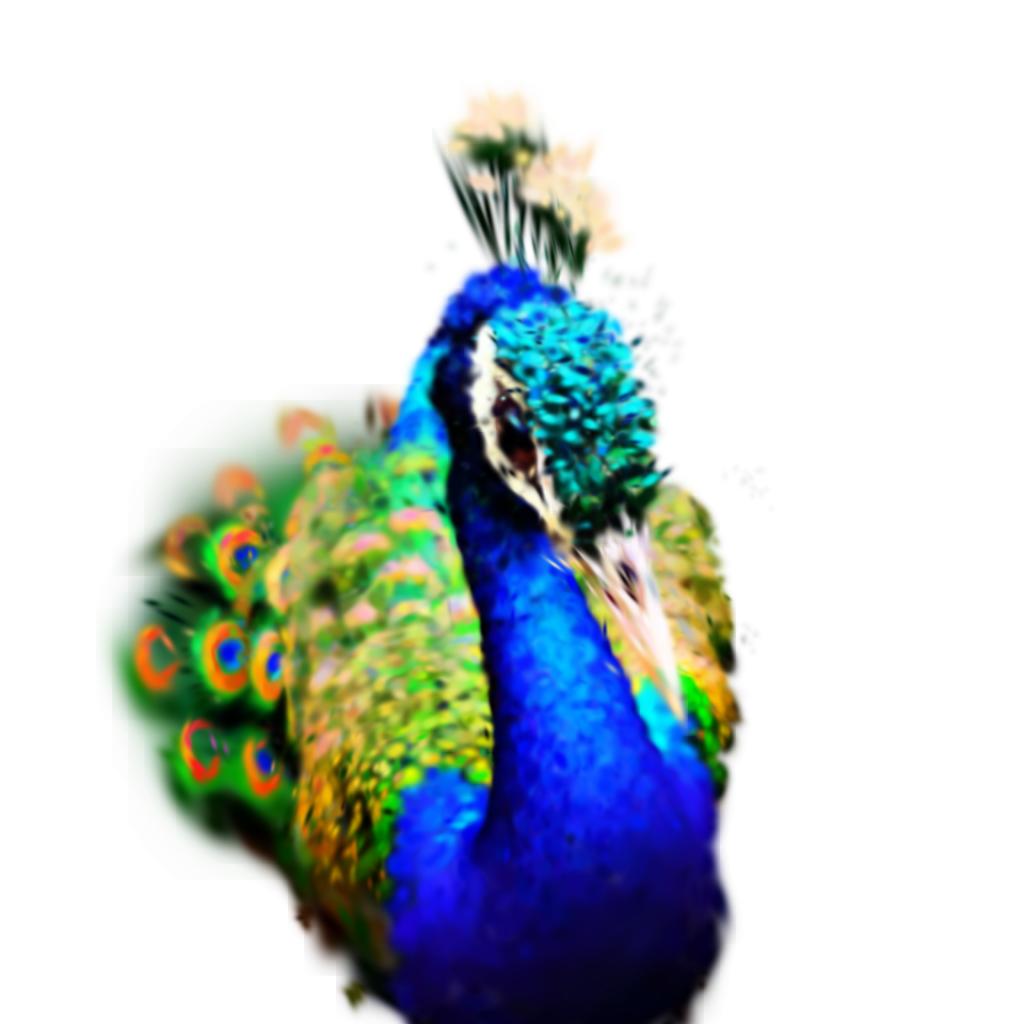} 
            \caption*{\textit{w/o Point-E color}}
        \end{subfigure}
    \end{minipage}
    \hfill
    \begin{minipage}{0.235\textwidth}
        \centering
        \begin{subfigure}{\linewidth}
            \centering
            \includegraphics[width=1.0\linewidth]{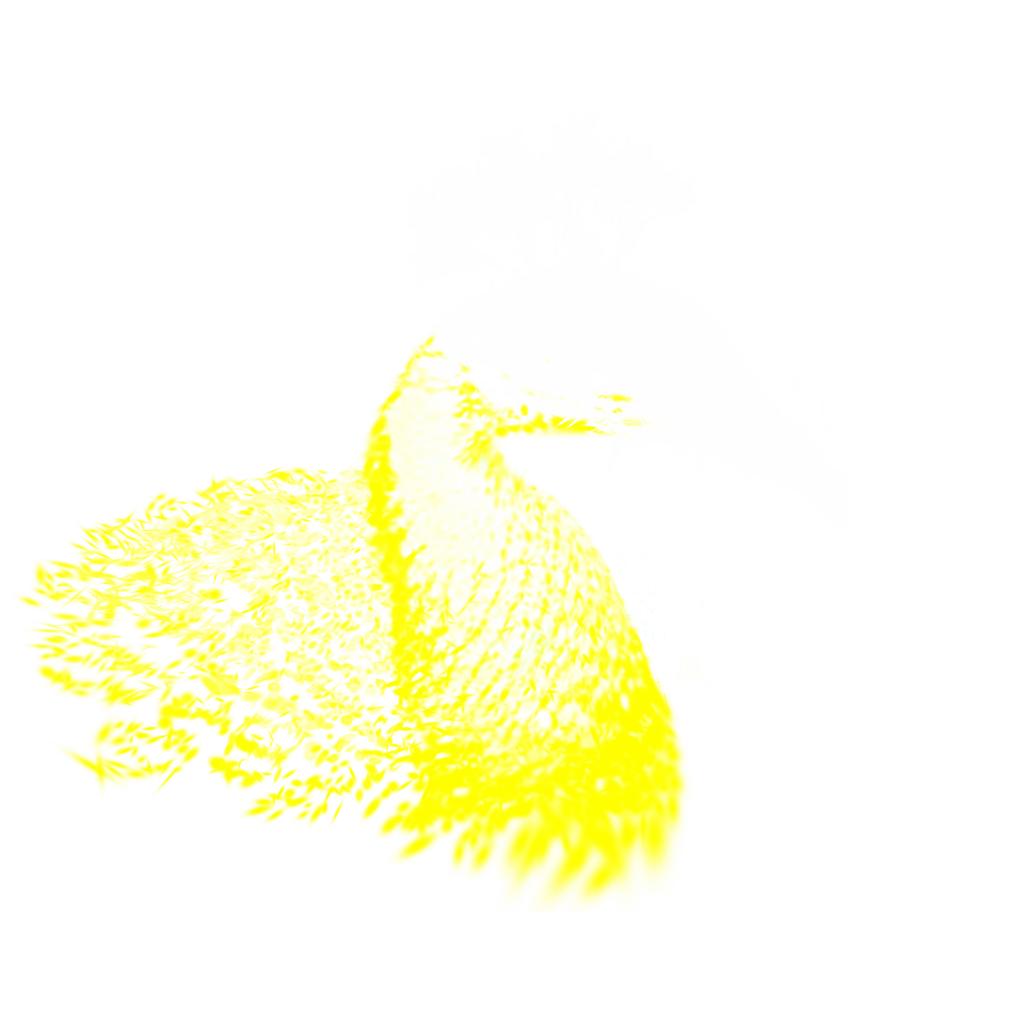} 
            \caption*{\textit{w/ Point-E color}}
        \end{subfigure}
    \end{minipage}
    \caption{The impact of adopting Point-E generated color.}
    \vspace{-4mm}
    \label{fig:no_pointe_color}
\end{figure}

\begin{figure}
    \centering
    \centering
    \begin{minipage}{0.23\textwidth}
        \centering
        \begin{subfigure}{\linewidth}
            \centering
            \caption*{Point-E}
            \includegraphics[width=1.0\linewidth]{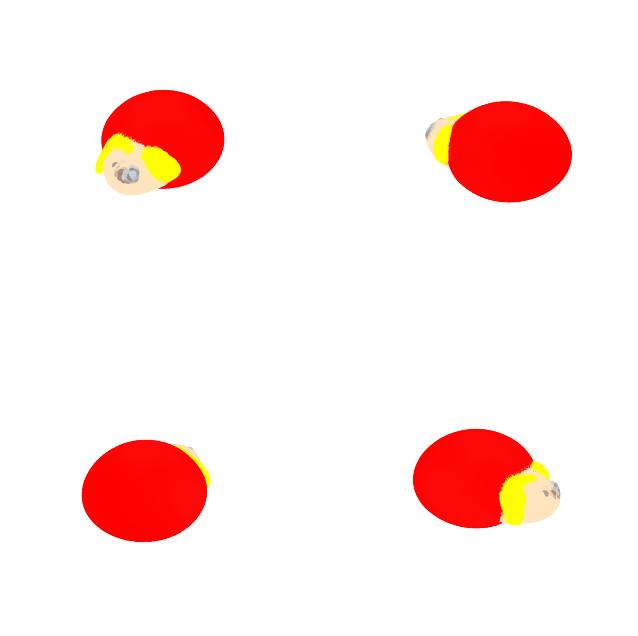} 
        \end{subfigure}
    \end{minipage}
    \hfill
    \begin{minipage}{0.23\textwidth}
        \centering
        \begin{subfigure}{\linewidth}
            \centering
            \caption*{\approach}
            \includegraphics[width=1.0\linewidth]{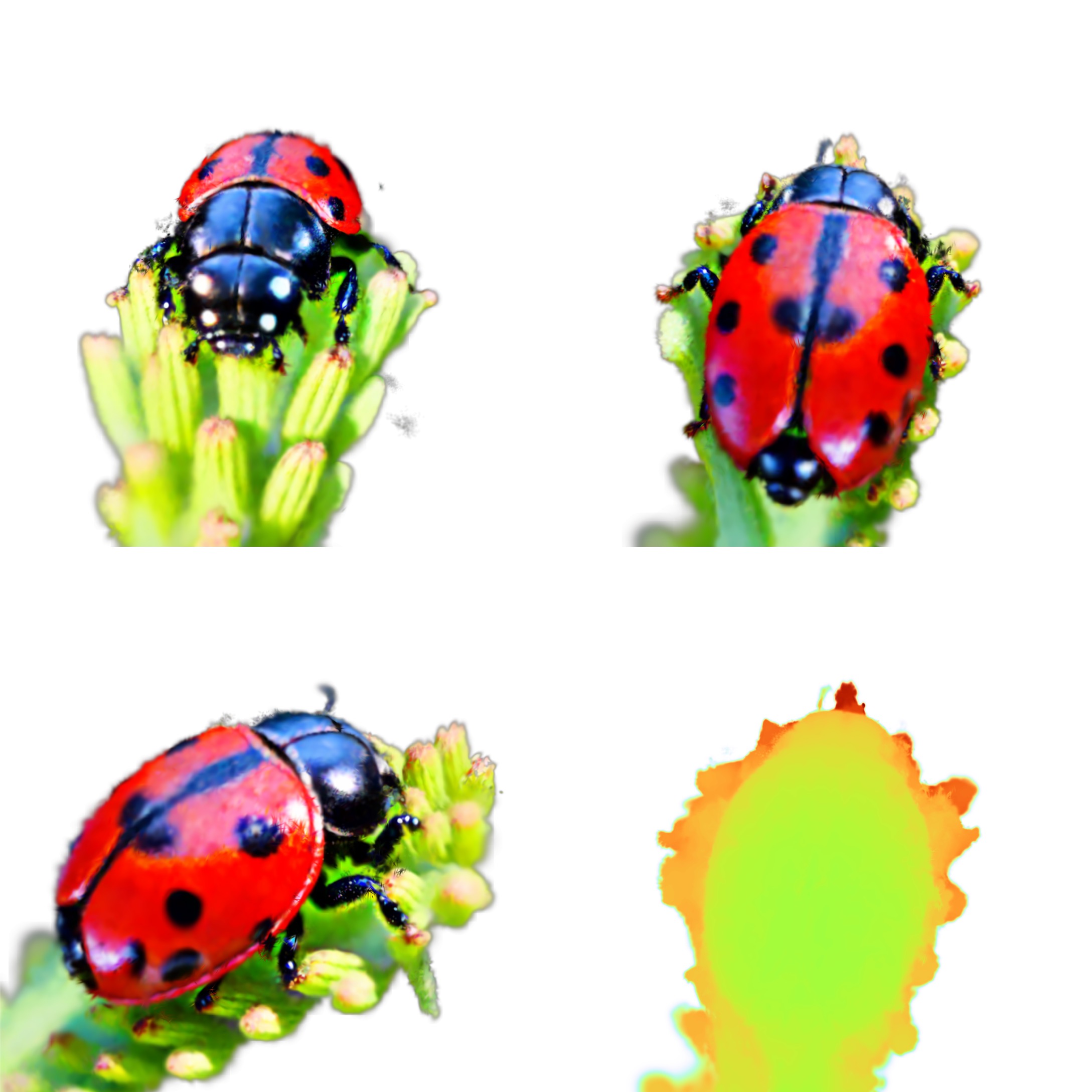} 
        \end{subfigure}
    \end{minipage}
    \\ \textit{a ladybug}

    \begin{minipage}{0.23\textwidth}
        \centering
        \begin{subfigure}{\linewidth}
            \centering
            \includegraphics[width=1.0\linewidth]{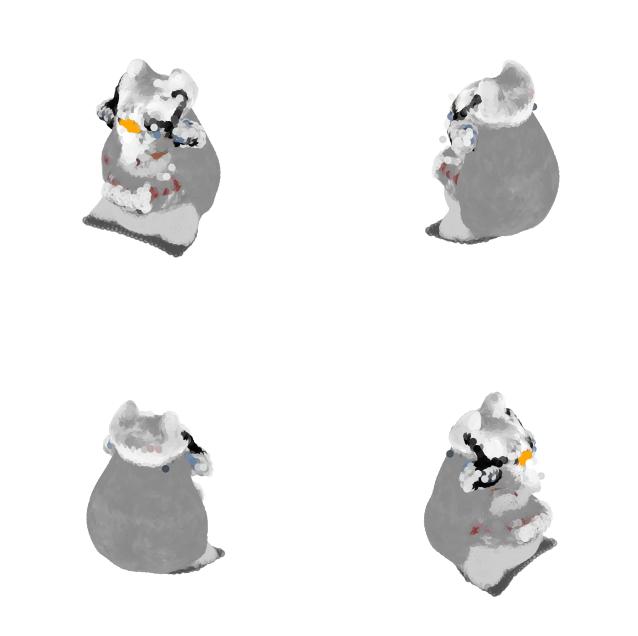} 
        \end{subfigure}
    \end{minipage}
    \hfill
    \begin{minipage}{0.23\textwidth}
        \centering
        \begin{subfigure}{\linewidth}
            \centering
            \includegraphics[width=1.0\linewidth]{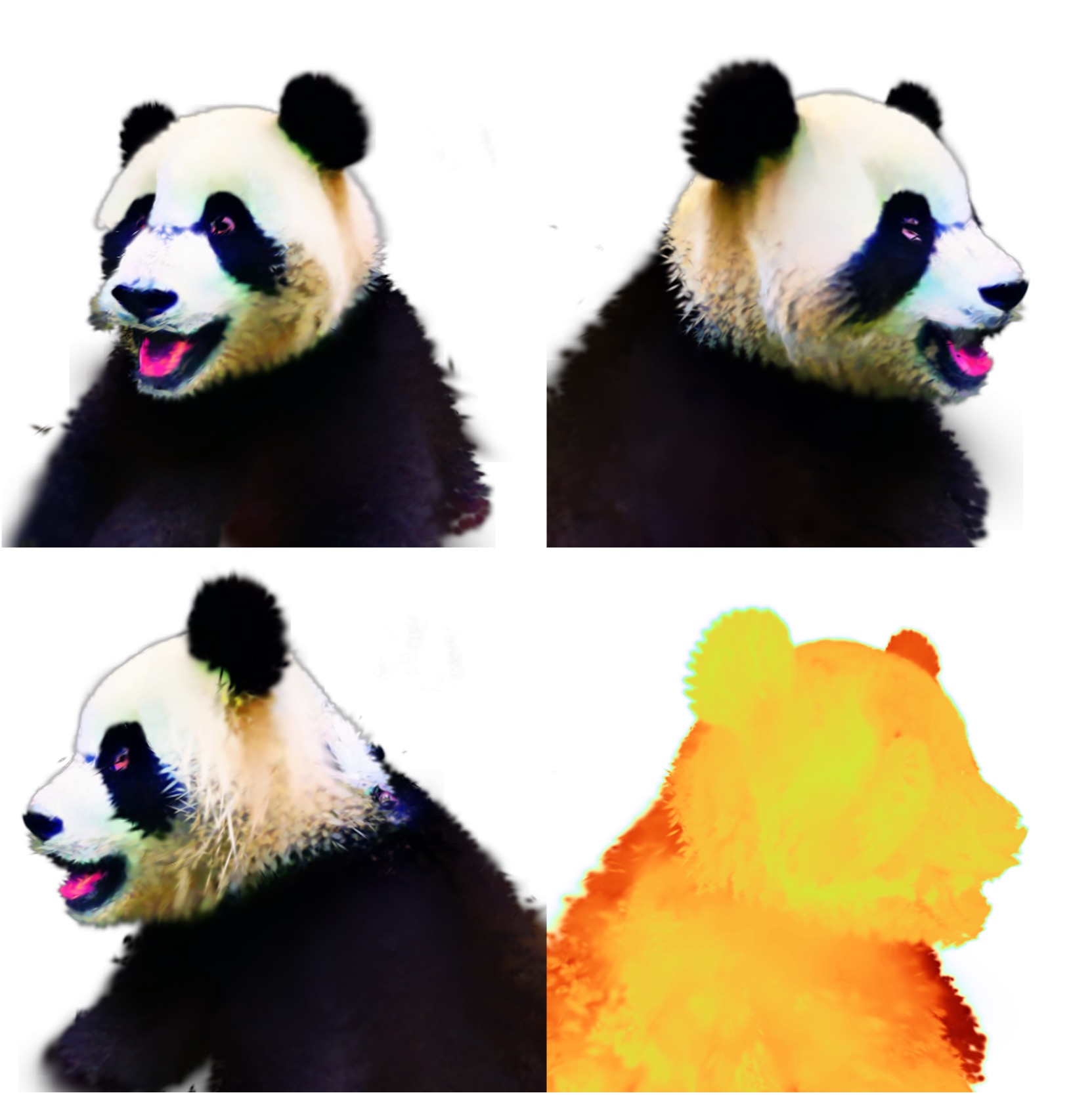} 
        \end{subfigure}
    \end{minipage}
    \\ \textit{a panda}
    \caption{Comparison between Point-E generated point clouds and \approach generated 3D assets. }
    \label{fig:point_e_gsgen}
    \vspace{-5mm}
\end{figure}

\subsection{More Experiments}
\begin{figure}[t]
    \centering
    \begin{minipage}{0.23\textwidth}
        \centering
        \begin{subfigure}{\linewidth}
            \centering
            \includegraphics[width=1.0\linewidth]{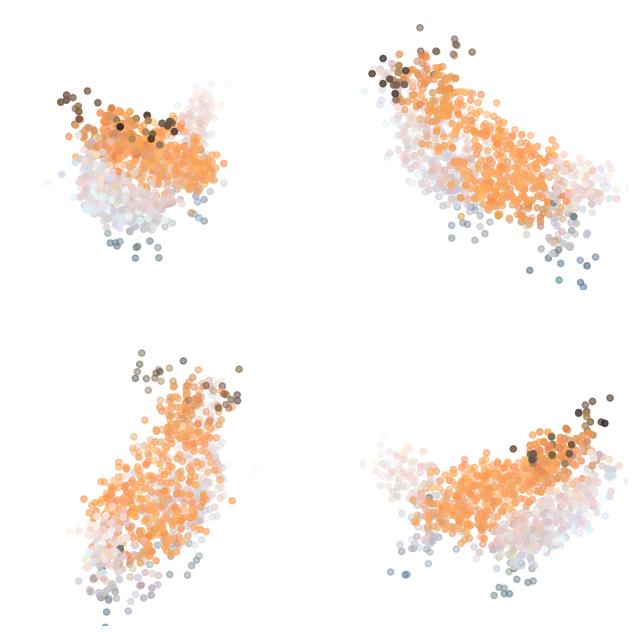} 
            \caption*{Point-E guidance}
        \end{subfigure}
    \end{minipage}
    \hfill
    \begin{minipage}{0.23\textwidth}
        \centering
        \begin{subfigure}{\linewidth}
            \centering
            \includegraphics[width=1.0\linewidth]{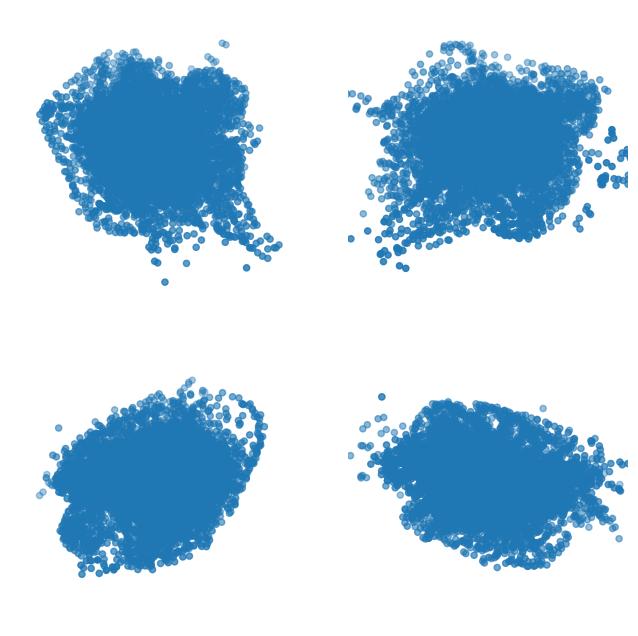} 
            \caption*{ULIP guidance}
        \end{subfigure}
    \end{minipage}
    \caption{Point clouds optimized under \textit{Point-E} and \textit{ULIP}. Prompt: \textit{A corgi}}
    \label{fig:3d_guidance}
\end{figure}
\subsubsection{Color Initialization}
As illustrated in the main text, \approach adopts random color initialization instead of directly applying Point-E generated colors. Fig.~\ref{fig:no_pointe_color} demonstrates the detrimental effect of direct utilization of Point-E generated texture.
\subsubsection{3D Point Cloud Guidance}

\begin{figure}[h]
    \centering
    \begin{minipage}{0.45\textwidth}
        \centering
        \begin{subfigure}{\linewidth}
            \centering
            \includegraphics[width=1.0\linewidth]{comparison/gsgen_sundae.jpg} 
            \caption*{\approach under Point-E guidance}
        \end{subfigure}
    \end{minipage}
    \hfill
    \begin{minipage}{0.45\textwidth}
        \centering
        \begin{subfigure}{\linewidth}
            \centering
            \includegraphics[width=1.0\linewidth]{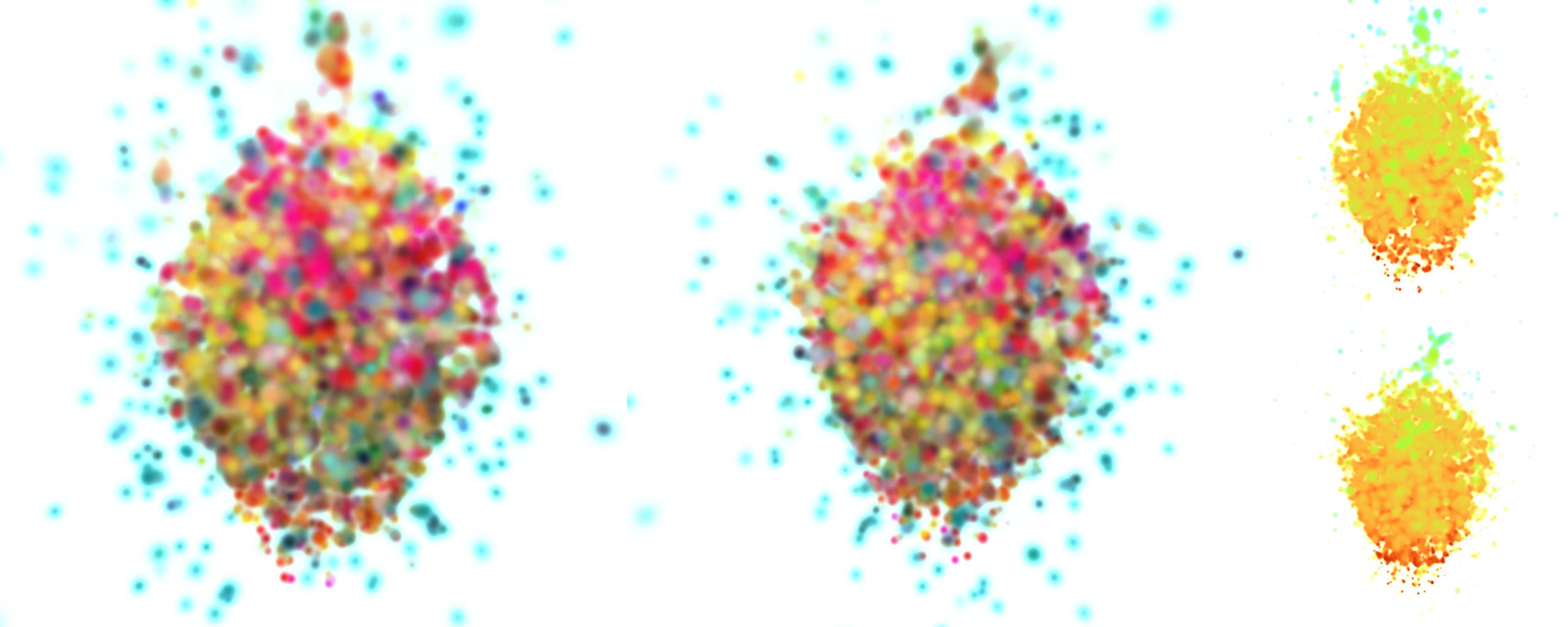} 
            \caption*{\approach under ULIP guidance}
        \end{subfigure}
    \end{minipage}
    \caption{Text-to-3D generation qualitative comparison with 3D prior as Point-E or ULIP. Prompt: \textit{A DSLR photo of an ice cream sundae.}}
    \label{fig:3d_guidance_comparison}
\end{figure}

Our empirical results demonstrate that \approach consistently delivers high performance, even in scenarios where Point-E operates sub-optimally. As illustrated in Fig.~~\ref{fig:point_e_gsgen}, we showcase point clouds generated by Point-E alongside the corresponding 3D assets created by \approach. 
\begin{figure*}[!t]
    \centering
    \begin{minipage}{0.32\textwidth}
        \centering
        \begin{subfigure}{\linewidth}
            \centering
            \includegraphics[width=1.0\linewidth]{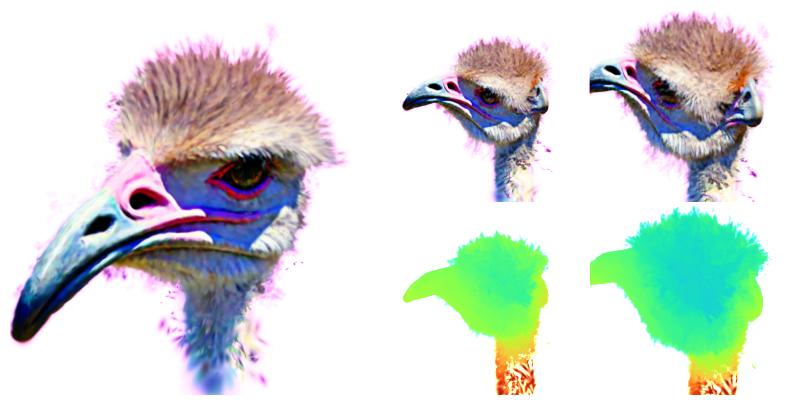} 
            \caption{\textit{\smaller A high quality photo of an ostrich}}
            \label{fig:failed_ostrich}
        \end{subfigure}
    \end{minipage}
    \hfill
    \begin{minipage}{0.32\textwidth}
        \centering
        \begin{subfigure}{\linewidth}
            \centering
            \includegraphics[width=1.0\linewidth]{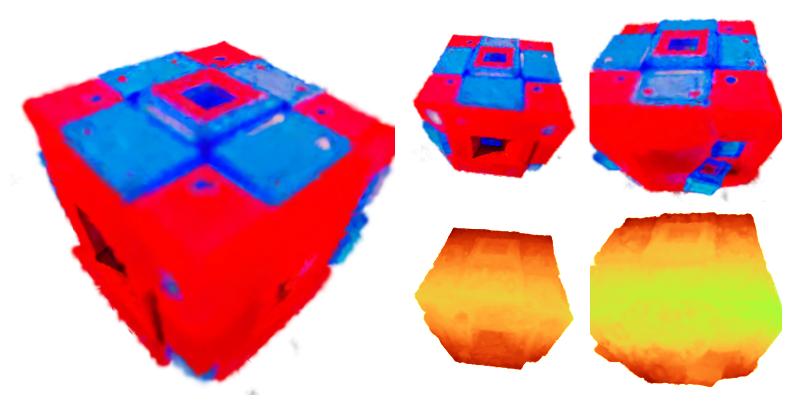} 
            \caption{\textit{\smaller A small red cube is sitting on top of a large blue cube. red on top, blue on bottom}}
            \label{fig:failed_boxes}
        \end{subfigure}
    \end{minipage}
    \hfill
    \begin{minipage}{0.32\textwidth}
        \centering
        \begin{subfigure}{\linewidth}
            \centering
            \includegraphics[width=1.0\linewidth]{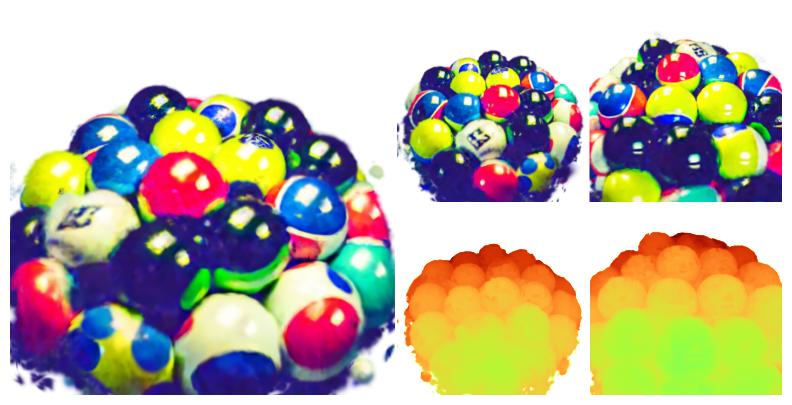} 
            \caption{\textit{\smaller a zoomed out DSLR photo of a few pool balls sitting on a pool table}}
            \label{fig:failed_balls}
        \end{subfigure}
    \end{minipage}
    \caption{Several typical failure cases of \approach.}
    \label{fig:failure_cases}
\end{figure*}
\begin{figure*}[t]
    \centering
    \begin{subfigure}{0.15\textwidth}
        \includegraphics[width=\textwidth]{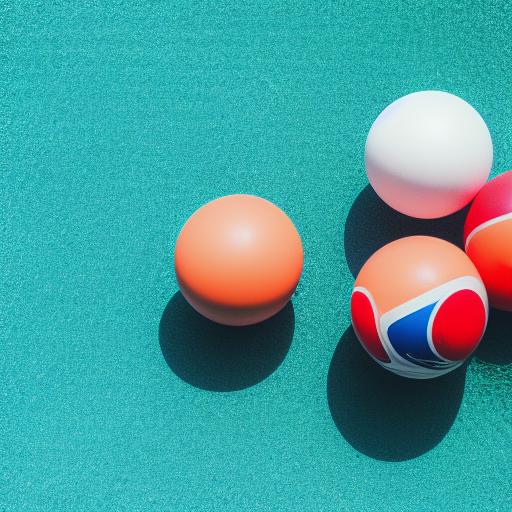}
    \end{subfigure}
    \begin{subfigure}{0.15\textwidth}
        \includegraphics[width=\textwidth]{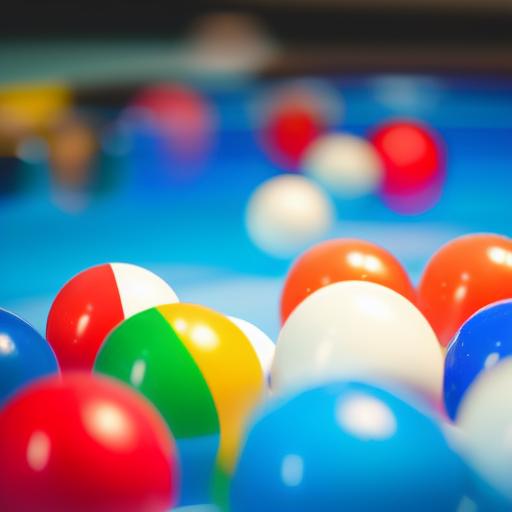}
    \end{subfigure}
    \begin{subfigure}{0.15\textwidth}
        \includegraphics[width=\textwidth]{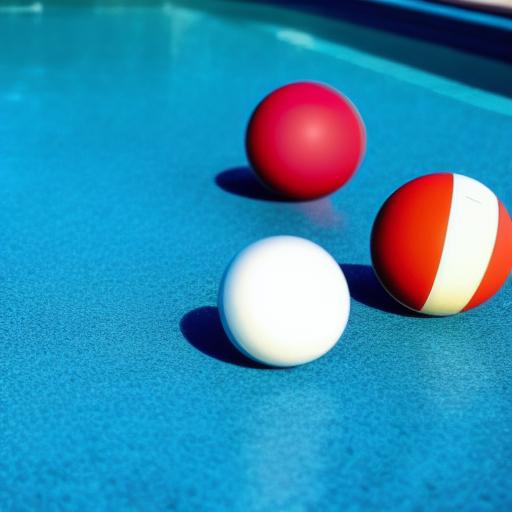}
    \end{subfigure}
    \begin{subfigure}{0.15\textwidth}
        \includegraphics[width=\textwidth]{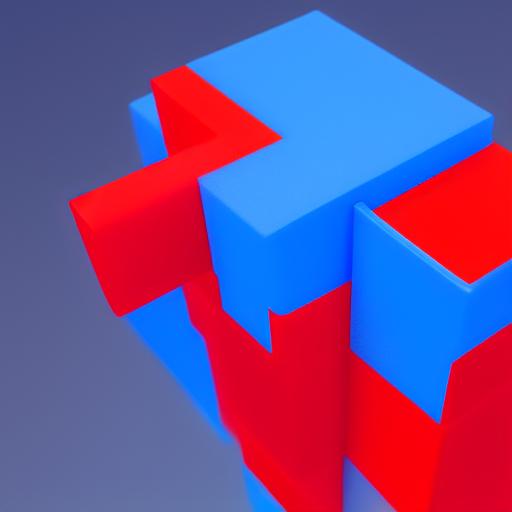}
    \end{subfigure}
    \begin{subfigure}{0.15\textwidth}
        \includegraphics[width=\textwidth]{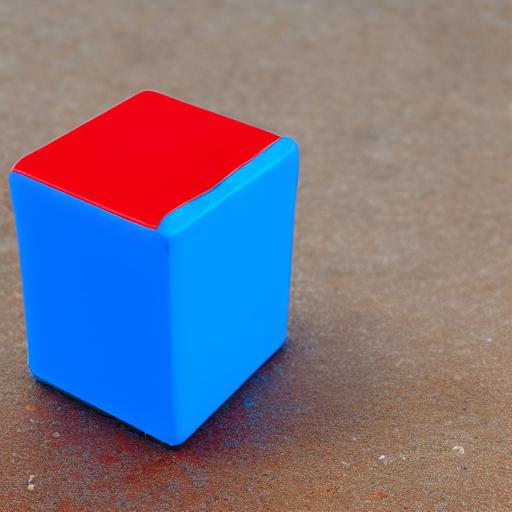}
    \end{subfigure}
    \begin{subfigure}{0.15\textwidth}
        \includegraphics[width=\textwidth]{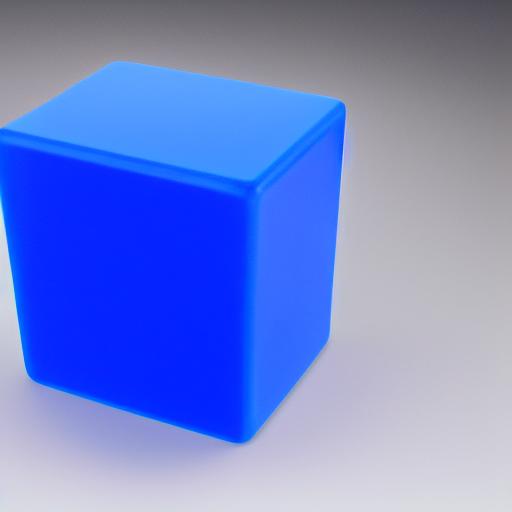}
    \end{subfigure}
    \caption{Prompts that StableDiffusion cannot correctly process, which leads to the failure of corresponding text-to-3D generation.}
    \label{fig:sd_fail}
\end{figure*}
Our approach demonstrates great performance when Point-E provides only rough guidance. We attribute this to direct 3D prior provided by Point-E assisting in geometrical consistency by correcting major shape deviations in the early stage, without the need to guide fine-grained geometric details.

Except for the Point-E \citep{pointe} used in our proposed \approach, we also test a CLIP-like point cloud understanding model ULIP \citep{xue2022ulip, xue2023ulip2}. While achieving superior performance in zero-shot point cloud classification, ULIP seems ineffective in the context of generation. Fig.~\ref{fig:3d_guidance} demonstrates point clouds generated under the guidance of ULIP and Point-E. Point-E can guide the point cloud to a consistent rough shape with SDS loss while the inner-product similarity provided by ULIP leads to a mess. We substitute the 3D prior in \approach from Point-E to ULIP in Fig.~\ref{fig:3d_guidance_comparison}, yielding similar results to point cloud optimization.

\subsubsection{2D Image Guidance}
\label{app:ablation_2d_guidance}

\begin{figure*}[ht]
    \vspace{-12mm}
    \centering
    \begin{minipage}{0.49\textwidth}
        \centering
        \begin{subfigure}{\linewidth}
            \centering
            \caption*{\approach with \textit{StableDiffusion}}
            \includegraphics[width=1.0\linewidth]{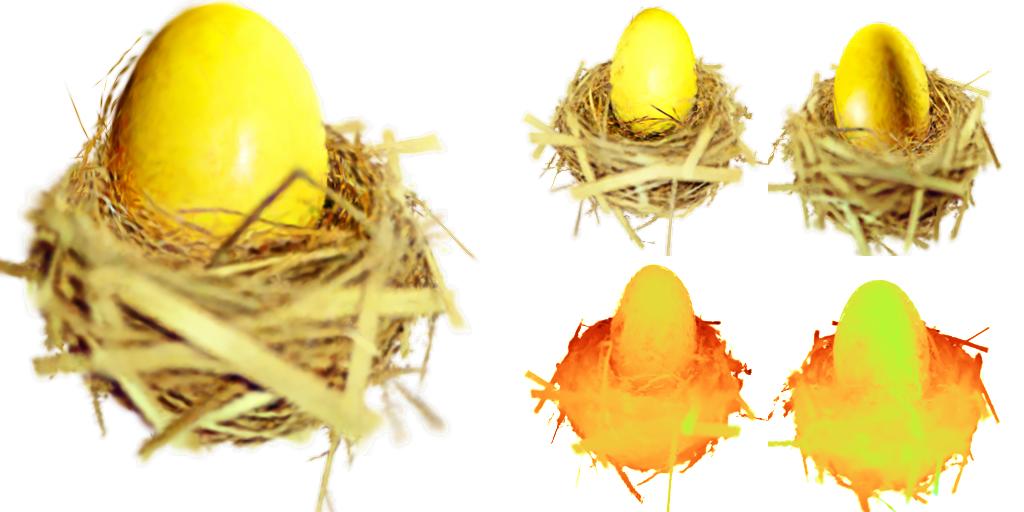} 
        \end{subfigure}
    \end{minipage}
    \hfill
    \begin{minipage}{0.49\textwidth}
        \centering
        \begin{subfigure}{\linewidth}
            \centering
            \caption*{\approach with \textit{DeepFloyd IF}}
            \includegraphics[width=1.0\linewidth]{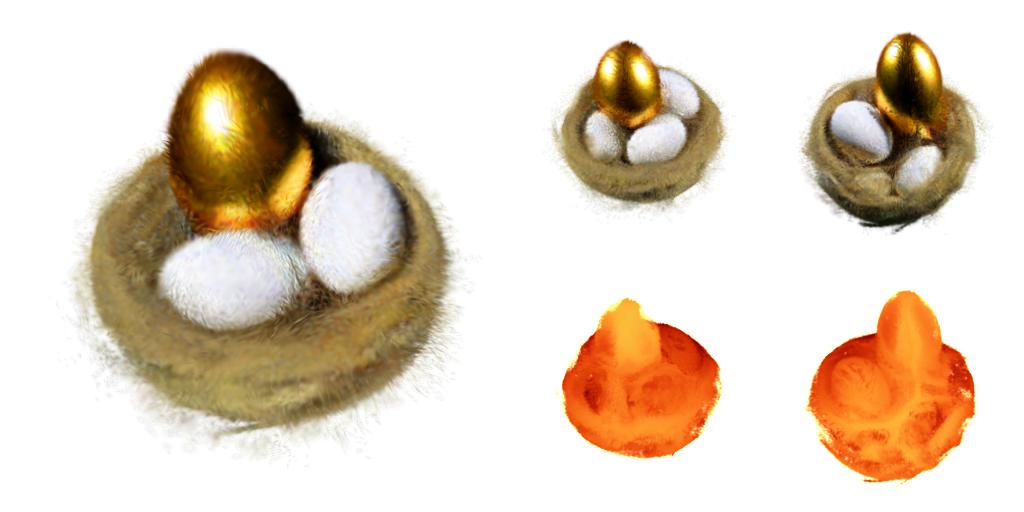} 
        \end{subfigure}
    \end{minipage}
    \\ \textit{A nest with a few white eggs and one golden egg}
    
    \begin{minipage}{0.49\textwidth}
        \centering
        \begin{subfigure}{\linewidth}
            \centering
            \includegraphics[width=1.0\linewidth]{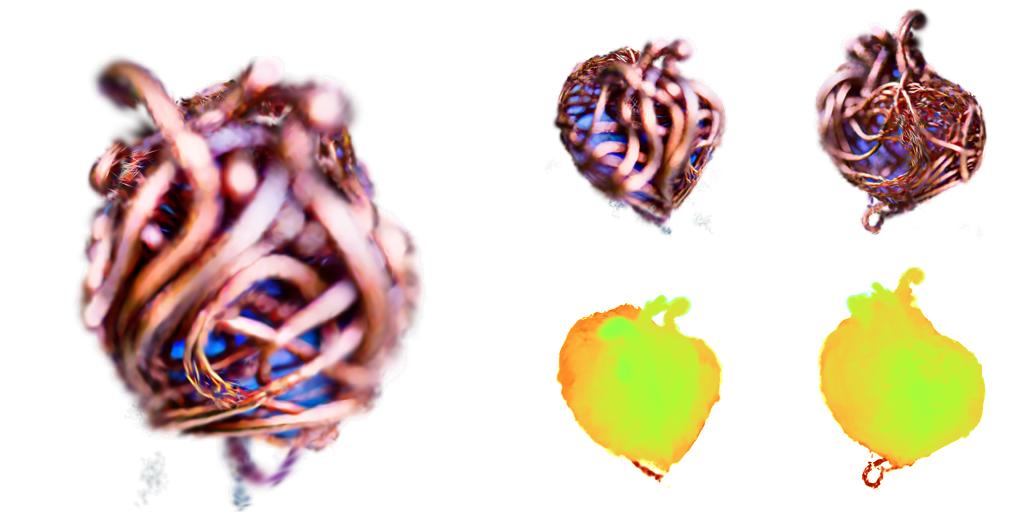} 
        \end{subfigure}
    \end{minipage}
    \hfill
    \begin{minipage}{0.49\textwidth}
        \centering
        \begin{subfigure}{\linewidth}
            \centering
            \includegraphics[width=1.0\linewidth]{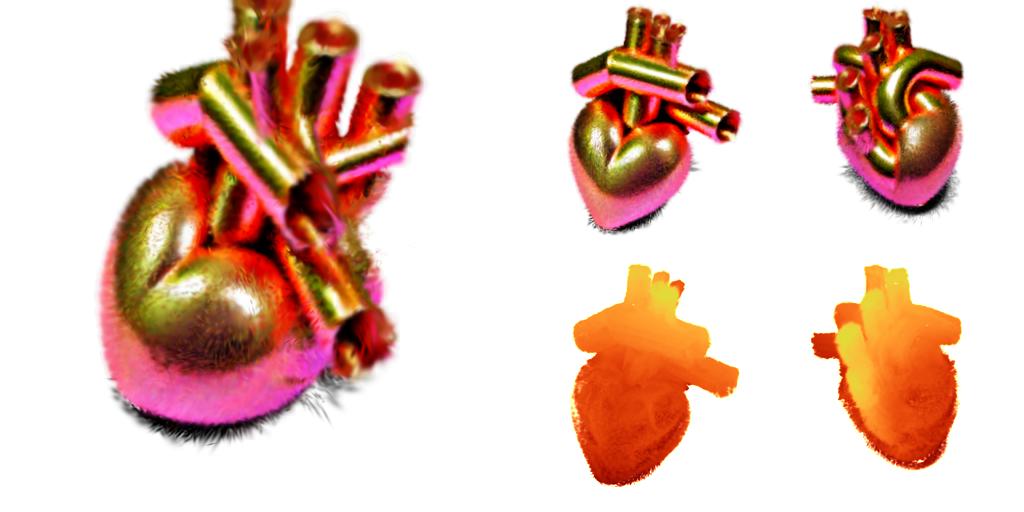} 
        \end{subfigure}
    \end{minipage}
    \\ \textit{A DSLR photo of a very beautiful tiny human heart organic sculpture made of copper wire and threaded pipes, very intricate, curved, Studio lighting, high resolution}

    \begin{minipage}{0.49\textwidth}
        \centering
        \begin{subfigure}{\linewidth}
            \centering
            \includegraphics[width=1.0\linewidth]{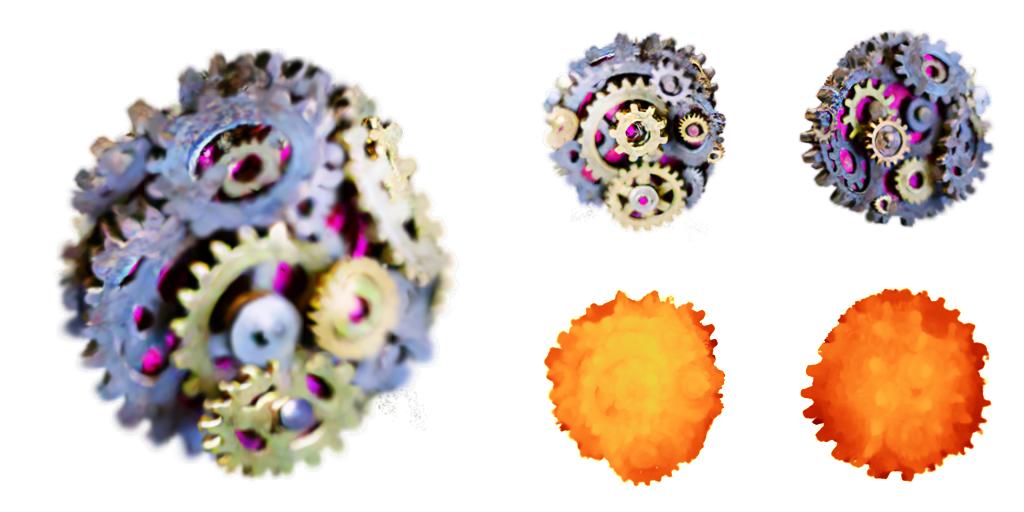} 
        \end{subfigure}
    \end{minipage}
    \hfill
    \begin{minipage}{0.49\textwidth}
        \centering
        \begin{subfigure}{\linewidth}
            \centering
            \includegraphics[width=1.0\linewidth]{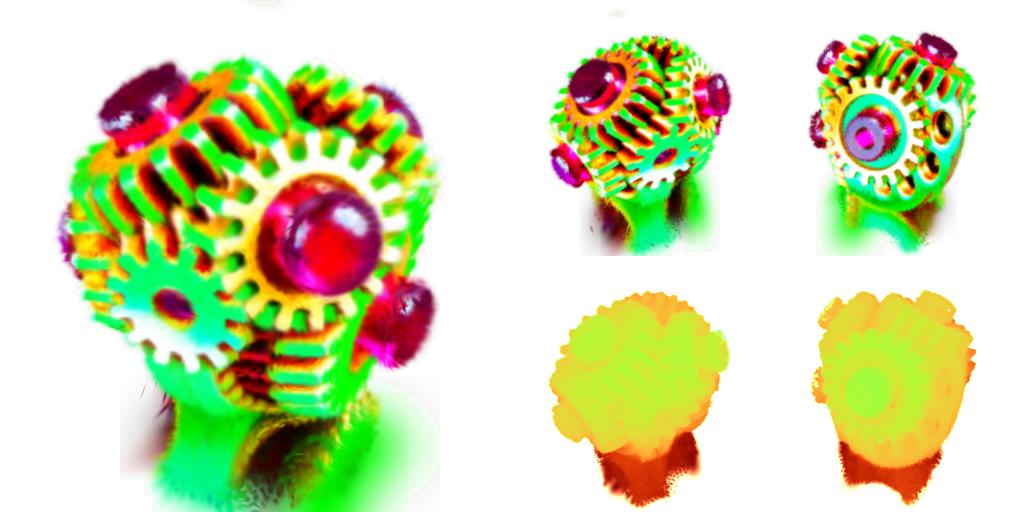} 
        \end{subfigure}
    \end{minipage}
    \\ \textit{A DSLR photo of a very beautiful small organic sculpture made of fine clockwork and gears with tiny ruby bearings, very intricate, caved, curved. Studio lighting, High resolution}

    \begin{minipage}{0.49\textwidth}
        \centering
        \begin{subfigure}{\linewidth}
            \centering
            \includegraphics[width=1.0\linewidth]{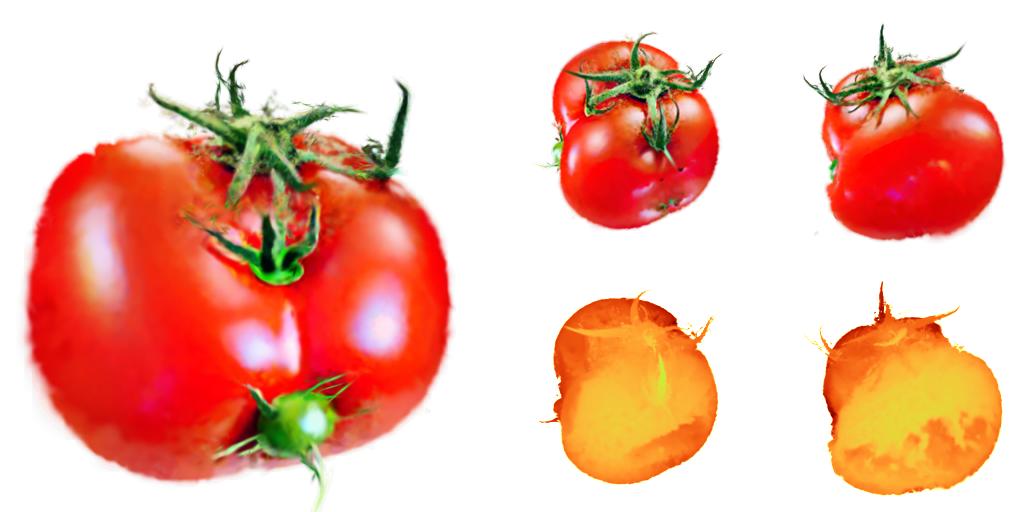} 
        \end{subfigure}
    \end{minipage}
    \hfill
    \begin{minipage}{0.49\textwidth}
        \centering
        \begin{subfigure}{\linewidth}
            \centering
            \includegraphics[width=1.0\linewidth]{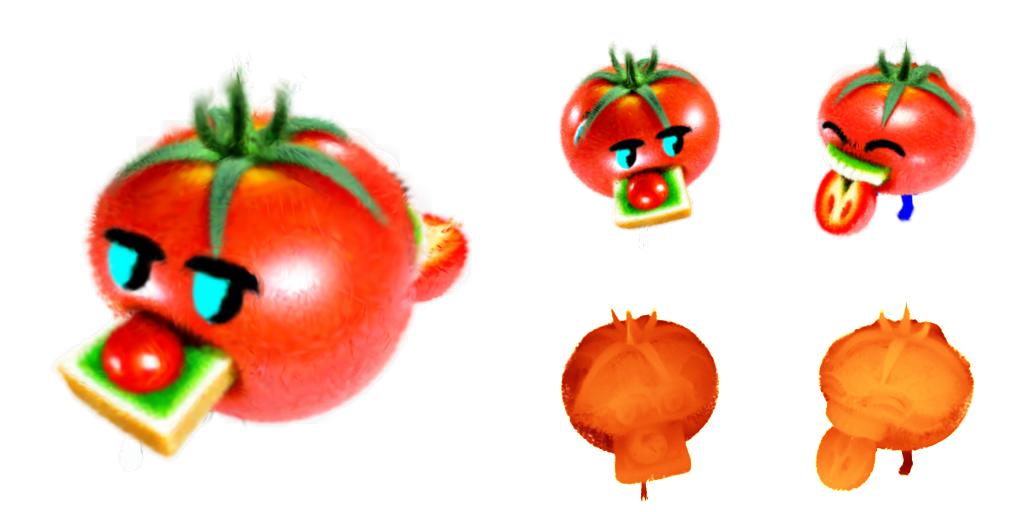} 
        \end{subfigure}
    \end{minipage}
    \\ \textit{An anthropomorphic tomato eating another tomato}
    
    \caption{Qualitative comparison of \approach under StableDiffusion guidance and DeepFloyd IF guidance.}
    \label{fig:sd_vs_if}
    \vspace{-4mm}
\end{figure*}

Except for StableDiffusion, we also test the performance of \approach under the guidance of \textit{DeepFloyd IF}, another open-sourced cutting-edge text-to-image diffusion model. Compared to StableDiffusion, DeepFloyd IF has an Imagen-like architecture and a much more powerful text encoder. We demonstrate the qualitative comparison between \approach under different guidance in Fig.~\ref{fig:sd_vs_if}. 
Obviously, assets generated with \textit{DeepFloyd IF} have a much better text-3D alignment, which is primarily attributed to the stronger text understanding provided by T-5 encoder than that of CLIP text encoder. However, due to the modular cascaded design, the input to \textit{DeepFloyd IF} has to be downsampled to $64 \times 64$, which may result in a blurry appearance compared to those generated under StableDiffusion.

\begin{figure*}[ht]
    \vspace{-5mm}
    \centering
    \begin{minipage}{0.32\textwidth}
        \centering
        \begin{subfigure}{\linewidth}
            \centering
            \includegraphics[width=1.0\linewidth]{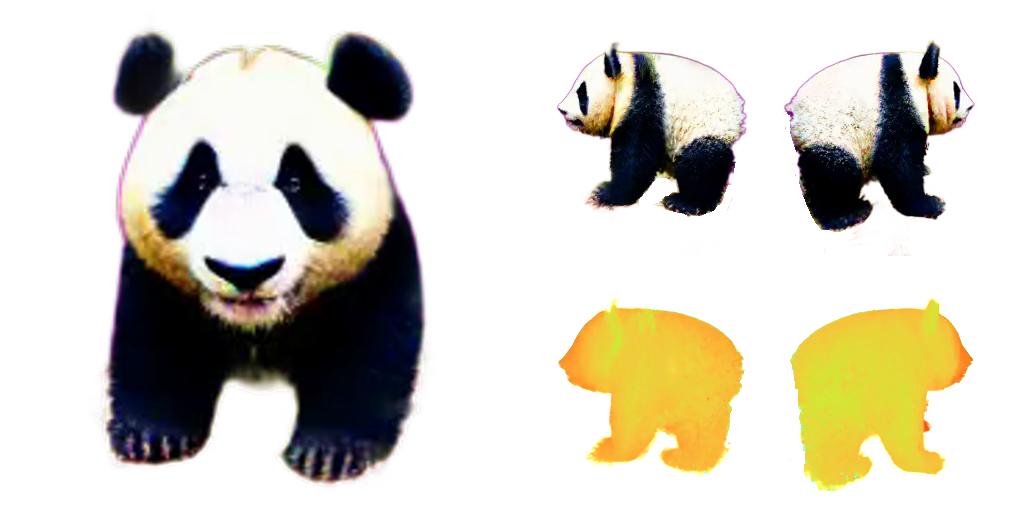} 
            \caption*{\textit{A DSLR photo of a panda}}
        \end{subfigure}
    \end{minipage}
    \hfill
    \begin{minipage}{0.32\textwidth}
        \centering
        \begin{subfigure}{\linewidth}
            \centering
            \includegraphics[width=1.0\linewidth]{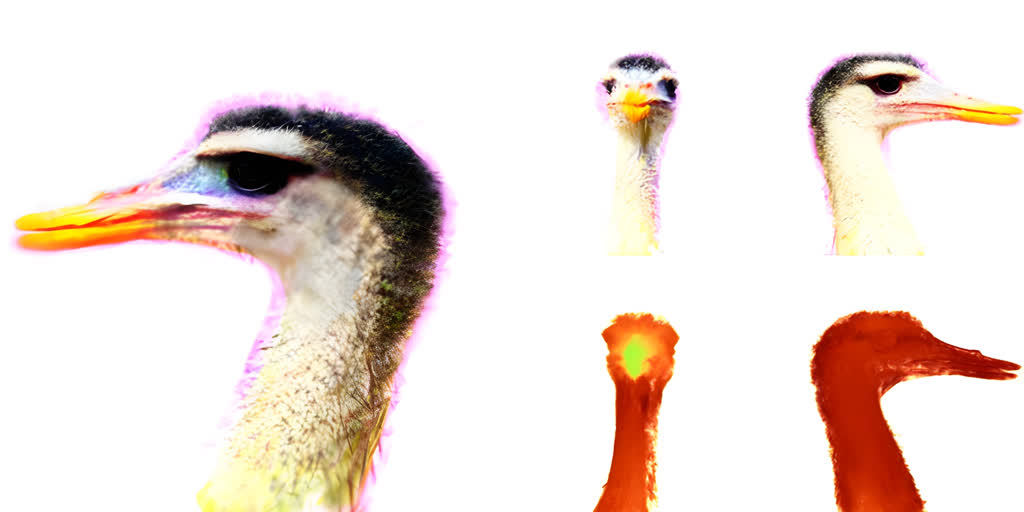} 
            \caption*{\textit{A DSLR photo of an ostrich}}
        \end{subfigure}
    \end{minipage}
    \hfill
    \begin{minipage}{0.32\textwidth}
        \centering
        \begin{subfigure}{\linewidth}
            \centering
            \includegraphics[width=1.0\linewidth]{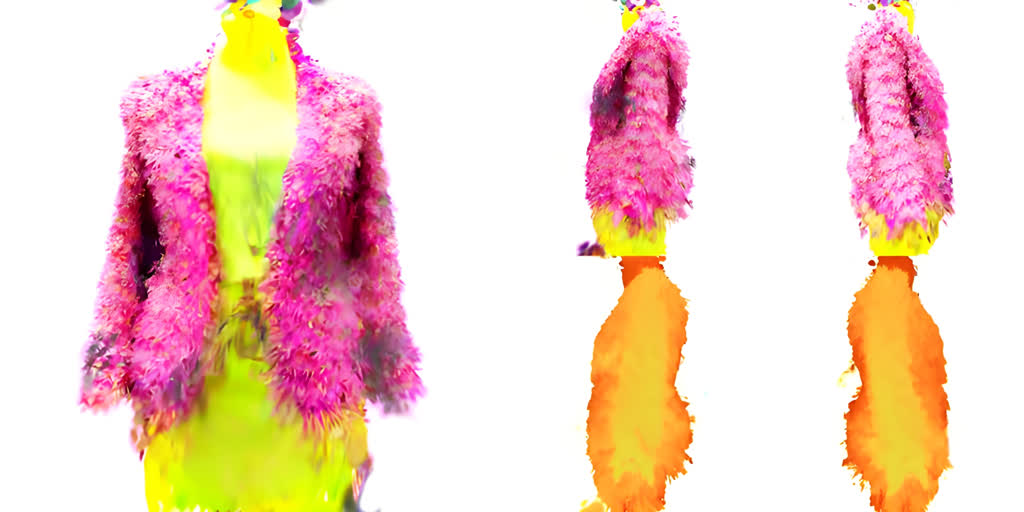} 
            \caption*{\textit{A beautiful suit made out of feather}}
        \end{subfigure}
    \end{minipage}
    
    \caption{3D assets generated under the guidance of our concurrent work MVDream~\citep{shi2023MVDream}. MVDream helps generate more accurate geometry and alleviate the Janus problem, e.g. more complete panda and suit, and the Janus-free ostrich.}
    \label{fig:mvdream}
\end{figure*}

\begin{figure*}[ht]
    \vspace{-10mm}
    \centering
    \begin{minipage}{0.49\textwidth}
        \centering
        \begin{subfigure}{\linewidth}
            \centering
            \caption*{MVDream}
            \includegraphics[width=1.0\linewidth]{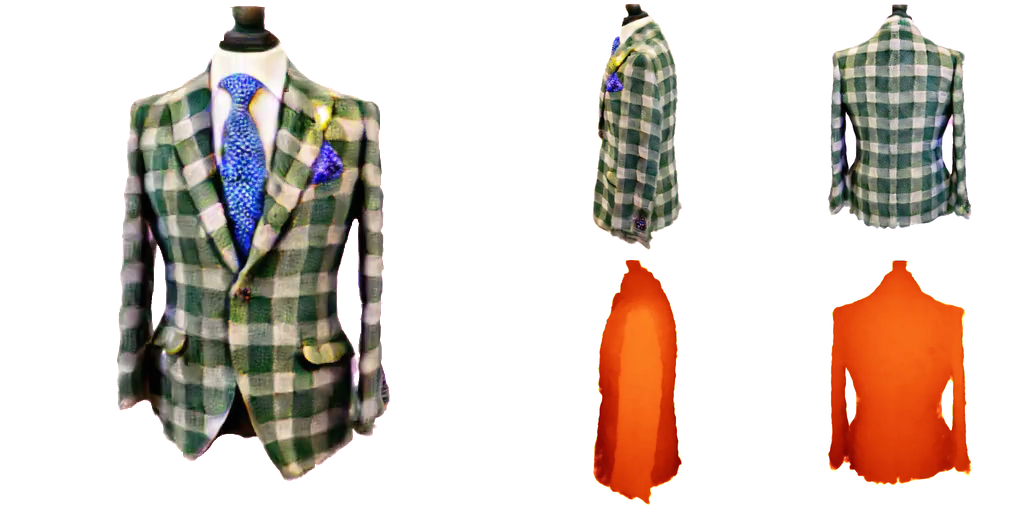} 
        \end{subfigure}
    \end{minipage}
    \begin{minipage}{0.49\textwidth}
        \centering
        \begin{subfigure}{\linewidth}
            \centering
            \caption*{\approach}
            \includegraphics[width=1.0\linewidth]{app/comparison/gsgen/A_zoomed_out_DSLR_photo_of_a_beautiful_suit_made_out_of_moss,_on_a_mannequin.jpg} 
        \end{subfigure}
    \end{minipage}
    \\ \textit{A zoomed out DSLR photo of a beautiful suit made out of moss, on a mannequin. Studio lighting, high quality, high resolution}

    \begin{minipage}{0.49\textwidth}
        \centering
        \begin{subfigure}{\linewidth}
            \centering
            \includegraphics[width=1.0\linewidth]{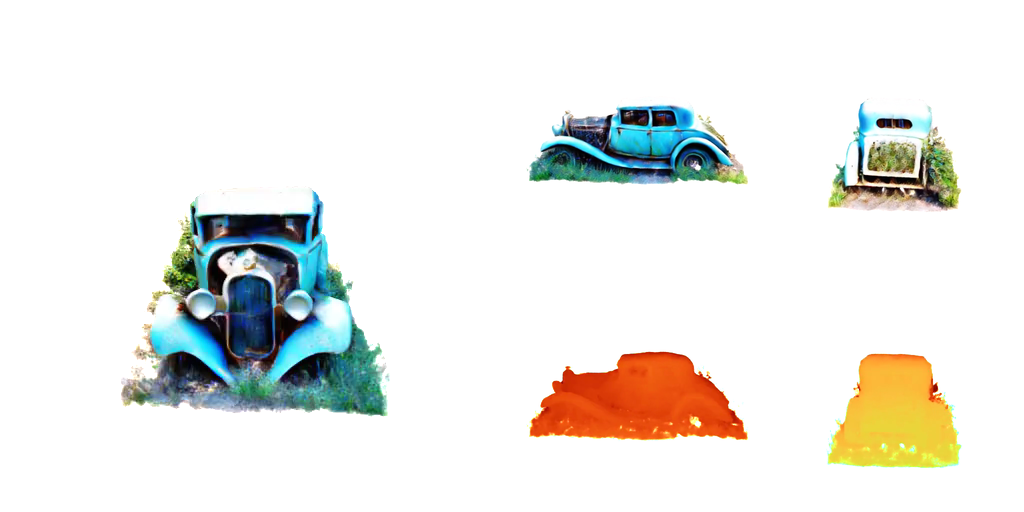} 
        \end{subfigure}
    \end{minipage}
    \begin{minipage}{0.49\textwidth}
        \centering
        \begin{subfigure}{\linewidth}
            \centering
            \includegraphics[width=1.0\linewidth]{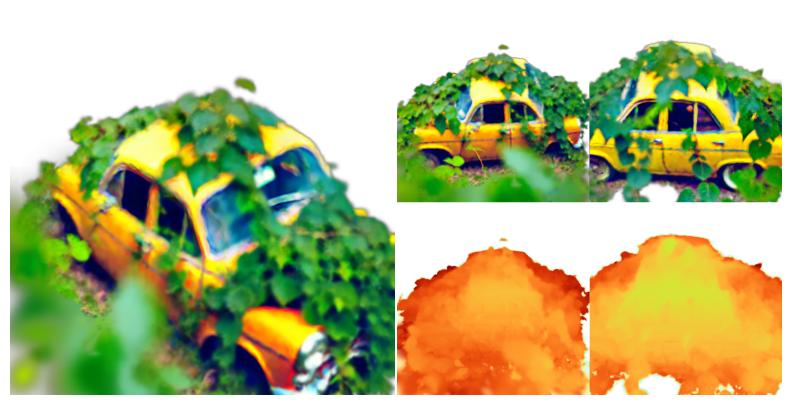} 
        \end{subfigure}
    \end{minipage}
    \\ \textit{A DSLR photo of an old car overgrown by vines and weeds}

    \begin{minipage}{0.49\textwidth}
        \centering
        \begin{subfigure}{\linewidth}
            \centering
            \includegraphics[width=1.0\linewidth]{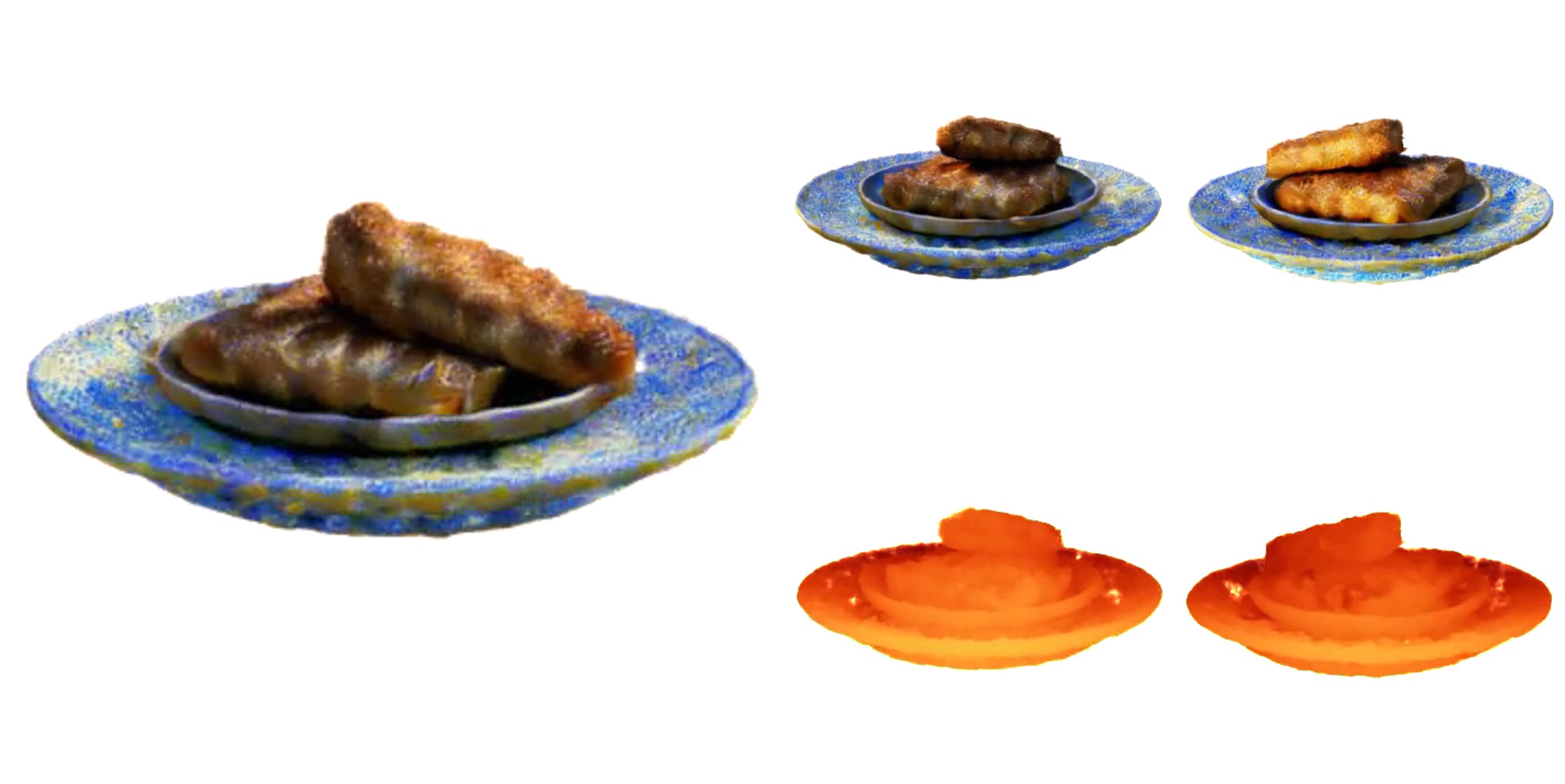} 
        \end{subfigure}
    \end{minipage}
    \begin{minipage}{0.49\textwidth}
        \centering
        \begin{subfigure}{\linewidth}
            \centering
            \includegraphics[width=1.0\linewidth]{app/results/A_zoomed_out_DSLR_photo_of_a_plate_of_fried_chicken_and_waffles_with_maple_syrup_on_them.jpg} 
        \end{subfigure}
    \end{minipage}
    \\ \textit{A zoomed out DSLR photo of a plate of fried chicken and waffles with maple syrup on them}

    \caption{Qualitative comparison of MVDream and \approach with StableDiffusion on complex prompts. }
    \label{fig:mvdream-gsgen}
    \vspace{-4mm}
\end{figure*}

Our concurrent work MVDream~\citep{shi2023MVDream} proposes to fine-tune StableDiffusion with 3D aware components on Objaverse~\citep{objaverse, objaverse-xl} to enhance multi-view consistency and alleviate the Janus problem. We also test the performance of \approach under MVDream. As shown in Fig.~\ref{fig:mvdream}, MVDream significantly contributes to multi-view consistency, resulting in more accurate geometry and complete 3D assets (such as more complete panda and Janus-free ostrich).
Although alleviating the Janus problem, we empirically find that MVDream demonstrates sub-optimal performance towards complex prompts, as shown in Fig.~\ref{fig:mvdream-gsgen}. 3D assets generated with MVDream tend to ignore some parts of the prompt compared to those under StableDiffusion guidance, e.g. the moss on the suit, the vines on the car, and the chicken and waffles on the plate. This demonstrates that introducing 3D prior while retaining the information from the original diffusion model presents a challenging problem, and we consider this issue as our future research.

\section{Failure Cases}
\label{app:failure_cases}

Despite the introduction of direct 3D prior, we could not completely eliminate the Janus problem, attributed to the ill-posed nature of text-to-3D generation through a 2D prior and the limited capability of the 3D prior we employed.

Fig.~\ref{fig:failure_cases} showcases several typical failure cases we encountered in our experiments. In Fig.~\ref{fig:failed_ostrich}, the geometrical structure is correctly established, but the Janus problem happens on the appearance (another ostrich head on the back head). Fig.~~\ref{fig:failed_boxes} and Fig.~~\ref{fig:failed_balls} demonstrates another failure case caused by the limited language understanding of the guidance model. StableDiffusion also fails to generate reasonable images with these prompts, as illustrated in Fig.~\ref{fig:sd_fail}.

    
\end{document}